\documentclass{article} % For LaTeX2e
\usepackage{iclr2027_conference,times}

\definecolor{cvprblue}{rgb}{0.21,0.49,0.74}
\definecolor{lightcarminepink}{rgb}{0.9, 0.4, 0.38}
\usepackage[pagebackref,breaklinks,colorlinks,citecolor=gray,linkcolor=BrickRed]{hyperref}
\usepackage{url}
\renewcommand{\cite}{\citep}

\title{PruneForget: Joint Unlearning and\\ Pruning of Vision Models}
\author{Yu-Shan Tai, Amber Yijia Zheng \& Raymond A. Yeh \\
Department of Computer Science\\
Purdue University\\
\texttt{\{tai50,zheng709,rayyeh\}@purdue.edu} \\
}

\iclrfinalcopy % Uncomment for camera-ready version, but NOT for submission.
\newcommand*{\ShowNotes}{} %Exist then 
\usepackage[T1]{fontenc}
\usepackage[utf8]{inputenc}

\usepackage{amsmath, amssymb, amsfonts}
\usepackage{mathtools}
\usepackage{stmaryrd}
\usepackage{nicefrac}
\usepackage{bm} % Better than \mathbf for bold math
\usepackage{pifont}

\usepackage{graphicx}
\usepackage{booktabs}   % Only need this once
\usepackage{tabularx}
\usepackage{makecell}
\usepackage{multirow}
\usepackage{colortbl}
\usepackage[export]{adjustbox}
\usepackage{wrapfig}
\usepackage{arydshln} % For dashed lines in tables

\usepackage{url}
\usepackage{microtype}
\usepackage{enumitem}
\usepackage{listings}
\usepackage{comment}
\usepackage{pmboxdraw}
\usepackage[multiple]{footmisc}

\usepackage{tikz}
\usepackage[accsupp]{axessibility} 
\usepackage{orcidlink}

\renewcommand{\bibnumfmt}[1]{[#1]}
\usepackage{amsmath}
\usepackage{amsfonts}
\usepackage{bm}

\def\eps{{\epsilon}}

\def\1{\bm{1}}

\def\vm{{\bm{m}}}

\def\vx{{\bm{x}}}

\DeclareMathAlphabet{\mathsfit}{\encodingdefault}{\sfdefault}{m}{sl}
\SetMathAlphabet{\mathsfit}{bold}{\encodingdefault}{\sfdefault}{bx}{n}

\def\gD{{\mathcal{D}}}

\def\gG{{\mathcal{G}}}

\def\gL{{\mathcal{L}}}

\def\gR{{\mathcal{R}}}
\def\gS{{\mathcal{S}}}
\def\gT{{\mathcal{T}}}
\def\gU{{\mathcal{U}}}

\def\gX{{\mathcal{X}}}

\newcommand{\norm}[1]{\left\lVert#1\right\rVert}

\DeclareMathOperator*{\argmin}{arg\,min}

\input{includes/macros.tex}
\usepackage{color}
\usepackage{soul}
\usepackage{xcolor}
\usepackage{wrapfig}
\usepackage[dvipsnames]{xcolor}

\definecolor{darkred}{rgb}{0.5, 0.0, 0.0}
\definecolor{lightred}{rgb}{1.0, 0.9, 0.9}
\definecolor{darkgreen}{rgb}{0.0, 0.5, 0.0}
\definecolor{lightgreen}{rgb}{0.9, 1.0, 0.9}
\definecolor{darkblue}{rgb}{0.0, 0.0, 0.5}
\definecolor{lightblue}{rgb}{0.9, 0.9, 1.0}

\definecolor{lightyellow}{rgb}{1.0, 1.0, 0.8}

\definecolor{orange}{rgb}{1.0, 0.5, 0.0}
\definecolor{cyan}{rgb}{0.7, 0.0, 0.7}
\definecolor{maroon}{rgb}{0.76, 0.13, 0.28}
\definecolor{burntorange}{rgb}{0.81, 0.33, 0}
\definecolor{tealblue}{rgb}{0.212, 0.459, 0.533}
\definecolor{skyblue}{rgb}{0.3 0.7, 1.0}
\definecolor{turquoise}{cmyk}{0.65,0,0.1,0.1}
\definecolor{purple}{rgb}{0.67, 0.28, 0.9}
\definecolor{darkpink}{rgb}{0.98,0.81,0.89}
\definecolor{lightgray}{gray}{0.88}
\definecolor{lightpurple}{rgb}{1.0, 0.9, 1.0}
\definecolor{mygreen}{rgb}{0.5, 0.7, 0.0}
\definecolor{brinkpink}{rgb}{0.98, 0.38, 0.5}

\definecolor{eccvblue}{rgb}{0.12,0.49,0.85}

\definecolor{pp}{rgb}{0.43921569, 0.18823529, 0.62745098}
\definecolor{rr}{rgb}{0.5254902 , 0.00784314, 0.12941176}
\definecolor{bb}{rgb}{0.09019608, 0.23529412, 0.37647059}
\definecolor{yy}{rgb}{0.49803922, 0.3372549 , 0.0}
\definecolor{gg}{rgb}{0.02352941, 0.3372549 , 0.17647059}
\definecolor{ourscolor}{rgb}{0.9, 1.0, 0.75}
\ifdefined\ShowNotes
  \newcommand{\colornote}[3]{{\color{#1}\bf{#2: #3}\normalfont}}
\else
  \newcommand{\colornote}[3]{}
\fi

\newif\ifproofread

\definecolor{checkcolor}{HTML}{305AFF}
\definecolor{crosscolor}{HTML}{E62020}
\definecolor{colorPretrain}{HTML}{E8F5E9} % Very soft mint green
\definecolor{colorOracle}{gray}{0.86}
\definecolor{colorBest}{HTML}{0072B2} 
\definecolor{colorSecondBest}{RGB}{0, 0, 0} 
\definecolor{stage1bg}{HTML}{F8CBAD} % Muted Orange
\definecolor{stage2bg}{HTML}{FFF2CC} % Muted Yellow
\definecolor{stage3bg}{HTML}{E2EFDA} % Muted Green
\definecolor{avg_highlight}{RGB}{230, 240, 250}

\usepackage[most]{tcolorbox}
\usepackage{xcolor}
\usepackage{amsmath}
\usepackage[export]{adjustbox}
\usepackage{graphicx}
\usepackage{subcaption}
\usepackage{multirow}
\usepackage{algorithm}
\usepackage[noend]{algpseudocode}
\usepackage{amsmath,amssymb} 
\usepackage{soul}
\sethlcolor{colorOracle}

\newcommand{\hlblue}[1]{{\sethlcolor{colorPretrain}\hl{#1}}}

\newcommand{\hlavg}[1]{{\setlength{\fboxsep}{1.5pt}\colorbox{avg_highlight}{#1}}}

\begin{document}

\maketitle

\begin{abstract}

Machine unlearning and model pruning are increasingly {\it coupled} in the real world. Models must support unlearning requests, \eg, for safety concerns, {\it while also} meeting requirements in latency and memory budget. Until recently, existing works have studied each aspect as an independent problem, \ie, running unlearning and pruning sequentially. In this work, we show that unlearning and pruning are naturally aligned and should be solved jointly to be aware of each other. Intuitively, parameters that encode information of the unlearned samples are natural pruning targets, as unlearning and pruning both call for the ``deletion'' of such parameters. We propose PruneForget, a method that uses the unlearn set as a guide for pruning, so that unlearning and pruning mutually benefit each other. Extensive experiments on image classifiers and generative models show that PruneForget removes the influence of the unlearned samples while producing a more compact model with reduced inference cost. %PruneForget achieves a negligible performance gap relative to an oracle that retrains from scratch for unlearning and then prunes.

% Today these goals are typically addressed in sequence (unlearn, then compress), which is costly and can be counterproductive: unlearning alters the parameter landscape, while pruning choices determine where information can persist. We argue they should be solved jointly. In particular, the unlearn set, \ie, the samples to be removed, provides a direct signal about which parameters carry the undesired influence and are therefore natural candidates for pruning. We propose PruneForget, which uses the unlearn set to guide pruning so that unlearning and pruning mutually benefit each other. Extensive experiments on image classifiers and generative models show that PruneForget removes the influence of the unlearned samples while producing a more compact model with reduced inference cost, with a negligible performance gap relative to an oracle that retrains from scratch for unlearning and then prunes.
\end{abstract}

\section{Introduction}

%%% Why is this topic interesting, and what's the scope %%%
Deploying vision models in real-world applications increasingly requires satisfying two constraints \textit{simultaneously}. This includes \textbf{(a)} compliance with safety or privacy requirements, \eg, GDPR~\cite{EU2016GDPR} and CCPA~\cite{bonta2022california}, and \textbf{(b)} computation requirements for efficient inference. Until recently~\cite{zhang2025llm}, each of these aspects is typically studied by two disjoint areas of machine unlearning or model pruning. 
%
%
%Although both requirements often need to be satisfied in deployment, the two research areas have mostly been studied separately.
%
%
%%% What do people do now, and why it's not good enough %%%
Machine unlearning focuses on approximating the model that would have been obtained if trained \textit{without} the data in a given unlearn set~\cite{thudi2022unrolling, neel2021descent,izzo2021approximate, koh2017understanding,fan2024salun, kurmanji2023towards,gandikota2023erasing, zhang2024forget, heng2023selective,huang2024unified, he2019filter, chundawat2023can, wu2025erasing, SSD_2024, zheng2026designing}. Model pruning aims to remove parameters from a trained model while maintaining performance~\cite{ding2019centripetal, liu2021group, fang2023structural, fang2023depgraph, fang2024isomorphic,dong2017learning, lee2019signal,park2020lookahead, guo2020multi, ding2019approximated, ye2018rethinking, yu2018nisp, gao2021network, lin2020hrank, orseau2020logarithmic, sanh2020movement, han2015deep, yin2026cut}. Both fields have made significant progress, and recent works~\cite{jia2023model, xiao2025right,shirkavand2025efficient} have begun to show that the two areas are not as separated as they may seem. 

\citet{jia2023model} show that imposing sparsity regularization, common in pruning techniques, can be beneficial to unlearning. However, they fall short of actually removing parameters to produce a smaller model. Other works~\cite{xiao2025right,shirkavand2025efficient} have studied unlearning and pruning together, but their pruning strategies do not explicitly consider the effect of the unlearn set. LLM-Eraser by~\citet{zhang2025llm} leverages the unlearn set for pruning, but focuses its study on large language models (LLMs). While it may seem that this technique could transfer to vision models, empirically, we find that a direct application of LLM-Eraser to vision models does not perform well. This motivates us to study machine unlearning and model pruning {\it jointly} for vision models.

\begin{figure}[t]
\centering
\includegraphics[width=\textwidth]{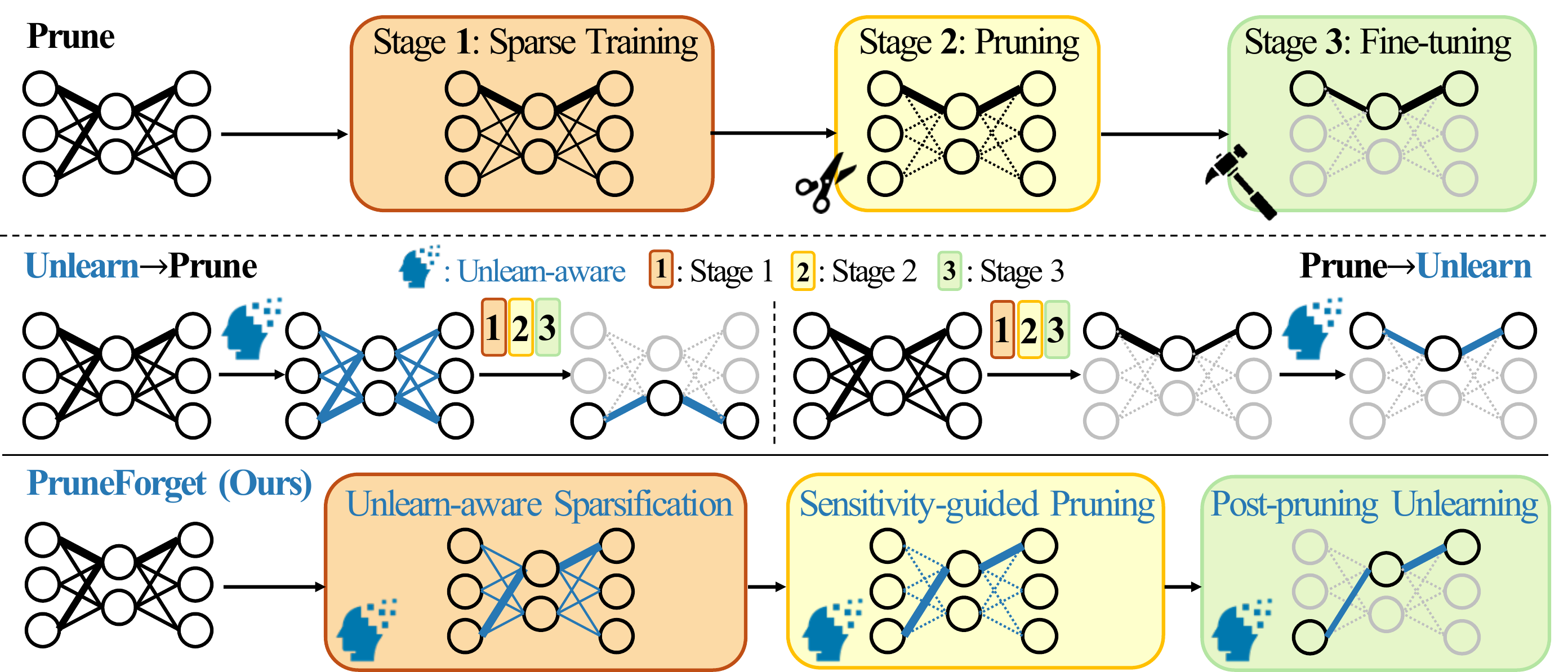}
\vspace{-0.1cm}
\caption{{\bf Sequential methods \vs PruneForget.} 
Model pruning typically consists of three stages: sparse training, pruning, and fine-tuning. Sequential baselines (\textcolor{colorBest}{Unlearn}$\to$Prune or Prune$\to$\textcolor{colorBest}{Unlearn}) treat unlearning and pruning as independent tasks. Our proposed \textcolor{colorBest}{\textbf{PruneForget}} integrates unlearning awareness into the three stages by introducing unlearn-aware sparsification, sensitivity-guided pruning, and post-pruning unlearning; achieving a joint unlearning and pruning framework. %This joint framework is able to achieve an unlearned and pruned model. 
%a seamless, joint optimization of both objectives.
% We introduce unlearning awareness throughout all three stages of the pruning process. PruneForget uses a sensitivity score which measures how important the unlearn set $\gU$
% relative to the retain set $\gD \setminus \gU$ to inform both the sparsification and pruning process. Finally, in the post-pruning stage, we also perform unlearning steps to produce a final pruned and unlearned model.
% \clover{TODO}
}
\label{fig:pipeline}
\vspace{-0.3cm}
\end{figure}

%%% What do we do? %%%
%We propose to study machine unlearning and model pruning {\it jointly} for vision models. 
In this work, we focus on structured pruning methods~\cite{fang2023depgraph, fang2024isomorphic, fang2023structural}, which have been successful in producing smaller vision models that can be efficiently deployed. At a high level, we observe that the set of data to be unlearned should help identify which of the model's parameters to prune, since pruning inherently removes encoded information from a model.
%as one can view ``pruning'' as an approach to remove information from a model. 
We propose \textbf{PruneForget}, which uses the unlearn set to guide structured pruning across all three pruning stages (sparse training, pruning, and fine-tuning). Specifically, we introduce a sensitivity score to quantify the information each parameter group retains with respect to the unlearn set and relative to the remaining data. A comparison between sequential methods vs. ours is provided in~\figref{fig:pipeline}.
%this metric informs both the training and pruning processes. See a visual overview of PruneForget in~\figref{fig:pipeline}.

%%% Results %%%%
To validate our method, we performed extensive unlearning experiments across image classification and generation, subject to target pruning ratios.
For classification, we evaluate on CIFAR-10~\cite{krizhevsky2009learning} and TinyImageNet~\cite{le2015tiny} with ResNet~\cite{he2016deep} and Swin Transformer~\cite{liu2021swin}. For generative model unlearning, we conduct experiments on CIFAR-10 and ImageNet~\cite{deng2009imagenet} using DDPM~\cite{ddpm} and DiT~\cite{peebles2023scalable} architectures. The experiments also cover three different pruning techniques~\cite{fang2023depgraph, fang2024isomorphic, fang2023structural}. In both domains, PruneForget achieves a negligible performance gap relative to the oracle (retrain-then-prune) model while requiring less unlearning and pruning time. %\vspace{3pt}\\

%%% Summarized contribution %%%
{\bf\noindent Our main contributions are as follows:}
\vspace{-3pt}
\begin{itemize}[topsep=0pt, leftmargin=12pt]
    \setlength{\itemsep}{2pt}
    \setlength{\parskip}{2.5pt}
    \item We demonstrate that the unlearn set is useful for guiding which model parameters to structurally prune, enabling effective joint pruning and unlearning of vision models.
    \item Our approach introduces a novel \textit{sensitivity score} that measures how much a pruning parameter group encodes information that should be unlearned. We use this score in \textbf{PruneForget}, the proposed three-stage framework for joint unlearning and pruning.
    \item Experiments across image classification and generation, multiple architectures, and pruning methods demonstrate the general applicability of PruneForget. We outperform existing sequential baselines %(prune-then-unlearn and unlearn-then-prune), 
    as well as recent baselines, \eg, LLM-Eraser.
\end{itemize}

\section{Related Work}
\myparagraph{Machine unlearning} aims to produce models that behave as if specific training data were never seen, without the cost of full retraining from scratch~\citep{cao2015towards}.
Methods include gradient-based forgetting~\citep{lu2026machine, thudi2022unrolling, neel2021descent}, influence function approximations~\citep{izzo2021approximate, koh2017understanding}, and Fisher information-guided parameter updates~\citep{golatkar2020eternal, becker2022evaluating, heng2023selective, SSD_2024}.
Fine-tuning-based approaches have since become the dominant paradigm, spanning image classifiers~\citep{fan2024salun, kurmanji2023towards, huang2024unified, jia2023model, chundawat2023can, heng2023selective}, generative models~\citep{gandikota2023erasing, zhang2024forget, heng2023selective, wu2025erasing, huang2024unified}, and LLMs~\citep{zhang2024negative, yuan2025closer, zhang2025llm}. Broadly speaking, these methods all follow the general framework of jointly optimizing an unlearning loss on the unlearn set and a retain loss on the remainder of the dataset~\citep{fan2024salun, yuan2025closer, huang2024unified, kurmanji2023towards, chundawat2023can, zheng2026designing}, with different designs of the loss function and optimization techniques. Regardless of formulation, these unlearning algorithms never modify the deep-net architecture, \ie, parameter count remains the same before and after unlearning. \vspace{4pt}\\
\myparagraph{Model pruning.} As modern deep-nets are typically over-parameterized, model pruning has been introduced to reduce computation and memory overhead. Model pruning generally involves evaluating an importance weight (mainly heuristic-based) to identify and discard less important parameters. The design of an importance weight depends on the specific pruning method. \citet{fang2023depgraph} propose using the $L_2$ norm of grouped parameters, \citet{liu2021group} utilize Fisher information to rank channel importance, and Diff-Pruning~\cite{fang2023structural} utilizes first-order gradients across timesteps in diffusion models to determine which parameters to prune. These approaches are broadly categorized into unstructured \cite{dong2017learning, lee2019signal, park2020lookahead} and structured pruning \cite{ding2019centripetal, liu2021group, fang2023structural, fang2023depgraph, fang2024isomorphic, guo2020multi, zhuang2020neuron, zhang2021aligned,schindler2020parameterized}. While unstructured pruning zeros out individual weights, structured pruning removes entire channels \cite{fang2023depgraph} or blocks \cite{guo2020multi}, enabling hardware speedups without requiring specialized accelerators. Recently, model-specific pruning methods have been tailored for CNNs~\cite{guo2020multi, ding2019approximated, ye2018rethinking}, Transformers \cite{zhu2021vision, yu2022width, mao2021tprune, song2022cp, yin2026cut}, diffusion models~\cite{fang2023structural, ganjdanesh2024not, castells2024ld}, and VLMs~\cite{wu2026collaborative, kang2026porta}. %, alongside the generalized pruning frameworks~\cite{fang2023depgraph, fang2024isomorphic}. 

\myparagraph{Unlearning with pruning.} While both machine unlearning and model pruning remove information from a trained model, existing works treat the two as independent tasks.
In $\ell_1$-MU, \citet{jia2023model} show that model sparsity can facilitate unlearning. However, they \textit{do not} actually prune the model.
Despite the success of existing unlearning and model pruning methods, naively applying the two sequentially often results in ineffective unlearning or reduced performance. 
To bridge this gap, Un-pruning~\cite{xiao2025right} interleaves pruning and unlearning steps, while 
\citet{shirkavand2025efficient} focus specifically on the post-pruning phase and propose a bilevel optimization strategy for joint fine-tuning and unlearning.
LLM-Eraser~\cite{zhang2025llm} further incorporates unlearning directly into the pruning phase to eliminate weights tied to unlearned knowledge in LLMs. We summarize in Appendix~\tabref{tab:method_comparison} that these approaches only partially integrate the two domains into the three stages of pruning (sparse training, pruning, post-pruning fine-tuning). In contrast, our proposed PruneForget makes all three stages unlearn-aware to further improve the unlearning performance of pruned models.

%%% OLD Stuff below %%%

% ~\citet{shirkavand2025efficient}
% ~\citet{jia2023model}
% ~\citet{xiao2025right}
% ~\citet{zhang2025llm}

% \clover{TODO:(1) Why need to combine the two (2) challenges: naive concat doesn't work (3) Existing methods. Review, then compare / contrast. **LLM-Eraser**.
% \begin{itemize}
%     \item L1-sparse MU~\cite{jia2023model}: (1) Motivation: Utilize sparsity to unlearn. Their goal is to unlearn, so they don't really prune the model. (2) L1 norm just uses weight magnitude, they don't consider the unlearned set.
%     \item LLM-Eraser~\cite{zhang2025llm}: (1) Motivation: Unlearn through pruning. (2) Didn't consider unlerned set in regularization term (they use L1).
%     \item Un-pruning~\cite{xiao2025right}: (1) Motivation: Eliminate the influence of unelearned data on pruning. (Similar to ours) (2) Just interleave unlearning \& pruning. (3) Without considering unlearned set during pruning (2nd stage).
%     \item bilevel~\cite{shirkavand2025efficient}: (1) Motivation: For a pruned model, fine-tune then unlearn is not optimal. (2) Only adjust the fine-tune stage (3rd stage).
% \end{itemize}
% }

\section{Preliminaries}\label{sec:prelim}
%We briefly review machine unlearning and model pruning to establish the notation and background.
\myparagraph{Machine unlearning.}
The oracle unlearned model is defined as
\bea\label{eq:unlearn}
\theta' \triangleq \argmin_\theta \gL(\gD \setminus \gU, \theta),
\eea
where a model is retrained using loss $\gL$ from scratch on $\gD\setminus\gU$ by removing the unlearn set $\gU$ from the training set $\gD$. However, such retraining following~\equref{eq:unlearn} is prohibitively expensive for modern deep networks and large datasets. Unlearning algorithms therefore aim to efficiently produce an approximation $\tilde{\theta}$ of the oracle $\theta'$. This approximation $\tilde{\theta}$ is expected to be close to the oracle $\theta'$, \ie, 
\bea 
F_{\tilde{\theta}}(\vx) \approx F_{\theta'}(\vx), \forall \vx \in \gX,
\eea
where $F_\theta(\cdot)$ denotes the model's output. 

Existing methods~\cite{fan2024salun, graves2021amnesiac, zhang2025llm, huang2024unified, kurmanji2023towards, thudi2022unrolling} approximate $\theta'$ by updating the pre-trained model $\theta^\star$ (trained on the full dataset $\gD$) using gradient-based optimization. The difference lies in the design choices of the unlearning loss $\gL_{\tt U}$ and the retain loss $\gL_{\tt R}$, and the gradient update algorithm
\bea
\tilde{\theta} \triangleq \texttt{GD}(\gL_{\tt U}, \gL_{\tt R}, \gU, \gD, \theta^\star),
\eea
where $\texttt{GD}$ denotes gradient descent starting from $\theta^\star$ that optimizes a combination of the unlearning loss $\gL_{\tt U}$ and the retain loss $\gL_{\tt R}$. The unlearning loss $\gL_{\tt U}$ encourages the model to ``forget'' the unlearn set $\gU$, with common choices including maximizing the cross-entropy loss~\cite{graves2021amnesiac} or pushing predictions of the unlearned samples toward a uniform distribution~\cite{zhang2025llm} (we denote this as $\gL_{\tt Uni}$).
Next, the retain loss $\gL_{\tt R}$ is the standard supervised loss on the retain set $\gD \setminus \gU$ to preserve model performance.

\myparagraph{Structured model pruning.} 
Model pruning aims to reduce the computational and memory costs of deep-nets by removing redundant or less important parameters while preserving model performance. Existing approaches~\cite{fang2023depgraph, liu2017learning, you2019gate, zhang2021aligned, fang2024isomorphic, fang2023structural} on structured pruning generally follow a three-step pipeline: sparsity training to prepare weights for removal, pruning to identify and remove unimportant parameters, and fine-tuning to recover performance on the pruned network.

\noindent\textit{(i) Sparse training (optional).}
Directly pruning a trained model $\theta^\star$ tends to remove weights that still carry useful information. To avoid this, weights $\theta$ are fine-tuned with a sparsity-inducing regularizer to encourage unimportant weights to be close to zero, \ie,
\bea\label{eq:sparse-train}
\theta^{\tt s} \triangleq \texttt{GD}(\gL_{\tt prune}, \gD, \theta^\star), \text{ where } \gL_{\tt prune}(\gD,\theta) \triangleq \gL(\gD, \theta) + \lambda \gR({\theta}).
\eea
Here $\gL$ is the standard supervised loss, $\gR$ is a sparsity-inducing regularizer, and $\lambda$ is a hyperparameter that controls the strength of the regularization. A popular choice for the regularizer is proposed by~\citet{fang2023depgraph}, which considers structural dependencies between the weights. In more detail, 
\bea\label{eq:group-reg}
\gR(\theta) \triangleq \sum_{g \in \gG} \sum_{l \in g} \sum_{k \in l} \gamma_{l[k]}\norm{\theta_{l[k]}}_2^2,
\eea
where they construct a set $\gG=\{g\}$ of parameter groups $g$, where layers $l$ are placed into $g$ based on their dependencies in the computational graph. The $k$ indexes the prunable dimension in a layer, \ie, $\theta_{l[k]}$ denotes the $k^{\text{th}}$ prunable weight of layer $l$. Finally, $\gamma$ is a heuristic that determines the penalty amount; please refer to~\citet{fang2023depgraph} for specifics.

%Common choices for $\gR$ include $\ell_1$ regularization, which encourages individual weights to be zero, and group lasso regularization, which encourages entire groups of weights (e.g., rows or blocks) to be zero.

\noindent\textit{(ii) Pruning.}
After the sparse training, pruning is performed by removing unimportant weights, \ie, masking out the parameters for the pruned model
\bea\label{eq:prune}
\theta^{\tt pr} \triangleq \theta^{\tt s} \odot \vm,
\eea
where $\odot$ is element-wise multiplication and $\vm \in \{0,1\}^{\text{DIM}(\theta)}$ is the binary pruning mask. An importance weight $I(\theta_{l[k]})$ is used to determine whether to keep the parameter, which depends on the specific pruning method. For example, in Group-level Pruning~\cite{fang2023depgraph}, the importance weight is determined per group, \ie,
$
I(\theta_{l[k]}) \triangleq \sum_{l'\in g(l)} \norm{\theta_{l'[k]}}_2^2,
$
where $g(l)$ is the parameter group. % that $l$ is in.

\noindent\textit{(iii) Fine-tuning.}
Performance usually drops after model pruning. Hence, to recover the performance, the pruned model is fine-tuned again on the training data. That is, another round of gradient-based updates is applied to obtain the final fine-tuned model $\theta^{\tt f}$.
% \bea\label{eq:finetune}
% \theta^{\tt f} \triangleq \texttt{GD}(\gL, \gD, \theta^{\tt pr}).
% % {\theta}_{\tt p}^\star  \triangleq \argmin_{\theta}\; \gL(\gD,\, \theta \odot \vm).
% \eea 

\section{Method}
%\begin{wrapfigure}[27]{r}{0.59\linewidth}

%
The two objectives of machine unlearning and model pruning are naturally aligned. The unlearn set $\gU$ identifies the very information that must be erased from the model, and the weights most responsible for encoding that information are essentially the ones that should be removed by pruning. Hence, in this work, our goal is to produce a pruned \textbf{and} unlearned model $\tilde{\theta}^{\tt f}$ from a model $\theta^\star$ trained on the full dataset $\gD$. Rather than treating unlearning and pruning as sequential post-hoc procedures, our proposed \texttt{PruneForget} leverages the unlearn set $\gU$ to guide the pruning process, \ie, the unlearning and pruning benefit each other within a single unified framework. The overall algorithm is stated in Alg.~\ref{alg:pruneforget}, where unlearn-aware terms %on the right-hand side (RHS) 
are highlighted in \textcolor{colorBest}{blue}. We now describe the details.  

\begin{figure}[t]
\input{sec/alg_main_new}
\end{figure}
\subsection{PruneForget}
At a high level, we make the three-stage structured pruning pipeline (reviewed in~\secref{sec:prelim}) aware of unlearning. For this, we introduce a per-parameter-group \emph{sensitivity score} $S_{l[k]}$,
which measures how important the unlearn set $\gU$
is relative to the retain set $\gD \setminus \gU$. In more detail, 
\begin{equation}\label{eq:sensitivity}
 S_{l[k]} \triangleq \sqrt{
        %\frac
        {\left(\sum_{l'\in g(l)} \norm{ \nabla_{\theta_{l'[k]}} \gL_{\tt Uni}(\gU,\,\theta) }_{2}^{2}\right)} \bigg/
             {\left(\sum_{l'\in g(l)} \norm{ \nabla_{\theta_{l'[k]}} \gL(\gD \setminus \gU,\,\theta) }_{2}^{2} + \eps\right)}
    },
\end{equation}
where $g(l)$ denotes the parameter group to which layer $l$ belongs, and $\gL_{\tt Uni}$
is the KL divergence of the prediction class probability with a uniform distribution. 
We use the gradient norm as a proxy for importance, due to its effectiveness in the model pruning literature~\cite{fang2024isomorphic, fang2023structural, liu2021group, zhang2025llm}. 
A large $S_{l[k]}$ indicates that the group encodes unlearn-set information more than
retain-set utility, making it a suitable target for removal.
The scores are then standardized, $S'_{l[k]} = \text{Standardize}(S_{l[k]})$, and
passed to the first two stages. % of the approach.

\myparagraph{Stage 1: Unlearn-aware sparsification.}
Before any structured pruning, Stage~1 encourages the forget-sensitive weights toward zero
via a sensitivity-weighted regularizer applied alongside an unlearning procedure. 
Over $T_s$ gradient steps, the model minimizes
\bea\label{eq:stage1}
    \gL_{\tt U} + \gL_{\tt prune} \triangleq 
    \rlap{$\overbrace{\phantom{-\alpha\,\gL(\gU,\theta)+ \eta\,\gL(\gD \setminus \gU,\theta)}}^{\text{Unlearning}}$}
    {-\alpha}\,\gL(\gU,\theta) + \underbrace{\eta\,\gL(\gD \setminus \gU,\theta) +  \gR'(\theta)}_{\text{Pruning}},
\eea
where the sensitivity-weighted regularizer is
\bea
\gR'(\theta) = \lambda \sum_{g \in \gG} \sum_{l \in g} \sum_{k \in l}
S'_{l[k]}\,\gamma_{l[k]}\,\norm{\theta_{l[k]}}_2^2.
\eea
This regularizer differs from the group regularizer reviewed in~\equref{eq:group-reg} in that the penalty on each group is scaled by $S'_{l[k]}$. Intuitively, a high sensitivity score $S'$ pushes the weights that influence the unlearn set to be smaller. At the same time, the gradient ascent term $-\alpha\,\gL(\gU,\theta)$ simultaneously performs unlearning in this stage.

\myparagraph{Stage 2: Sensitivity-guided pruning.}
Structured pruning uses an importance weight $I(\theta_{l[k]})$ to determine which parameters to prune. That is, if $I(\theta_{l[k]})$ is less than a threshold $\tau$, then the parameter is pruned. To make the importance weight aware of the unlearn set, we scale it by the inverse of the proposed sensitivity (\equref{eq:sensitivity}), \ie,
\begin{equation}\label{eq:importance}
    I'(\theta_{l[k]}) \triangleq I(\theta_{l[k]})\;/\;S'_{l[k]}.
\end{equation}
% A binary mask is then obtained by thresholding $I'$ at $\tau$, chosen to satisfy the
% target prune ratio, and the model is pruned as $\theta \leftarrow \theta \odot \vm$.
Dividing by $S'_{l[k]}$ lowers this sensitivity-guided importance weight of the parameters. Parameters with higher sensitivity are more related to the unlearn set; hence, dividing by sensitivity makes the parameter more likely to be pruned. Note that this approach is importance-weight-agnostic, \ie, any existing pruning
method's $I$ can be made unlearn-aware by applying~\equref{eq:importance}.

\myparagraph{Stage 3: Post-pruning unlearning.}
As in model pruning, further fine-tuning is needed to recover the model's performance. As our goal is to have an unlearned model, we perform standard unlearning in this Stage~3, \ie,
%Structural pruning inevitably disrupts the model's predictions on both the retain and unlearn sets.
%Stage~3 recovers the retain-set performance and completes residual forgetting by fine-tuning the pruned model for $T_f$ iterations with
\begin{equation}\label{eq:stage3}
    \theta \leftarrow \theta
    - \nabla_\theta\!\Big(
        \underbrace{-\alpha\,\gL(\gU,\theta)}_{\gL_{\tt U}}
        + \underbrace{\eta\,\gL(\gD \setminus \gU,\theta)}_{\gL_{\tt R}}
    \Big).
\end{equation}
The retain-set term $\gL_{\tt R}$ aims to recover the model's performance, whereas $\gL_{\tt U}$ removes the influence of data in the unlearn set.

\subsection{Practical details of PruneForget} \label{sec:additional_details}
%We now describe several details of the approach for completeness.
\myparagraph{Unlearning loss $\gL_{\tt Uni}$.}
For classification, $\gL_{\tt Uni}$ is the KL divergence between the model's predicted class probabilities and a uniform distribution, as defined in~\secref{sec:prelim}. For generative models, we use the MSE between the model's predicted noise and randomly sampled Gaussian noise, which serves to drive the model's output on $\gU$ toward an uninformative target~\cite{aravindan2025sealing}. 

\myparagraph{Sensitivity-weighted penalty coefficient $\gamma_{l[k]}$.}
As in~\citet{fang2023depgraph}, our method also has this $\gamma_{l[k]}$ coefficient defined as
$\gamma_{l[k]} \triangleq 2^{\kappa(I^{\max}_{g(l)}-I_{l[k]})/(I^{\max}_{g(l)}-I^{\min}_{g(l)})},
$
where $I^{\max}_{g(l)}$ and $I^{\min}_{g(l)}$ are the maximum and minimum importance weights within group $g(l)$. In~\citet{fang2023depgraph}, they choose a fixed $\kappa=4$. In this work, we find it beneficial to use a scheduler for this term to gradually increase the importance. Specifically, we choose $\kappa = 4\cdot{t/T_s}$ across the $T_s$ steps of Stage~1. 

\myparagraph{Unlearning \& retain balance.}
As the unlearn set $\gU$ is typically much smaller than the retain set
$\gD \setminus \gU$, running the same number of gradient steps for each of them
leads to unbalanced updates. Following SFRon~\cite{huang2024unified}, we adopt a
forget-frequency schedule, \ie, running $f$ gradient steps on $\gL_{\tt R}$ for
every one step on $\gL_{\tt U}$. The values of $f$ are found by grid
search; we provide more details in an appendix.
Moreover, we employ a dynamic unlearning loss weight $\alpha_t = \alpha\,\zeta_t$ for classification tasks. Specifically, $\zeta_t = \mathrm{clip}\big((\ell_{\max}-\gL(\gU,\theta))/\ell_{\max},\,0,\,1\big)$, where $\ell_{\max}$ is the cross-entropy of a uniform predictor. Once $\gL(\gU,\theta) \ge \ell_{\max}$, $\gL_{\tt U}$ is permanently switched off. For generative models, $\gL_{\tt U}$ is turned off once the generation of the unlearned concepts deviates too far from a random Gaussian; we threshold based on MSE (equivalent to KL divergence up to constant scaling).

\begin{table*}[t]
    \setlength{\tabcolsep}{3pt}
    \renewcommand{\arraystretch}{1.15}
\centering
\caption{\textbf{CIFAR-10 (ResNet-18) random unlearning on classification averaged over 10 trials.} PruneForget achieves performance comparable to the \hl{oracle} (Retrain$\to$3SP). Baselines include SFRon~\cite{huang2024unified}, $\ell_1$-MU~\cite{jia2023model}, Un-pruning~\cite{xiao2025right}, Bilevel~\cite{shirkavand2025efficient}, and LLM-Eraser~\cite{zhang2025llm}. Performance gaps ($\Delta$) relative to the \hl{oracle} are reported first, with the raw UA, RA, and TA values provided in parentheses. \hlavg{$\text{Avg }\Delta$} reflects overall performance. Best results are in \textbf{\textcolor{colorBest}{blue}}, while second best are \textit{\textbf{italicized}}.}
\vspace{-0.2cm}
\resizebox{1.01\textwidth}{!}{
\begin{tabular}{l|ccc | >{\columncolor{avg_highlight}}c | cc | ccc | >{\columncolor{avg_highlight}}c | cc}
\specialrule{.15em}{.05em}{.05em}
\multirow{2}{*}{\textbf{Method}} & \multicolumn{6}{c|}{\textbf{Random Unlearn 10\%}} & \multicolumn{6}{c}{\textbf{Random Unlearn 50\%}} \\
 & \textbf{$\Delta$UA $\downarrow$} & \textbf{$\Delta$RA $\downarrow$} & \textbf{$\Delta$TA $\downarrow$} & \textbf{Avg $\Delta$ $\downarrow$} & \textbf{$D_{\text{KL}}$ $\downarrow$} & \textbf{Time} & \textbf{$\Delta$UA $\downarrow$} & \textbf{$\Delta$RA $\downarrow$} & \textbf{$\Delta$TA $\downarrow$} & \textbf{Avg $\Delta$ $\downarrow$} & \textbf{$D_{\text{KL}}$ $\downarrow$} & \textbf{Time} \\
\hline
\cellcolor{colorOracle}Retrain$\to$3SP & \cellcolor{colorOracle} 0.00 (6.03$_{\pm 0.56}$) & \cellcolor{colorOracle} 0.00 (100.00$_{\pm 0.00}$) & \cellcolor{colorOracle} 0.00 (93.74$_{\pm 0.32}$) & \cellcolor{colorOracle} 0.00 & \cellcolor{colorOracle} 0.00 & \cellcolor{colorOracle} 2215 & \cellcolor{colorOracle} 0.00 (7.60$_{\pm 0.22}$) & \cellcolor{colorOracle} 0.00 (100.00$_{\pm 0.00}$) & \cellcolor{colorOracle} 0.00 (92.20$_{\pm 0.30}$) & \cellcolor{colorOracle} 0.00 & \cellcolor{colorOracle} 0.00 & \cellcolor{colorOracle} 1518 \\
Pretrain$\to$3SP & 0.89 (5.15$_{\pm 0.38}$) & \textbf{\textcolor{colorBest}{0.00}} (100.00$_{\pm 0.00}$) & 0.41 (94.15$_{\pm 0.20}$) & 0.43 & \textbf{\textcolor{colorBest}{0.20}} & 984 & 1.54 (6.06$_{\pm 0.17}$) & \textbf{\textcolor{colorBest}{0.00}} (100.00$_{\pm 0.00}$) & 0.81 (93.01$_{\pm 0.27}$) & 0.78 & \textbf{\textcolor{colorBest}{0.23}} & 556 \\
3SP$\to$SFRon & 14.84 (20.88$_{\pm 0.77}$) & 1.88 (98.12$_{\pm 0.36}$) & 3.38 (90.37$_{\pm 0.37}$) & 6.70 & 0.24 & 1141 & 5.39 (12.99$_{\pm 0.54}$) & 0.74 (99.25$_{\pm 0.11}$) & 3.42 (88.79$_{\pm 0.56}$) & 3.19 & 0.27 & 593 \\
SFRon$\to$3SP & \textit{\textbf{0.16}} (5.87$_{\pm 0.57}$) & \textbf{\textcolor{colorBest}{0.00}} (100.00$_{\pm 0.01}$) & \textit{\textbf{0.04}} (93.70$_{\pm 0.31}$) & \textit{\textbf{0.07}} & \textit{\textbf{0.22}} & 1152 & \textbf{\textcolor{colorBest}{0.15}} (7.45$_{\pm 0.21}$) & \textbf{\textcolor{colorBest}{0.00}} (100.00$_{\pm 0.00}$) & \textit{\textbf{0.34}} (91.87$_{\pm 0.28}$) & \textit{\textbf{0.16}} & 0.27 & 627 \\
$\ell_1$-MU$\to$2SP & 5.00 (11.03$_{\pm 0.46}$) & 5.19 (94.81$_{\pm 0.26}$) & 4.90 (88.85$_{\pm 0.21}$) & 5.03 & 0.27 & 813 & 4.95 (12.55$_{\pm 0.20}$) & 3.43 (96.57$_{\pm 0.27}$) & 5.05 (87.16$_{\pm 0.16}$) & 4.48 & 0.34 & 478 \\
Un-pruning & 1.70 (7.74$_{\pm 0.25}$) & 0.05 (99.95$_{\pm 0.01}$) & 1.66 (92.08$_{\pm 0.19}$) & 1.14 & 0.41 & 937 & 2.72 (10.31$_{\pm 0.16}$) & 0.07 (99.93$_{\pm 0.02}$) & 2.94 (89.26$_{\pm 0.31}$) & 1.91 & 0.48 & 550 \\
Bilevel & 7.71 (13.74$_{\pm 0.76}$) & 7.01 (92.99$_{\pm 0.76}$) & 7.04 (86.71$_{\pm 0.69}$) & 7.25 & 0.55 & 942 & 6.61 (14.20$_{\pm 0.88}$) & 3.72 (96.28$_{\pm 0.76}$) & 6.73 (85.48$_{\pm 0.91}$) & 5.69 & 0.63 & 642 \\
LLM-Eraser & 12.60 (18.63$_{\pm 1.37}$) & 16.33 (83.67$_{\pm 1.24}$) & 12.51 (81.23$_{\pm 1.35}$) & 13.81 & 0.53 & 1482 & 10.46 (18.05$_{\pm 1.62}$) & 14.00 (85.99$_{\pm 1.60}$) & 10.59 (81.61$_{\pm 1.47}$) & 11.68 & 0.45 & 810 \\
\hline
PruneForget & \textbf{\textcolor{colorBest}{0.03}} (6.01$_{\pm 0.28}$) & \textbf{\textcolor{colorBest}{0.00}} (100.00$_{\pm 0.00}$) & \textbf{\textcolor{colorBest}{0.01}} (93.73$_{\pm 0.22}$) & \textbf{\textcolor{colorBest}{0.01}} & \textit{\textbf{0.22}} & 998 & \textit{\textbf{0.16}} (7.44$_{\pm 0.32}$) & \textbf{\textcolor{colorBest}{0.00}} (100.00$_{\pm 0.00}$) & \textbf{\textcolor{colorBest}{0.02}} (92.22$_{\pm 0.23}$) & \textbf{\textcolor{colorBest}{0.06}} & \textit{\textbf{0.25}} & 594 \\
\specialrule{.15em}{.05em}{.05em}
\end{tabular}
}
\label{tab:cls_cifar_rand_pr50}
\vspace{-0.2cm}
\end{table*}

\begin{table*}[t]
    \setlength{\tabcolsep}{3pt}
    \renewcommand{\arraystretch}{1.15}
\centering
\caption{\textbf{CIFAR-10 (ResNet-18) class-wise unlearning on classification averaged over 10 classes.} PruneForget achieves close performance to the \hl{oracle} and the lowest \hlavg{$\text{Avg }\Delta$} in both scenarios.}
\vspace{-0.2cm}
\resizebox{1.01\textwidth}{!}{
\begin{tabular}{l|ccc | >{\columncolor{avg_highlight}}c | cc | ccc | >{\columncolor{avg_highlight}}c | cc}
\specialrule{.15em}{.05em}{.05em}
\multirow{2}{*}{\textbf{Method}} & \multicolumn{6}{c|}{\textbf{Unlearn 1 Class (10\%)}} & \multicolumn{6}{c}{\textbf{Unlearn 5 Classes (50\%)}} \\
 & \textbf{$\Delta$UA $\downarrow$} & \textbf{$\Delta$RA $\downarrow$} & \textbf{$\Delta$TA $\downarrow$} & \textbf{Avg $\Delta$ $\downarrow$} & \textbf{$D_{\text{KL}}$ $\downarrow$} & \textbf{Time} & \textbf{$\Delta$UA $\downarrow$} & \textbf{$\Delta$RA $\downarrow$} & \textbf{$\Delta$TA $\downarrow$} & \textbf{Avg $\Delta$ $\downarrow$} & \textbf{$D_{\text{KL}}$ $\downarrow$} & \textbf{Time} \\
\hline
\cellcolor{colorOracle}Retrain$\to$3SP & \cellcolor{colorOracle} 0.00 (100.00$_{\pm 0.00}$) & \cellcolor{colorOracle} 0.00 (100.00$_{\pm 0.00}$) & \cellcolor{colorOracle} 0.00 (94.27$_{\pm 0.82}$) & \cellcolor{colorOracle} 0.00 & \cellcolor{colorOracle} 0.00 & \cellcolor{colorOracle} 2241 & \cellcolor{colorOracle} 0.00 (100.00$_{\pm 0.00}$) & \cellcolor{colorOracle} 0.00 (100.00$_{\pm 0.00}$) & \cellcolor{colorOracle} 0.00 (96.38$_{\pm 1.39}$) & \cellcolor{colorOracle} 0.00 & \cellcolor{colorOracle} 0.00 & \cellcolor{colorOracle} 1554 \\
Pretrain$\to$3SP & \textbf{\textcolor{colorBest}{0.00}} (100.00$_{\pm 0.00}$) & \textbf{\textcolor{colorBest}{0.00}} (100.00$_{\pm 0.00}$) & \textit{\textbf{0.44}} (94.72$_{\pm 0.57}$) & \textit{\textbf{0.15}} & \textbf{\textcolor{colorBest}{0.25}} & 989 & \textbf{\textcolor{colorBest}{0.00}} (100.00$_{\pm 0.00}$) & \textbf{\textcolor{colorBest}{0.00}} (100.00$_{\pm 0.00}$) & \textit{\textbf{0.22}} (96.60$_{\pm 1.23}$) & \textbf{\textcolor{colorBest}{0.07}} & \textbf{\textcolor{colorBest}{0.36}} & 566 \\
3SP$\to$SFRon & \textbf{\textcolor{colorBest}{0.00}} (100.00$_{\pm 0.00}$) & 2.77 (97.22$_{\pm 1.03}$) & 2.34 (91.94$_{\pm 1.31}$) & 1.70 & 0.46 & 1078 & \textbf{\textcolor{colorBest}{0.00}} (100.00$_{\pm 0.00}$) & 3.46 (96.54$_{\pm 2.84}$) & 3.53 (92.85$_{\pm 3.20}$) & 2.33 & 1.54 & 619 \\
SFRon$\to$3SP & \textbf{\textcolor{colorBest}{0.00}} (100.00$_{\pm 0.00}$) & \textit{\textbf{0.01}} (99.99$_{\pm 0.01}$) & 0.67 (93.60$_{\pm 0.68}$) & 0.23 & 0.33 & 1074 & \textbf{\textcolor{colorBest}{0.00}} (100.00$_{\pm 0.00}$) & \textit{\textbf{0.01}} (99.99$_{\pm 0.01}$) & 0.89 (95.48$_{\pm 2.00}$) & 0.30 & 0.47 & 615 \\
$\ell_1$-MU$\to$2SP & \textbf{\textcolor{colorBest}{0.00}} (100.00$_{\pm 0.00}$) & 4.93 (95.07$_{\pm 0.70}$) & 4.44 (89.84$_{\pm 0.78}$) & 3.12 & 0.31 & 805 & \textbf{\textcolor{colorBest}{0.00}} (100.00$_{\pm 0.00}$) & 1.74 (98.26$_{\pm 0.86}$) & 3.18 (93.20$_{\pm 1.94}$) & 1.64 & 0.45 & 453 \\
Un-pruning & \textbf{\textcolor{colorBest}{0.00}} (100.00$_{\pm 0.00}$) & 0.05 (99.95$_{\pm 0.02}$) & 1.53 (92.75$_{\pm 0.56}$) & 0.52 & 0.48 & 937 & \textbf{\textcolor{colorBest}{0.00}} (100.00$_{\pm 0.00}$) & 0.04 (99.96$_{\pm 0.03}$) & 1.60 (94.77$_{\pm 1.47}$) & 0.55 & 0.66 & 533 \\
Bilevel & \textbf{\textcolor{colorBest}{0.00}} (100.00$_{\pm 0.00}$) & 58.23 (41.77$_{\pm 14.96}$) & 52.53 (41.74$_{\pm 14.83}$) & 36.92 & 2.19 & 920 & \textbf{\textcolor{colorBest}{0.00}} (100.00$_{\pm 0.00}$) & 17.57 (82.43$_{\pm 11.71}$) & 15.82 (80.56$_{\pm 10.23}$) & 11.13 & 1.34 & 643 \\
LLM-Eraser & \textbf{\textcolor{colorBest}{0.00}} (100.00$_{\pm 0.00}$) & 21.50 (78.49$_{\pm 11.71}$) & 17.58 (76.69$_{\pm 11.08}$) & 13.03 & 0.78 & 1512 & \textbf{\textcolor{colorBest}{0.00}} (100.00$_{\pm 0.00}$) & 11.06 (88.94$_{\pm 3.39}$) & 9.36 (87.02$_{\pm 3.55}$) & 6.81 & 0.71 & 795 \\
\hline
PruneForget & \textbf{\textcolor{colorBest}{0.00}} (100.00$_{\pm 0.00}$) & \textbf{\textcolor{colorBest}{0.00}} (100.00$_{\pm 0.00}$) & \textbf{\textcolor{colorBest}{0.41}} (93.86$_{\pm 0.71}$) & \textbf{\textcolor{colorBest}{0.14}} & \textit{\textbf{0.28}} & 984 & \textbf{\textcolor{colorBest}{0.00}} (100.00$_{\pm 0.00}$) & \textbf{\textcolor{colorBest}{0.00}} (100.00$_{\pm 0.00}$) & \textbf{\textcolor{colorBest}{0.20}} (96.17$_{\pm 1.21}$) & \textbf{\textcolor{colorBest}{0.07}} & \textit{\textbf{0.38}} & 596 \\
\specialrule{.15em}{.05em}{.05em}
\end{tabular}
}
\label{tab:cls_cifar_class_pr50}
\vspace{-0.3cm}
\end{table*}

\section{Experiments}
To evaluate the effectiveness of PruneForget, we conduct experiments on both image classification and class-conditioned generation tasks. %The primary objective is to 
We assess the unlearning efficacy of models pruned to a \textbf{fixed target ratio, \ie, all the pruned models are equally efficient}. See the Appendix for implementation details, extended results, and global pruning experiments.
% \ray{mention global pruning.}

\subsection{Unlearning for classification} \label{sec:exp_cls}
\myparagraph{Setup.} Experiments incorporate aspects from both the unlearning and pruning literature. For the unlearning setup, we follow the setting of~\citet{huang2024unified} and evaluate ResNet-18~\cite{he2016deep} on CIFAR-10~\cite{krizhevsky2009learning} and pre-trained Swin-T~\cite{liu2021swin} on TinyImageNet~\cite{le2015tiny}. We report on both random unlearning (\ie, randomly removing a subset of the training data) and class-wise unlearning (\ie, removing all images of a specific class). For the pruning setup, we follow the setup of Group-level Pruning~\cite{fang2023depgraph} for ResNet-18 and Isomorphic Pruning~\cite{fang2024isomorphic} for Swin-T as the default structural model pruning mechanisms. We enforce a target pruning ratio $p$ of $0.5$ for ResNet-18 and $0.2$ for Swin-T, \ie, every baseline evaluated has exactly $100\cdot p\%$ channels per layer removed. 
Because the pruning methods may differ in the number of stages (either the last two or three described in \secref{sec:prelim}), we introduce the notation \textbf{$n$SP} ($n$-Stage Pruning) to denote the number of stages applied. For a fair comparison, PruneForget adopts the \textit{same} pruning mechanism (2SP or 3SP) as the baselines it is compared against. % in any given table.

\myparagraph{Evaluation metrics.}
To measure unlearning performance, prior works~\cite{fan2024salun,jia2023model} report the {\it gap} relative to the oracle model. In our setting, the oracle model (Retrain$\to$$n$SP) is created by first obtaining an unlearned model by training from scratch, followed by $n$-stage pruning. To quantify this gap, we report $\Delta \text{Acc}(\gS) \triangleq |\text{Acc}(\tilde{\theta},\gS) - \text{Acc}(\theta',\gS)|$, which measures the absolute accuracy difference between the unlearned model $\tilde{\theta}$ and the pruned oracle $\theta'$ over a dataset split $\gS$. We evaluate this gap on the \textbf{U}nlearn set $\gU$ ($\Delta$UA), the \textbf{R}etain set $\gD \setminus \gU$ ($\Delta$RA), and the \textbf{T}est set $\gT$ ($\Delta$TA). Note we report the error rate ($100 - \text{Acc}$) as the UA to show unlearning efficacy. 
To assess overall performance, we report the mean gap \hlavg{$\text{\textbf{Avg} }\Delta$} as our primary evaluation metric.
Additionally, we report the KL divergence ($D_{\text{KL}}$) to quantify the distributional difference between the model outputs and the oracle's, alongside the total execution {\bf Time} (in seconds) for the unlearn and prune procedure.

\begin{table*}[t]
    \setlength{\tabcolsep}{3pt}
    \renewcommand{\arraystretch}{1.15}
\centering
\caption{\textbf{TinyImageNet (Swin-T) random unlearning on classification averaged over 10 trials.} PruneForget achieves near-oracle performance across all metrics with the lowest \hlavg{$\text{Avg }\Delta$}. 
% Note that the retrained Swin-T is initialized from pretrained weights.
}
\vspace{-0.2cm}
\resizebox{1.01\textwidth}{!}{
\begin{tabular}{l|ccc | >{\columncolor{avg_highlight}}c | cc | ccc | >{\columncolor{avg_highlight}}c | cc}
\specialrule{.15em}{.05em}{.05em}
\multirow{2}{*}{\textbf{Method}} & \multicolumn{6}{c|}{\textbf{Random Unlearn 10\%}} & \multicolumn{6}{c}{\textbf{Random Unlearn 50\%}} \\
 & \textbf{$\Delta$UA $\downarrow$} & \textbf{$\Delta$RA $\downarrow$} & \textbf{$\Delta$TA $\downarrow$} & \textbf{Avg $\Delta$ $\downarrow$} & \textbf{$D_{\text{KL}}$ $\downarrow$} & \textbf{Time} & \textbf{$\Delta$UA $\downarrow$} & \textbf{$\Delta$RA $\downarrow$} & \textbf{$\Delta$TA $\downarrow$} & \textbf{Avg $\Delta$ $\downarrow$} & \textbf{$D_{\text{KL}}$ $\downarrow$} & \textbf{Time} \\
\hline
\cellcolor{colorOracle}Retrain$\to$2SP & \cellcolor{colorOracle} 0.00 (19.88$_{\pm 0.00}$) & \cellcolor{colorOracle} 0.00 (98.66$_{\pm 0.00}$) & \cellcolor{colorOracle} 0.00 (80.18$_{\pm 0.00}$) & \cellcolor{colorOracle} 0.00 & \cellcolor{colorOracle} 0.00 & \cellcolor{colorOracle} 8814 & \cellcolor{colorOracle} 0.00 (22.61$_{\pm 0.00}$) & \cellcolor{colorOracle} 0.00 (99.24$_{\pm 0.00}$) & \cellcolor{colorOracle} 0.00 (77.56$_{\pm 0.00}$) & \cellcolor{colorOracle} 0.00 & \cellcolor{colorOracle} 0.00 & \cellcolor{colorOracle} 5704 \\
Pretrain$\to$2SP & 1.64 (18.24$_{\pm 0.36}$) & 0.05 (98.71$_{\pm 0.03}$) & \textit{\textbf{0.09}} (80.27$_{\pm 0.21}$) & \textit{\textbf{0.59}} & \textbf{\textcolor{colorBest}{0.25}} & 6278 & \textit{\textbf{2.32}} (20.30$_{\pm 0.13}$) & \textbf{\textcolor{colorBest}{0.07}} (99.32$_{\pm 0.04}$) & \textit{\textbf{0.26}} (77.82$_{\pm 0.18}$) & \textit{\textbf{0.88}} & \textbf{\textcolor{colorBest}{0.30}} & 3771 \\
2SP$\to$SFRon & 7.26 (27.14$_{\pm 0.87}$) & 0.62 (98.04$_{\pm 0.18}$) & 4.83 (75.35$_{\pm 0.78}$) & 4.24 & 0.45 & 6968 & 4.61 (27.22$_{\pm 0.72}$) & 0.28 (98.96$_{\pm 0.08}$) & 5.64 (71.92$_{\pm 0.59}$) & 3.51 & 0.53 & 4779 \\
SFRon$\to$2SP & \textit{\textbf{1.63}} (18.25$_{\pm 0.31}$) & \textbf{\textcolor{colorBest}{0.03}} (98.69$_{\pm 0.04}$) & 0.20 (80.38$_{\pm 0.22}$) & 0.62 & \textit{\textbf{0.30}} & 7039 & 2.51 (20.11$_{\pm 0.22}$) & \textit{\textbf{0.09}} (99.34$_{\pm 0.05}$) & 0.44 (78.00$_{\pm 0.27}$) & 1.01 & \textit{\textbf{0.34}} & 4865 \\
$\ell_1$-MU$\to$2SP & 29.54 (49.42$_{\pm 2.31}$) & 39.17 (59.49$_{\pm 2.60}$) & 29.36 (50.82$_{\pm 2.10}$) & 32.69 & 1.36 & 7412 & 33.71 (56.33$_{\pm 0.48}$) & 45.53 (53.71$_{\pm 0.78}$) & 34.04 (43.51$_{\pm 0.50}$) & 37.76 & 1.55 & 4486 \\
Un-pruning & 5.66 (25.54$_{\pm 0.61}$) & 2.85 (95.81$_{\pm 0.32}$) & 6.30 (73.88$_{\pm 0.52}$) & 4.94 & 0.51 & 10040 & 5.92 (28.53$_{\pm 0.42}$) & 1.87 (97.37$_{\pm 0.32}$) & 6.88 (70.67$_{\pm 0.62}$) & 4.89 & 0.58 & 5694 \\
Bilevel & 13.00 (6.87$_{\pm 0.24}$) & 2.52 (96.14$_{\pm 0.04}$) & 2.99 (83.17$_{\pm 0.20}$) & 6.17 & 1.21 & 18672 & 16.29 (6.33$_{\pm 0.07}$) & 2.11 (97.13$_{\pm 0.09}$) & 5.36 (82.92$_{\pm 0.11}$) & 7.92 & 1.47 & 18785 \\
LLM-Eraser & 6.40 (13.47$_{\pm 0.28}$) & 3.05 (95.61$_{\pm 0.18}$) & 2.13 (82.31$_{\pm 0.25}$) & 3.86 & 0.35 & 7804 & 8.35 (14.27$_{\pm 0.19}$) & 2.76 (96.49$_{\pm 0.10}$) & 3.40 (80.96$_{\pm 0.17}$) & 4.83 & 0.41 & 4241 \\
\hline
PruneForget & \textbf{\textcolor{colorBest}{1.50}} (18.38$_{\pm 0.24}$) & \textbf{\textcolor{colorBest}{0.03}} (98.69$_{\pm 0.03}$) & \textbf{\textcolor{colorBest}{0.07}} (80.25$_{\pm 0.17}$) & \textbf{\textcolor{colorBest}{0.54}} & \textit{\textbf{0.30}} & 7026 & \textbf{\textcolor{colorBest}{0.98}} (21.64$_{\pm 0.12}$) & 0.10 (99.35$_{\pm 0.03}$) & \textbf{\textcolor{colorBest}{0.21}} (77.35$_{\pm 0.27}$) & \textbf{\textcolor{colorBest}{0.43}} & 0.36 & 4912 \\
\specialrule{.15em}{.05em}{.05em}
\end{tabular}
}
\label{tab:cls_swin_rand_pr20}
\vspace{-0.5cm}
\end{table*}

\begin{table}[t]
\vspace{-0.05cm}
    \small
    \setlength{\tabcolsep}{3pt}
    \centering
    \caption{\textbf{Ablation of pipeline stages.} Evaluating the contribution of each optimization stage in PruneForget across 50\% random unlearning on CIFAR-10. Integrating all three stages yields the optimal balance between unlearning efficacy and model utility.}
\vspace{-0.2cm}
    {
    \begin{tabular}{ccc|ccc | >{\columncolor{avg_highlight}}c | cc}
        \specialrule{.15em}{.05em}{.05em}
        \textbf{Stage} & \textbf{Stage} & \textbf{Stage} & \multicolumn{6}{c}{\textbf{Random Unlearn 50\%}} \\
        \textbf{1} & \textbf{2} & \textbf{3} &  \textbf{$\Delta$UA $\downarrow$} & \textbf{$\Delta$RA $\downarrow$} & \textbf{$\Delta$TA $\downarrow$} & \textbf{Avg $\Delta$ $\downarrow$} & \textbf{$D_{\text{KL}}$ $\downarrow$} & \textbf{Time} \\
        \hline
         & & & 1.54 (6.06$_{\pm 0.17}$) & \textbf{\textcolor{colorBest}{0.00}} (100.00$_{\pm 0.00}$) & 0.81 (93.01$_{\pm 0.27}$) & 0.78 & \textbf{\textcolor{colorBest}{0.23}} & 556 \\
        \checkmark & & & 0.93 (6.67$_{\pm 0.17}$) & \textbf{\textcolor{colorBest}{0.00}} (100.00$_{\pm 0.00}$) & 0.29 (92.50$_{\pm 0.24}$) & 0.41 & 0.25 & 588 \\
        \checkmark & \checkmark & & \textit{\textbf{0.80}} (6.80$_{\pm 0.16}$) & \textbf{\textcolor{colorBest}{0.00}} (100.00$_{\pm 0.00}$) & \textit{\textbf{0.22}} (92.43$_{\pm 0.20}$) & \textit{\textbf{0.34}} & \textit{\textbf{0.24}} & 590 \\
        \checkmark & \checkmark & \checkmark & \textbf{\textcolor{colorBest}{0.16}} (7.44$_{\pm 0.32}$) & \textbf{\textcolor{colorBest}{0.00}} (100.00$_{\pm 0.00}$) & \textbf{\textcolor{colorBest}{0.02}} (92.22$_{\pm 0.23}$) & \textbf{\textcolor{colorBest}{0.06}} & 0.25 & 594 \\
        \specialrule{.15em}{.05em}{.05em}
    \end{tabular}
    }
\vspace{-.5cm}
    \label{tab:ablation}
\end{table}

\myparagraph{Baselines.} We compare our method against several baselines: \textbf{Pretrain$\to$$n$SP}, which prunes the pre-trained model without unlearning; \textbf{$n$SP$\to$SFRon}~\cite{huang2024unified}, which applies $n$ pruning stages followed by the SFRon unlearning algorithm; \textbf{SFRon}~\cite{huang2024unified}\textbf{$\to$$n$SP}, which unlearns using SFRon followed by the pruning stages; \textbf{$\ell_1$-MU}~\cite{jia2023model}\textbf{$\to$2SP}, utilizing sparsity-inducing weights for unlearning followed by the last two pruning stages; \textbf{Un-pruning}~\cite{xiao2025right}, which iteratively interleaves unlearning and partial weight reinitialization before a final pruning step; \textbf{Bilevel}~\cite{shirkavand2025efficient}, integrating unlearning into the post-pruning fine-tuning stage via bilevel optimization; and \textbf{LLM-Eraser}~\cite{zhang2025llm}, which incorporates unlearning into the pruning and fine-tuning stages and applies them iteratively.
% We compare with \textbf{(1) (Pretrain $\to$) Reprune} prunes the pretrained model without explicit unlearning, \textbf{(2) Reprune$\to$SFRon}~\cite{huang2024unified} and \textbf{(3) SFRon$\to$Reprune}~\cite{huang2024unified} naively sequences the two processes in both orders, \textbf{(4) L1-sparseMU}~\cite{jia2023model} utilizes sparsity-inducing weights to achieve unlearning, \textbf{(5) Un-pruning}~\cite{xiao2025right} and \textbf{(6) LLM-Eraser}~\cite{zhang2025llm} interleave unlearning and pruning steps iteratively, and \textbf{(7) Bilevel}~\cite{shirkavand2025efficient} integrates unlearning into the post-pruning fine-tuning phase. We apply the same pruning ratio to all baselines.

\myparagraph{Results on \textit{random} unlearning on CIFAR-10 (ResNet-18).}
We report the 10\% and 50\% random unlearning results in~\tabref{tab:cls_cifar_rand_pr50}. 
Without an explicit unlearning objective, the Pretrain$\to$3SP baseline fails to erase the influence of the data, resulting in a UA that is 1.54\% lower than the oracle model (6.06\% \textit{vs.} 7.60\%). Conversely, while other prior works achieve some unlearning effectiveness, they struggle to maintain low $\Delta$TA in both cases. In contrast, PruneForget consistently achieves the lowest $\text{Avg }\Delta$ under both unlearning ratios. Overall, our method shows robust unlearning with a zero $\Delta$RA alongside low $\Delta$UA, $\Delta$TA, and $D_{\text{KL}}$, while consuming less than 50\% of the oracle's runtime.\vspace{5pt}\\
\myparagraph{Results on \textit{class-wise} unlearning on CIFAR-10 (ResNet-18).}
In~\tabref{tab:cls_cifar_class_pr50}, we evaluate class-wise unlearning on 1-class and 5-class scenarios. PruneForget achieves the lowest $\text{Avg }\Delta$ and zero gap in both UA and RA, with a near-zero gap in TA and a low $D_{\text{KL}}$, outperforming the compared baselines. This result is expected, as class-wise unlearning is typically the easier task vs. random unlearning. 
%These results show that PruneForget effectively balances precise forgetting with robust performance recovery.

\myparagraph{Results on \textit{random} unlearning on TinyImageNet (Swin-T).}
We report the 10\% and 50\% random unlearning results in \tabref{tab:cls_swin_rand_pr20}. 
The Pretrain$\to$2SP baseline again suffers from a significant unlearning deficit ($\Delta$UA = 2.32 on 50\% unlearning). While other baselines also struggle to balance UA and TA, PruneForget consistently achieves the lowest $\text{Avg }\Delta$. Notably, our method yields the smallest $\Delta$UA and $\Delta$TA in both cases while maintaining a robust performance across all metrics. This result demonstrates PruneForget's applicability to different datasets and model architectures.
%
%
%
% \ray{@Yijia.}
% Because early attention modules capture local spatial features~\cite{raghu2021vision}, they likely encode forget-specific visual features. Simultaneously, the terminal attention layers integrate global semantic representations for final decision-making~\cite{liu2021swin}, which are equally critical for unlearning. Aggressively pruning these boundary layers thus effectively removes the targeted information from the model. Conversely, PruneForget preserves greater capacity in the final MLP modules. Since MLPs encode high-level semantics~\cite{geva2021transformer}, safeguarding them prevents catastrophic forgetting of the retain data. This strategic redistribution demonstrates that PruneForget successfully embeds the unlearning objective directly into its pruning pattern.
%\ray{TODO: something about QKV is important.}
%
%

\myparagraph{Ablation on the three stages of PruneForget.} We conduct an ablation study on the CIFAR-10 50\% random unlearning task to demonstrate the contribution of each stage in PruneForget. When a specific stage is ablated, we revert to the default pruning procedure. In \tabref{tab:ablation}, we observe that all stages meaningfully contribute to the overall unlearning efficacy. Specifically, progressively incorporating Stage~1 and Stage~2 reduces $\Delta\text{UA}$ from 1.54\% to 0.93\% and 0.80\% and $\Delta\text{TA}$ from 0.81\% to 0.29\% and 0.22\%. Integrating all three stages further achieves the best trade-off, lowering $\Delta\text{UA}$ to 0.16\% and $\Delta\text{TA}$ to a near-oracle 0.02\%, yielding the lowest overall {$\text{Avg }\Delta$} of 0.06\%. Overall, these results show the value of integrating unlearning awareness into every stage of the pruning pipeline.

% omitting Stage 1 increases the $\Delta$UA from 0.54\% to 0.85\%, while removing Stage 2 and Stage 3 increases it to 0.58\% and 0.62\%, respectively. Furthermore, in terms of model utility, removing Stage 1 and Stage 2 worsens the $\Delta$TA from 0.03\% to 0.37\% and 0.09\%. 

\begin{table}[t]
\small
    \setlength{\tabcolsep}{3pt}
    \centering
    \caption{\textbf{Sensitivity score design comparison.} We evaluate the impact of various designs of $\mathcal{S}$ on 50\% random unlearning on CIFAR-10. Ours achieves the lowest \hlavg{$\text{Avg }\Delta$} among all methods.}
\vspace{-0.25cm}
    {
    \begin{tabular}{l|ccc | >{\columncolor{avg_highlight}}c | cc}
        \specialrule{.15em}{.05em}{.05em}
        \multirow{2}{*}{\textbf{Design of} $\mathcal{S}$} & \multicolumn{6}{c}{\textbf{Random Unlearn 50\%}} \\
        &  \textbf{$\Delta$UA $\downarrow$} & \textbf{$\Delta$RA $\downarrow$} & \textbf{$\Delta$TA $\downarrow$} & \textbf{Avg $\Delta$ $\downarrow$} & \textbf{$D_{\text{KL}}$ $\downarrow$} & \textbf{Time} \\
        \hline
        $1$ (Uniform) & 1.49 (6.11$_{\pm 0.33}$) & \textbf{\textcolor{colorBest}{0.00}} (100.00$_{\pm 0.00}$) & 0.71 (92.91$_{\pm 0.21}$) & 0.73 & \textbf{\textcolor{colorBest}{0.23}} & 575 \\
        $\text{Random}$ & 2.29 (5.31$_{\pm 0.45}$) & \textbf{\textcolor{colorBest}{0.00}} (100.00$_{\pm 0.00}$) & 0.98 (93.19$_{\pm 0.19}$) & 1.09 & \textbf{\textcolor{colorBest}{0.23}} & 574 \\
        Unlearn Only  & \textbf{\textcolor{colorBest}{0.15}} (7.44$_{\pm 0.15}$) & \textbf{\textcolor{colorBest}{0.00}} (100.00$_{\pm 0.00}$) & \textit{\textbf{0.05}} (92.25$_{\pm 0.22}$) & \textit{\textbf{0.07}} & 0.26 & 587 \\
        Retain Only & 1.05 (8.64$_{\pm 0.43}$) & \textbf{\textcolor{colorBest}{0.00}} (100.00$_{\pm 0.00}$) & 1.09 (91.11$_{\pm 0.50}$) & 0.71 & 0.27 & 589 \\
        Fisher Score & \textbf{\textcolor{colorBest}{0.15}} (7.74$_{\pm 0.58}$) & \textbf{\textcolor{colorBest}{0.00}} (100.00$_{\pm 0.00}$) & 0.15 (92.05$_{\pm 0.36}$) & 0.10 & \textit{\textbf{0.25}} & 620 \\
        \textbf{Ours} & \textit{\textbf{0.16}} (7.44$_{\pm 0.32}$) & \textbf{\textcolor{colorBest}{0.00}} (100.00$_{\pm 0.00}$) & \textbf{\textcolor{colorBest}{0.02}} (92.22$_{\pm 0.23}$) & \textbf{\textcolor{colorBest}{0.06}} & \textit{\textbf{0.25}} & 594 \\ 
        \specialrule{.15em}{.05em}{.05em}
    \end{tabular}
    }
    \label{tab:ablation_S}
    \vspace{-0.3cm}
\end{table}
% \myparagraph{Sensitivity score design comparison.}
% In~\tabref{tab:ablation_S}, we evaluate the impact of using different data subsets in computing the sensitivity score~\equref{eq:sensitivity} on the CIFAR-10 50\% random unlearning task. Specifically, we compare using only the retain set ($\gD \setminus \gU$), only the unlearn set ($\gU$), or a combination of both.  Using only the retain set achieves the tightest unlearning gap ($\Delta$UA of 0.18) but degrades generalization ($\Delta$TA of 0.20) due to overfitting. Conversely, using only the unlearn set results in a $\Delta$UA of 0.34 and 
% hurts the test accuracy ($\Delta$TA of 0.09). Hence, using the entire training dataset $\gD$ gives the best overall performance. The 10\% unlearning results are provided in the Appendix, where we can draw the same conclusion.
%Combining both sets establishes a self-normalizing mechanism that contrasts gradients to protect shared, generalizable features while penalizing forget-specific weights. This optimal balance best preserves test accuracy ($\Delta$TA of 0.03) while maintaining effective unlearning ($\Delta$UA of 0.54). 
%.

\myparagraph{Sensitivity score design comparison.}
In~\tabref{tab:ablation_S}, we evaluate the impact of various formulations for the sensitivity score $\mathcal{S}$ on the CIFAR-10 50\% random unlearning task. Naive baselines, such as uniform or random assignments, lead to substantial performance gaps ({$\text{Avg }\Delta$} $\ge 0.73\%$). Relying solely on the retain set impairs forgetting efficacy ($\Delta\text{UA} = 1.05\%$). While the unlearn-only subset and the Fisher score improve forgetting, they still result in suboptimal test retention ($\Delta\text{TA} \ge 0.05\%$). In contrast, our proposed formulation achieves the lowest overall {$\text{Avg }\Delta$} alongside a near-oracle $\Delta\text{TA}$ of 0.02\%, which is the best trade-off across all other designs.

\begin{table}[t]
\vspace{-0.05cm}
    \setlength{\tabcolsep}{2.5pt}
\centering
\small
    \caption{\textbf{Comparison of updating vs. static sensitivity score.} We compare an updating sensitivity score versus a static one. The static approach achieves a smaller gap to the \hl{oracle} across all metrics.}
\vspace{-0.25cm}
{
\begin{tabular}{l|ccc | >{\columncolor{avg_highlight}}c | cc}
\specialrule{.15em}{.05em}{.05em}
\multirow{2}{*}{\textbf{Method}} & \multicolumn{6}{c}{\textbf{Random Unlearn 50\%}} \\
 & \textbf{$\Delta$UA $\downarrow$} & \textbf{$\Delta$RA $\downarrow$} & \textbf{$\Delta$TA $\downarrow$} & \textbf{Avg $\Delta$ $\downarrow$} & \textbf{$D_{\text{KL}}$ $\downarrow$} & \textbf{Time (s)} \\
\hline
\cellcolor{colorOracle}Retrain$\to$3SP & \cellcolor{colorOracle} 0.00 (7.60$_{\pm 0.22}$) & \cellcolor{colorOracle} 0.00 (100.00$_{\pm 0.00}$) & \cellcolor{colorOracle} 0.00 (92.20$_{\pm 0.30}$) & \cellcolor{colorOracle} 0.00 & \cellcolor{colorOracle} 0.00 & \cellcolor{colorOracle} 1518 \\
Updating Sensitivity & 0.41 (8.00$_{\pm 0.49}$) & \textbf{\textcolor{colorBest}{0.00}} (100.00$_{\pm 0.00}$) & 0.51 (91.70$_{\pm 0.35}$) & 0.30 & 0.26 & 604 \\
Static (PruneForget) & \textbf{\textcolor{colorBest}{0.16}} (7.44$_{\pm 0.32}$) & \textbf{\textcolor{colorBest}{0.00}} (100.00$_{\pm 0.00}$) & \textbf{\textcolor{colorBest}{0.02}} (92.22$_{\pm 0.23}$) & \textbf{\textcolor{colorBest}{0.06}} & \textbf{\textcolor{colorBest}{0.25}} & 594 \\
\specialrule{.15em}{.05em}{.05em}
\end{tabular}
}
\label{tab:dynamic_S}
\vspace{-.6cm}
\end{table}
\myparagraph{Ablation on iteratively updating the sensitivity score.}
In Alg.~\ref{alg:pruneforget}, the sensitivity score $S_{l[k]}$ is computed once on the pre-trained model. We now conduct an ablation to determine whether updating this score iteratively during Stage~1 would be beneficial. In \tabref{tab:dynamic_S}, we use the setting of CIFAR-10 50\% random unlearning for this ablation. We use a running mean across Stage 1 to iteratively update $S_{l[k]}$. Overall, the static approach (PruneForget) achieves better {$\text{Avg }\Delta$}, indicating that analyzing the entire dataset beforehand is sufficient to determine the contribution of weights to both the unlearned and the retained information. This design choice is consistent with SFRon~\cite{huang2024unified}, which computes its gradient mask once on the pre-trained model and reuses it throughout training.

%Furthermore, by eliminating the overhead of continuous gradient updates, the static method achieves significantly higher computational efficiency.
%We also present experiments on the 10\% random unlearning scenario in the appendix, where the static score even yields superior results across all evaluated metrics.

\begin{table}[t]
\vspace{-0.05cm}
    \small
    \setlength{\tabcolsep}{3pt}
    \centering
\caption{\textbf{Ablation of dynamic weighted unlearning.} We evaluate dynamic weighted unlearning on 50\% random unlearning of CIFAR-10, demonstrating that it effectively improves \hlavg{$\text{Avg }\Delta$}.}
\vspace{-0.2cm}
{
\begin{tabular}{l|ccc | >{\columncolor{avg_highlight}}c | cc}
\specialrule{.15em}{.05em}{.05em}
\multirow{2}{*}{\textbf{Method}} & \multicolumn{6}{c}{\textbf{Random Unlearn 50\%}} \\
 & \textbf{$\Delta$UA $\downarrow$} & \textbf{$\Delta$RA $\downarrow$} & \textbf{$\Delta$TA $\downarrow$} & \textbf{Avg $\Delta$ $\downarrow$} & \textbf{$D_{\text{KL}}$ $\downarrow$} & \textbf{Time} \\
\hline
\cellcolor{colorOracle}Retrain$\to$3SP & \cellcolor{colorOracle} 0.00 (7.60$_{\pm 0.22}$) & \cellcolor{colorOracle} 0.00 (100.00$_{\pm 0.00}$) & \cellcolor{colorOracle} 0.00 (92.20$_{\pm 0.30}$) & \cellcolor{colorOracle} 0.00 & \cellcolor{colorOracle} 0.00 & \cellcolor{colorOracle} 1518 \\
$\zeta_t=1$ & 0.20 (7.80$_{\pm 0.33}$) & \textbf{\textcolor{colorBest}{0.00}} (100.00$_{\pm 0.00}$) & 0.17 (92.03$_{\pm 0.22}$) & 0.12 & 0.26 & 612 \\
Dynamic $\zeta_t$ & \textbf{\textcolor{colorBest}{0.16}} (7.44$_{\pm 0.32}$) & \textbf{\textcolor{colorBest}{0.00}} (100.00$_{\pm 0.00}$) & \textbf{\textcolor{colorBest}{0.02}} (92.22$_{\pm 0.23}$) & \textbf{\textcolor{colorBest}{0.06}} & \textbf{\textcolor{colorBest}{0.25}} & 594 \\
\specialrule{.15em}{.05em}{.05em}
\end{tabular}
}
\vspace{-0.1cm}
\label{tab:ablation_ga}
\end{table}

\myparagraph{Ablation on dynamic weighted unlearning.}
In~\tabref{tab:ablation_ga}, we evaluate dynamic weighted unlearning against a fixed baseline on CIFAR-10 with 50\% random unlearning. Compared to the fixed variant, our dynamic scheme effectively reduces both $\Delta\text{UA}$ and $\Delta\text{TA}$, lowering {$\text{Avg }\Delta$} from 0.12\% to 0.06\%. That is, the proposed dynamic weighting improves the trade-off between unlearn and retain sets.

\begin{table}[t]
\vspace{-0.1cm}
    \setlength{\tabcolsep}{3pt}
\centering
\small
\caption{\textbf{Ablation on iterative pruning rounds.} We evaluate the impact of the number of iterative pruning rounds ($R$). A single round ($R=1$) achieves the closest oracle alignment.}
\vspace{-0.2cm}
{
\begin{tabular}{l|ccc | >{\columncolor{avg_highlight}}c | cc}
\specialrule{.15em}{.05em}{.05em}
\multirow{2}{*}{\textbf{Method}} & \multicolumn{6}{c}{\textbf{Random Unlearn 50\%}} \\
 & \textbf{$\Delta$UA $\downarrow$} & \textbf{$\Delta$RA $\downarrow$} & \textbf{$\Delta$TA $\downarrow$} & \textbf{Avg $\Delta$ $\downarrow$} & \textbf{$D_{\text{KL}}$ $\downarrow$} & \textbf{Time (s)} \\
\hline
\cellcolor{colorOracle}Retrain$\to$3SP & \cellcolor{colorOracle} 0.00 (7.60$_{\pm 0.22}$) & \cellcolor{colorOracle} 0.00 (100.00$_{\pm 0.00}$) & \cellcolor{colorOracle} 0.00 (92.20$_{\pm 0.30}$) & \cellcolor{colorOracle} 0.00 & \cellcolor{colorOracle} 0.00 & \cellcolor{colorOracle} 1518 \\
$R=1$ (Ours) & \textbf{\textcolor{colorBest}{0.16}} (7.44$_{\pm 0.32}$) & \textbf{\textcolor{colorBest}{0.00}} (100.00$_{\pm 0.00}$) & \textbf{\textcolor{colorBest}{0.02}} (92.22$_{\pm 0.23}$) & \textbf{\textcolor{colorBest}{0.06}} & \textbf{\textcolor{colorBest}{0.25}} & 594 \\
$R=2$ & 0.57 (7.02$_{\pm 0.14}$) & \textbf{\textcolor{colorBest}{0.00}} (100.00$_{\pm 0.00}$) & 0.49 (92.69$_{\pm 0.22}$) & 0.35 & \textbf{\textcolor{colorBest}{0.25}} & 599 \\
$R=5$ & 0.67 (6.93$_{\pm 0.14}$) & \textbf{\textcolor{colorBest}{0.00}} (100.00$_{\pm 0.00}$) & 0.57 (92.77$_{\pm 0.16}$) & 0.41 & \textbf{\textcolor{colorBest}{0.25}} & 646 \\
$R=10$ & 0.45 (7.15$_{\pm 0.18}$) & \textbf{\textcolor{colorBest}{0.00}} (100.00$_{\pm 0.00}$) & 0.48 (92.69$_{\pm 0.14}$) & 0.31 & \textbf{\textcolor{colorBest}{0.25}} & 702 \\
\specialrule{.15em}{.05em}{.05em}
\end{tabular}
}
\label{tab:cls_cifar_iter}
\vspace{-1cm}
\end{table}

\myparagraph{Ablation on iterative pruning rounds.}
In~\tabref{tab:cls_cifar_iter}, we evaluate iterative pruning rounds ($R$) by alternating between Stage~1 and Stage~2. To match the compute, we divided the total iterations by a factor of $R$
 for each iteration of Stage 1 and 2, and iterated for $R$
 times. Increasing $R$ yields no performance gains; thus, we maintain single-round pruning as our default.

\subsection{Unlearning for class-conditioned generation}

\myparagraph{Setup \& evaluation metrics.} For unlearning, we follow~\citet{huang2024unified} using DDPMs~\cite{ddpm} on CIFAR-10 and DiT~\cite{peebles2023scalable} on ImageNet-1K~\cite{deng2009imagenet}. As for pruning, we enforce $p=0.3$ via Diff-Pruning~\cite{fang2023structural} and compare against the same baselines detailed in~\secref{sec:exp_cls}. Note that DiT results are deferred to Appendix~\secref{sec:supp_additional_exp_gene}. %\vspace{5pt}
%
%\myparagraph{Evaluation metrics.} 
We evaluate performance by reporting the gap relative to the oracle model (Retrain$\to$$n$SP) using three metrics: \textbf{UA}, the unlearned accuracy of a pre-trained classifier on images generated for the unlearn classes; \textbf{FID}, the Fréchet Inception Distance~\cite{heusel2017gans} assessing image generation quality on the retain classes; and \textbf{Time}, the total execution time in seconds.

\myparagraph{Results on class-wise unlearning on CIFAR-10.\vspace{-10pt}\\}%
\begin{wraptable}[11]{r}{0.46\textwidth}
    \setlength{\tabcolsep}{3pt}
\centering
\small
\vspace{-26pt} % Adjust this to move the table up/down
\caption{\textbf{CIFAR-10 (DDPM) class-wise unlearning results.} Averaged across 5 classes, PruneForget obtains close performance to the \hl{oracle} with over 10$\times$ acceleration.}
\vspace{-.2cm}
\label{tab:ddpm_cifar_avg}
\resizebox{1.01\linewidth}{!}{
\begin{tabular}{l|cc|c}
\specialrule{.15em}{.05em}{.05em}
    \multirow{2}{*}{\textbf{Method}} & \multicolumn{2}{c|}{\textbf{Average (5 Classes)}} & \multirow{2}{*}{\textbf{Time}} \\
 & \textbf{$\Delta$UA $\downarrow$} & \textbf{$\Delta$FID $\downarrow$} & \\
\hline
\cellcolor{colorOracle}Retrain$\to$2SP & \cellcolor{colorOracle} 0.00 (98.39$_{\pm 1.65}$) & \cellcolor{colorOracle} 0.00 (16.73$_{\pm 0.63}$) & \cellcolor{colorOracle} 398622 \\
Pretrain$\to$2SP & 0.45 (97.94$_{\pm 1.70}$) & 0.30 (16.95$_{\pm 0.47}$) & 21304 \\
2SP$\to$SFRon & 0.39 (98.00$_{\pm 1.52}$) & \textbf{\textcolor{colorBest}{0.22}} (16.92$_{\pm 0.70}$) & 25652 \\
SFRon$\to$2SP & 0.55 (97.83$_{\pm 1.77}$) & \textit{\textbf{0.23}} (16.96$_{\pm 0.71}$) & 26674 \\
$\ell_1$-MU$\to$2SP & \textit{\textbf{0.20}} (98.19$_{\pm 1.76}$) & 0.76 (17.49$_{\pm 0.73}$) & 22675 \\
Un-pruning & 0.75 (98.89$_{\pm 2.23}$) & 326.24 (342.96$_{\pm 31.06}$) &  36586 \\
Bilevel & 4.30 (94.09$_{\pm 2.68}$) & 0.98 (17.71$_{\pm 0.40}$) & 33863 \\
% LLM-Eraser & 3.66 (94.73$_{\pm 5.92}$) & 62.01 (78.74$_{\pm 17.65}$) & 1835 \\
LLM-Eraser & 3.08 (95.31$_{\pm 4.50}$) & 7.41 (24.13$_{\pm 2.52}$) & 27570 \\
\hline
PruneForget & \textbf{\textcolor{colorBest}{0.13}} (98.26$_{\pm 1.64}$) & 0.26 (16.92$_{\pm 0.76}$) & 26254 \\
\specialrule{.15em}{.05em}{.05em}
\end{tabular}
}
\end{wraptable}% 
We show the average class-wise unlearning results of DDPM over 5 classes (\textit{Automobile}, \textit{Cat}, \textit{Dog}, \textit{Horse}, and \textit{Truck}) in~\tabref{tab:ddpm_cifar_avg}. We observe PruneForget effectively matches the oracle model across all cases. It yields the lowest $\Delta$UA alongside highly competitive $\Delta$FID scores, demonstrating its ability to unlearn targeted concepts while maintaining overall image quality effectively. Importantly, PruneForget achieves these near-zero performance gaps while accelerating runtime by over 10$\times$ relative to the oracle model. Qualitative results are provided in Appendix~\figref{fig:supp_qualitative_gen}, and we observe them to be consistent with the quantitative metrics.

% \myparagraph{Qualitative results.} 
% In Appendix~\figref{fig:supp_qualitative_gen} we present the images generated from the unlearned models on the unlearn classes: \textit{Automobile}, \textit{Cat}, \textit{Dog}, \textit{Horse}, and \textit{Truck}. 
% We observe that the pre-trained model generates recognizable images for all categories, while the oracle (Retrain$\rightarrow$2SP) and PruneForget successfully erase the forgotten concepts. Specifically, for the unlearn classes (columns I1–I5), the outputs from PruneForget lose their original class semantics, the images become ambiguous or another class in the retain classes, which closely mirrors the behavior of the oracle. Conversely, for the retain classes (columns C1–C9), PruneForget preserves high visual fidelity, producing images that are very similar to those generated by the oracle model. These visual results confirm that PruneForget effectively unlearns the class information.

\section{Conclusion}
We presented PruneForget, a method that jointly performs machine unlearning and model pruning by using the unlearn set to guide which model parameters to remove. Intuitively, if a weight encodes information that should be unlearned, then it is a good pruning candidate. To realize this intuition, we proposed a sensitivity score that measures the ratio between the unlearned information and the information in the retain set. We then incorporated this score into the three-stage pruning pipeline, which consists of sparse training, pruning, and fine-tuning. By doing so, {PruneForget} produces an unlearned model that is pruned to satisfy a compute budget. Experiments across image classification/generation tasks, different architectures, and pruning mechanisms demonstrate that PruneForget is effective and outperforms existing baselines that jointly unlearn and prune models.
\clearpage
\subsection*{AI use statement}
% \clover{todo}
% (This section is \textbf{required} and does not count toward the page limit.)

% In this work, we used generative AI tools for [tasks with required disclosure].
% We have not used generative AI tools for [other tasks with required disclosure],
% and [the rest of the required disclosure tasks] are not applicable to this work.
% Additionally, we used generative AI tools for [tasks with recommended
% disclosure]. We have reviewed all AI-assisted work. [Elaborate. For example, “we
% checked LLM-generated research ideas for potential plagiarism through a manual
% literature survey”, “LLM-generated code was verified and tested for correctness
% by 2 authors”, etc.]. We take responsibility for the final content of this work,
% including text, claims or artifacts produced with the aid of generative AI.

%(This section is \textbf{required} and does not count toward the page limit.)

In this work, we used generative AI tools for polishing writing, searching on related works, and adapting prior works' code for experiments. We have not used generative AI tools for drafting sections of the paper, generating synthetic datasets, or proving mathematical claims. We have reviewed all AI-assisted work. For example, any related work suggested by the AI was checked manually. Furthermore, AI-assisted code adaptations from prior works used in our experiments were manually verified and tested for correctness within our pipeline. We take responsibility for the final content of this work, including text, claims, or artifacts produced with the aid of generative AI.
%\clearpage
\bibliography{main}

\begin{thebibliography}{72}
\providecommand{\natexlab}[1]{#1}
\providecommand{\url}[1]{\texttt{#1}}
\expandafter\ifx\csname urlstyle\endcsname\relax
  \providecommand{\doi}[1]{doi: #1}\else
  \providecommand{\doi}{doi: \begingroup \urlstyle{rm}\Url}\fi

\bibitem[Aravindan et~al.(2025)Aravindan, Jha, Salaway, Bhide, and Yaldiz]{aravindan2025sealing}
Ashwath~Vaithinathan Aravindan, Abha Jha, Matthew Salaway, Atharva~Sandeep Bhide, and Duygu~Nur Yaldiz.
\newblock Sealing the backdoor: Unlearning adversarial text triggers in diffusion models using knowledge distillation.
\newblock \emph{arXiv preprint arXiv:2508.18235}, 2025.

\bibitem[Becker \& Liebig(2022)Becker and Liebig]{becker2022evaluating}
Alexander Becker and Thomas Liebig.
\newblock Evaluating machine unlearning via epistemic uncertainty.
\newblock \emph{arXiv preprint arXiv:2208.10836}, 2022.

\bibitem[Bonta(2022)]{bonta2022california}
Rob Bonta.
\newblock California consumer privacy act ({CCPA}).
\newblock \emph{Retrieved from State of California Department of Justice: \url{https://oag. ca. gov/privacy/ccpa}}, 2022.

\bibitem[Cao \& Yang(2015)Cao and Yang]{cao2015towards}
Yinzhi Cao and Junfeng Yang.
\newblock Towards making systems forget with machine unlearning.
\newblock In \emph{Proc. IEEE S\&P}, 2015.

\bibitem[Castells et~al.(2024)Castells, Song, Kim, and Choi]{castells2024ld}
Thibault Castells, Hyoung-Kyu Song, Bo-Kyeong Kim, and Shinkook Choi.
\newblock {{LD-Pruner}}: Efficient pruning of latent diffusion models using task-agnostic insights.
\newblock In \emph{Proc. CVPR}, 2024.

\bibitem[Chundawat et~al.(2023)Chundawat, Tarun, Mandal, and Kankanhalli]{chundawat2023can}
Vikram~S Chundawat, Ayush~K Tarun, Murari Mandal, and Mohan Kankanhalli.
\newblock Can bad teaching induce forgetting? unlearning in deep networks using an incompetent teacher.
\newblock In \emph{Proc. AAAI}, 2023.

\bibitem[Deng et~al.(2009)Deng, Dong, Socher, Li, Li, and Fei-Fei]{deng2009imagenet}
Jia Deng, Wei Dong, Richard Socher, Li-Jia Li, Kai Li, and Li~Fei-Fei.
\newblock Image{N}et: A large-scale hierarchical image database.
\newblock In \emph{Proc. CVPR}, 2009.

\bibitem[Ding et~al.(2019{\natexlab{a}})Ding, Ding, Guo, and Han]{ding2019centripetal}
Xiaohan Ding, Guiguang Ding, Yuchen Guo, and Jungong Han.
\newblock Centripetal {{SGD}} for pruning very deep convolutional networks with complicated structure.
\newblock In \emph{Proc. CVPR}, 2019{\natexlab{a}}.

\bibitem[Ding et~al.(2019{\natexlab{b}})Ding, Ding, Guo, Han, and Yan]{ding2019approximated}
Xiaohan Ding, Guiguang Ding, Yuchen Guo, Jungong Han, and Chenggang Yan.
\newblock Approximated oracle filter pruning for destructive {CNN} width optimization.
\newblock In \emph{Proc. ICML}, 2019{\natexlab{b}}.

\bibitem[Dong et~al.(2017)Dong, Chen, and Pan]{dong2017learning}
Xin Dong, Shangyu Chen, and Sinno Pan.
\newblock Learning to prune deep neural networks via layer-wise optimal brain surgeon.
\newblock In \emph{Proc. NeurIPS}, 2017.

\bibitem[{European Union}(2016)]{EU2016GDPR}
{European Union}.
\newblock Regulation {(EU)} 2016/679 of the european parliament and of the council of 27 {April} 2016 on the protection of natural persons with regard to the processing of personal data and on the free movement of such data (general data protection regulation), 2016.
\newblock URL \url{https://eur-lex.europa.eu/eli/reg/2016/679/oj}.
\newblock OJ L 119, 4.5.2016, p. 1–88.

\bibitem[Fan et~al.(2024)Fan, Liu, Zhang, Wong, Wei, and Liu]{fan2024salun}
Chongyu Fan, Jiancheng Liu, Yihua Zhang, Eric Wong, Dennis Wei, and Sijia Liu.
\newblock {{SalUn}}: Empowering machine unlearning via gradient-based weight saliency in both image classification and generation.
\newblock In \emph{Proc. ICLR}, 2024.

\bibitem[Fang et~al.(2023{\natexlab{a}})Fang, Ma, Song, Mi, and Wang]{fang2023depgraph}
Gongfan Fang, Xinyin Ma, Mingli Song, Michael~Bi Mi, and Xinchao Wang.
\newblock {{DepGraph}}: Towards any structural pruning.
\newblock In \emph{Proc. CVPR}, 2023{\natexlab{a}}.

\bibitem[Fang et~al.(2023{\natexlab{b}})Fang, Ma, and Wang]{fang2023structural}
Gongfan Fang, Xinyin Ma, and Xinchao Wang.
\newblock Structural pruning for diffusion models.
\newblock In \emph{Proc. NeurIPS}, 2023{\natexlab{b}}.

\bibitem[Fang et~al.(2024)Fang, Ma, Mi, and Wang]{fang2024isomorphic}
Gongfan Fang, Xinyin Ma, Michael~Bi Mi, and Xinchao Wang.
\newblock Isomorphic pruning for vision models.
\newblock In \emph{Proc. ECCV}, 2024.

\bibitem[Foster et~al.(2024)Foster, Schoepf, and Brintrup]{SSD_2024}
Jack Foster, Stefan Schoepf, and Alexandra Brintrup.
\newblock Fast machine unlearning without retraining through selective synaptic dampening.
\newblock In \emph{Proc. AAAI}, 2024.

\bibitem[Frankle \& Carbin(2019)Frankle and Carbin]{frankle2018lottery}
Jonathan Frankle and Michael Carbin.
\newblock The lottery ticket hypothesis: Finding sparse, trainable neural networks.
\newblock In \emph{Proc. ICLR}, 2019.

\bibitem[Gandikota et~al.(2023)Gandikota, Materzynska, Fiotto-Kaufman, and Bau]{gandikota2023erasing}
Rohit Gandikota, Joanna Materzynska, Jaden Fiotto-Kaufman, and David Bau.
\newblock Erasing concepts from diffusion models.
\newblock In \emph{Proc. ICCV}, 2023.

\bibitem[Ganjdanesh et~al.(2024)Ganjdanesh, Shirkavand, Gao, and Huang]{ganjdanesh2024not}
Alireza Ganjdanesh, Reza Shirkavand, Shangqian Gao, and Heng Huang.
\newblock Not all prompts are made equal: Prompt-based pruning of text-to-image diffusion models.
\newblock \emph{arXiv preprint arXiv:2406.12042}, 2024.

\bibitem[Gao et~al.(2021)Gao, Huang, Cai, and Huang]{gao2021network}
Shangqian Gao, Feihu Huang, Weidong Cai, and Heng Huang.
\newblock Network pruning via performance maximization.
\newblock In \emph{Proc. CVPR}, 2021.

\bibitem[Golatkar et~al.(2020)Golatkar, Achille, and Soatto]{golatkar2020eternal}
Aditya Golatkar, Alessandro Achille, and Stefano Soatto.
\newblock Eternal sunshine of the spotless net: Selective forgetting in deep networks.
\newblock In \emph{Proc. CVPR}, 2020.

\bibitem[Graves et~al.(2021)Graves, Nagisetty, and Ganesh]{graves2021amnesiac}
Laura Graves, Vineel Nagisetty, and Vijay Ganesh.
\newblock Amnesiac machine learning.
\newblock In \emph{Proc. AAAI}, 2021.

\bibitem[Guo et~al.(2020)Guo, Ouyang, and Xu]{guo2020multi}
Jinyang Guo, Wanli Ouyang, and Dong Xu.
\newblock Multi-dimensional pruning: A unified framework for model compression.
\newblock In \emph{Proc. CVPR}, 2020.

\bibitem[Han et~al.(2015)Han, Mao, and Dally]{han2015deep}
Song Han, Huizi Mao, and William~J Dally.
\newblock Deep compression: Compressing deep neural networks with pruning, trained quantization and huffman coding.
\newblock \emph{arXiv preprint arXiv:1510.00149}, 2015.

\bibitem[He et~al.(2016)He, Zhang, Ren, and Sun]{he2016deep}
Kaiming He, Xiangyu Zhang, Shaoqing Ren, and Jian Sun.
\newblock Deep residual learning for image recognition.
\newblock In \emph{Proc. CVPR}, 2016.

\bibitem[He et~al.(2019)He, Liu, Wang, Hu, and Yang]{he2019filter}
Yang He, Ping Liu, Ziwei Wang, Zhilan Hu, and Yi~Yang.
\newblock Filter pruning via geometric median for deep convolutional neural networks acceleration.
\newblock In \emph{Proc. CVPR}, 2019.

\bibitem[Heng \& Soh(2023)Heng and Soh]{heng2023selective}
Alvin Heng and Harold Soh.
\newblock Selective amnesia: A continual learning approach to forgetting in deep generative models.
\newblock In \emph{Proc. NeurIPS}, 2023.

\bibitem[Heusel et~al.(2017)Heusel, Ramsauer, Unterthiner, Nessler, and Hochreiter]{heusel2017gans}
Martin Heusel, Hubert Ramsauer, Thomas Unterthiner, Bernhard Nessler, and Sepp Hochreiter.
\newblock {{GANs}} trained by a two time-scale update rule converge to a local {{Nash}} equilibrium.
\newblock In \emph{Proc. NeurIPS}, 2017.

\bibitem[Ho et~al.(2020)Ho, Jain, and Abbeel]{ddpm}
Jonathan Ho, Ajay Jain, and Pieter Abbeel.
\newblock Denoising diffusion probabilistic models.
\newblock In \emph{Proc. NeurIPS}, 2020.

\bibitem[Huang et~al.(2024)Huang, Cheng, Zheng, Wang, He, Li, and Huang]{huang2024unified}
Zhehao Huang, Xinwen Cheng, JingHao Zheng, Haoran Wang, Zhengbao He, Tao Li, and Xiaolin Huang.
\newblock Unified gradient-based machine unlearning with remain geometry enhancement.
\newblock In \emph{Proc. NeurIPS}, 2024.

\bibitem[Izzo et~al.(2021)Izzo, Smart, Chaudhuri, and Zou]{izzo2021approximate}
Zachary Izzo, Mary~Anne Smart, Kamalika Chaudhuri, and James Zou.
\newblock Approximate data deletion from machine learning models.
\newblock In \emph{Proc. AISTATS}, 2021.

\bibitem[Jia et~al.(2023)Jia, Liu, Ram, Yao, Liu, Liu, Sharma, and Liu]{jia2023model}
Jinghan Jia, Jiancheng Liu, Parikshit Ram, Yuguang Yao, Gaowen Liu, Yang Liu, Pranay Sharma, and Sijia Liu.
\newblock Model sparsity can simplify machine unlearning.
\newblock In \emph{Proc. NeurIPS}, 2023.

\bibitem[Kang et~al.(2026)Kang, Kim, Kim, Kim, Lee, and Jung]{kang2026porta}
Minseok Kang, Hyunwoo Kim, Chanyoung Kim, Minwoo Kim, Jaekoo Lee, and Dahuin Jung.
\newblock Prune once: Retraining-free task-agnostic pruning for vision-language models.
\newblock In \emph{Proc. ECCV}, 2026.

\bibitem[Koh \& Liang(2017)Koh and Liang]{koh2017understanding}
Pang~Wei Koh and Percy Liang.
\newblock Understanding black-box predictions via influence functions.
\newblock In \emph{Proc. ICML}, 2017.

\bibitem[Krizhevsky et~al.(2009)Krizhevsky, Hinton, et~al.]{krizhevsky2009learning}
Alex Krizhevsky, Geoffrey Hinton, et~al.
\newblock Learning multiple layers of features from tiny images.
\newblock 2009.

\bibitem[Kurmanji et~al.(2023)Kurmanji, Triantafillou, Hayes, and Triantafillou]{kurmanji2023towards}
Meghdad Kurmanji, Peter Triantafillou, Jamie Hayes, and Eleni Triantafillou.
\newblock Towards unbounded machine unlearning.
\newblock In \emph{Proc. NeurIPS}, 2023.

\bibitem[Le et~al.(2015)Le, Yang, et~al.]{le2015tiny}
Yann Le, Xuan Yang, et~al.
\newblock Tiny {ImageNet} visual recognition challenge.
\newblock \emph{CS 231N}, 2015.

\bibitem[Lee et~al.(2019)Lee, Ajanthan, Gould, and Torr]{lee2019signal}
Namhoon Lee, Thalaiyasingam Ajanthan, Stephen Gould, and Philip~HS Torr.
\newblock A signal propagation perspective for pruning neural networks at initialization.
\newblock \emph{arXiv preprint arXiv:1906.06307}, 2019.

\bibitem[Lin et~al.(2020)Lin, Ji, Wang, Zhang, Zhang, Tian, and Shao]{lin2020hrank}
Mingbao Lin, Rongrong Ji, Yan Wang, Yichen Zhang, Baochang Zhang, Yonghong Tian, and Ling Shao.
\newblock {{HRank}}: Filter pruning using high-rank feature map.
\newblock In \emph{Proc. CVPR}, 2020.

\bibitem[Liu et~al.(2021{\natexlab{a}})Liu, Zhang, Kuang, Zhou, Xue, Wang, Chen, Yang, Liao, and Zhang]{liu2021group}
Liyang Liu, Shilong Zhang, Zhanghui Kuang, Aojun Zhou, Jing-Hao Xue, Xinjiang Wang, Yimin Chen, Wenming Yang, Qingmin Liao, and Wayne Zhang.
\newblock Group fisher pruning for practical network compression.
\newblock In \emph{Proc. ICML}, 2021{\natexlab{a}}.

\bibitem[Liu et~al.(2021{\natexlab{b}})Liu, Lin, Cao, Hu, Wei, Zhang, Lin, and Guo]{liu2021swin}
Ze~Liu, Yutong Lin, Yue Cao, Han Hu, Yixuan Wei, Zheng Zhang, Stephen Lin, and Baining Guo.
\newblock {{Swin}} transformer: Hierarchical vision transformer using shifted windows.
\newblock In \emph{Proc. ICCV}, 2021{\natexlab{b}}.

\bibitem[Liu et~al.(2017)Liu, Li, Shen, Huang, Yan, and Zhang]{liu2017learning}
Zhuang Liu, Jianguo Li, Zhiqiang Shen, Gao Huang, Shoumeng Yan, and Changshui Zhang.
\newblock Learning efficient convolutional networks through network slimming.
\newblock In \emph{Proc. ICCV}, 2017.

\bibitem[Lu et~al.(2026)Lu, Shi, Wang, and Zhang]{lu2026machine}
Juxin Lu, Haoyu Shi, Mengyao Wang, and Huaiwen Zhang.
\newblock Machine unlearning via adaptive gradient reweighting and multi-stage objective optimization.
\newblock In \emph{Proc. CVPR}, 2026.

\bibitem[Mao et~al.(2021)Mao, Yang, Li, Li, and Chen]{mao2021tprune}
Jiachen Mao, Huanrui Yang, Ang Li, Hai Li, and Yiran Chen.
\newblock {{TPrune}}: Efficient transformer pruning for mobile devices.
\newblock \emph{ACM Transactions on Cyber-Physical Systems}, 2021.

\bibitem[Neel et~al.(2021)Neel, Roth, and Sharifi-Malvajerdi]{neel2021descent}
Seth Neel, Aaron Roth, and Saeed Sharifi-Malvajerdi.
\newblock Descent-to-delete: Gradient-based methods for machine unlearning.
\newblock In \emph{Proc. ALT}, 2021.

\bibitem[Orseau et~al.(2020)Orseau, Hutter, and Rivasplata]{orseau2020logarithmic}
Laurent Orseau, Marcus Hutter, and Omar Rivasplata.
\newblock Logarithmic pruning is all you need.
\newblock In \emph{Proc. NeurIPS}, 2020.

\bibitem[Park et~al.(2020)Park, Lee, Mo, and Shin]{park2020lookahead}
Sejun Park, Jaeho Lee, Sangwoo Mo, and Jinwoo Shin.
\newblock Lookahead: A far-sighted alternative of magnitude-based pruning.
\newblock \emph{arXiv preprint arXiv:2002.04809}, 2020.

\bibitem[Peebles \& Xie(2023)Peebles and Xie]{peebles2023scalable}
William Peebles and Saining Xie.
\newblock Scalable diffusion models with transformers.
\newblock In \emph{Proc. CVPR}, 2023.

\bibitem[Sanh et~al.(2020)Sanh, Wolf, and Rush]{sanh2020movement}
Victor Sanh, Thomas Wolf, and Alexander Rush.
\newblock Movement pruning: Adaptive sparsity by fine-tuning.
\newblock In \emph{Proc. NeurIPS}, 2020.

\bibitem[Schindler et~al.(2020)Schindler, Roth, Pernkopf, and Fr{\"o}ning]{schindler2020parameterized}
G{\"u}nther Schindler, Wolfgang Roth, Franz Pernkopf, and Holger Fr{\"o}ning.
\newblock Parameterized structured pruning for deep neural networks.
\newblock In \emph{Proc. LOD}, 2020.

\bibitem[Seo et~al.(2025)Seo, Kim, and Han]{seo2025revisiting}
Seonguk Seo, Dongwan Kim, and Bohyung Han.
\newblock Revisiting machine unlearning with dimensional alignment.
\newblock In \emph{Proc. WACV}. IEEE, 2025.

\bibitem[Shirkavand et~al.(2025)Shirkavand, Yu, Gao, Somepalli, Goldstein, and Huang]{shirkavand2025efficient}
Reza Shirkavand, Peiran Yu, Shangqian Gao, Gowthami Somepalli, Tom Goldstein, and Heng Huang.
\newblock Efficient fine-tuning and concept suppression for pruned diffusion models.
\newblock In \emph{Proc. CVPR}, 2025.

\bibitem[Song \& Mittal(2021)Song and Mittal]{song2021systematic}
Liwei Song and Prateek Mittal.
\newblock Systematic evaluation of privacy risks of machine learning models.
\newblock In \emph{USENIX security symposium}, 2021.

\bibitem[Song et~al.(2022)Song, Xu, He, Jiang, Jing, and Liang]{song2022cp}
Zhuoran Song, Yihong Xu, Zhezhi He, Li~Jiang, Naifeng Jing, and Xiaoyao Liang.
\newblock {CP-ViT}: Cascade vision transformer pruning via progressive sparsity prediction.
\newblock \emph{arXiv preprint arXiv:2203.04570}, 2022.

\bibitem[Thudi et~al.(2022)Thudi, Deza, Chandrasekaran, and Papernot]{thudi2022unrolling}
Anvith Thudi, Gabriel Deza, Varun Chandrasekaran, and Nicolas Papernot.
\newblock Unrolling {{SGD}}: Understanding factors influencing machine unlearning.
\newblock In \emph{Proc. EuroS\&P}, 2022.

\bibitem[Warnecke et~al.(2023)Warnecke, Pirch, Wressnegger, and Rieck]{warnecke2023machine}
Alexander Warnecke, Lukas Pirch, Christian Wressnegger, and Konrad Rieck.
\newblock Machine unlearning of features and labels.
\newblock In \emph{Network and Distributed System Security Symposium (NDSS)}, 2023.

\bibitem[Wu et~al.(2025)Wu, Le, Hayat, and Harandi]{wu2025erasing}
Jing Wu, Trung Le, Munawar Hayat, and Mehrtash Harandi.
\newblock Erasing undesirable influence in diffusion models.
\newblock In \emph{Proc. CVPR}, 2025.

\bibitem[Wu et~al.(2026)Wu, Wang, Wang, and Chen]{wu2026collaborative}
Zimeng Wu, Yunhong Wang, Donghao Wang, and Jiaxin Chen.
\newblock Collaborative multi-mode pruning for vision-language models.
\newblock In \emph{Proc. CVPR}, 2026.

\bibitem[Xiao et~al.(2025)Xiao, Li, Ji, Ye, Ma, and Hui]{xiao2025right}
Yang Xiao, Gen Li, Jie Ji, Ruimeng Ye, Xiaolong Ma, and Bo~Hui.
\newblock The right to be forgotten in pruning: Unveil machine unlearning on sparse models.
\newblock \emph{arXiv preprint arXiv:2507.18725}, 2025.

\bibitem[Ye et~al.(2018)Ye, Lu, Lin, and Wang]{ye2018rethinking}
Jianbo Ye, Xin Lu, Zhe Lin, and James~Z Wang.
\newblock Rethinking the smaller-norm-less-informative assumption in channel pruning of convolution layers.
\newblock \emph{arXiv preprint arXiv:1802.00124}, 2018.

\bibitem[Yin et~al.(2026)Yin, Li, Chen, Chen, Yao, and Wang]{yin2026cut}
Jianjian Yin, Liulei Li, Tao Chen, Yi~Chen, Yazhou Yao, and Wenguan Wang.
\newblock Cut-vit: Task-specific model pruning via gram anchoring subspace consistency.
\newblock In \emph{Proc. ECCV}, 2026.

\bibitem[You et~al.(2019)You, Yan, Ye, Ma, and Wang]{you2019gate}
Zhonghui You, Kun Yan, Jinmian Ye, Meng Ma, and Ping Wang.
\newblock Gate decorator: Global filter pruning method for accelerating deep convolutional neural networks.
\newblock In \emph{Proc. NeurIPS}, 2019.

\bibitem[Yu et~al.(2022)Yu, Huang, Wang, Cheng, Chu, and Cui]{yu2022width}
Fang Yu, Kun Huang, Meng Wang, Yuan Cheng, Wei Chu, and Li~Cui.
\newblock Width \& depth pruning for vision transformers.
\newblock In \emph{Proc. AAAI}, 2022.

\bibitem[Yu et~al.(2018)Yu, Li, Chen, Lai, Morariu, Han, Gao, Lin, and Davis]{yu2018nisp}
Ruichi Yu, Ang Li, Chun-Fu Chen, Jui-Hsin Lai, Vlad~I Morariu, Xintong Han, Mingfei Gao, Ching-Yung Lin, and Larry~S Davis.
\newblock {{NISP}}: Pruning networks using neuron importance score propagation.
\newblock In \emph{Proc. CVPR}, 2018.

\bibitem[Yuan et~al.(2025)Yuan, Pang, Du, Chen, Zhang, and Lin]{yuan2025closer}
Xiaojian Yuan, Tianyu Pang, Chao Du, Kejiang Chen, Weiming Zhang, and Min Lin.
\newblock A closer look at machine unlearning for large language models.
\newblock In \emph{Proc. ICLR}, 2025.

\bibitem[Zhang et~al.(2024{\natexlab{a}})Zhang, Wang, Xu, Wang, and Shi]{zhang2024forget}
Gong Zhang, Kai Wang, Xingqian Xu, Zhangyang Wang, and Humphrey Shi.
\newblock Forget-me-not: Learning to forget in text-to-image diffusion models.
\newblock In \emph{Proc. CVPR}, 2024{\natexlab{a}}.

\bibitem[Zhang et~al.(2024{\natexlab{b}})Zhang, Lin, Bai, and Mei]{zhang2024negative}
Ruiqi Zhang, Licong Lin, Yu~Bai, and Song Mei.
\newblock Negative preference optimization: From catastrophic collapse to effective unlearning.
\newblock In \emph{Proc. COLM}, 2024{\natexlab{b}}.

\bibitem[Zhang et~al.(2025)Zhang, Zhang, Zhou, Zheng, and Xiong]{zhang2025llm}
Shengming Zhang, Le~Zhang, Jingbo Zhou, Zhi Zheng, and Hui Xiong.
\newblock {LLM-Eraser}: Optimizing large language model unlearning through selective pruning.
\newblock In \emph{Proc. KDD}, 2025.

\bibitem[Zhang et~al.(2021)Zhang, Wang, Qin, and Fu]{zhang2021aligned}
Yulun Zhang, Huan Wang, Can Qin, and Yun Fu.
\newblock Aligned structured sparsity learning for efficient image super-resolution.
\newblock In \emph{Proc. NeurIPS}, 2021.

\bibitem[Zheng et~al.(2026)Zheng, Tai, and Yeh]{zheng2026designing}
Amber~Yijia Zheng, Yu-Shan Tai, and Raymond~A Yeh.
\newblock Designing to forget: Deep semi-parametric models for unlearning.
\newblock In \emph{Proc. CVPR}, 2026.

\bibitem[Zhu et~al.(2021)Zhu, Tang, and Han]{zhu2021vision}
Mingjian Zhu, Yehui Tang, and Kai Han.
\newblock Vision transformer pruning.
\newblock \emph{arXiv preprint arXiv:2104.08500}, 2021.

\bibitem[Zhuang et~al.(2020)Zhuang, Zhang, Huang, Zeng, Shuang, and Li]{zhuang2020neuron}
Tao Zhuang, Zhixuan Zhang, Yuheng Huang, Xiaoyi Zeng, Kai Shuang, and Xiang Li.
\newblock Neuron-level structured pruning using polarization regularizer.
\newblock In \emph{Proc. NeurIPS}, 2020.

\end{thebibliography}
\bibliographystyle{iclr2027_conference}

\appendix
\clearpage
\appendix
\label{sec:appendix}

{\noindent \bf \Large Appendix}

\setcounter{section}{0}
\renewcommand{\theHsection}{A\arabic{section}}
\renewcommand{\thesection}{A\arabic{section}}
\renewcommand{\thetable}{A\arabic{table}}
\setcounter{table}{0}
\setcounter{figure}{0}
\renewcommand{\thetable}{A\arabic{table}}
\renewcommand\thefigure{A\arabic{figure}}
\renewcommand{\theHtable}{A.Tab.\arabic{table}}%<---!!!!---
\renewcommand{\theHfigure}{A.Abb.\arabic{figure}}%<---!!!!---
\renewcommand\theequation{A\arabic{equation}}
\renewcommand{\theHequation}{A.Abb.\arabic{equation}}%<---!!!!---

\vspace{6pt}
{\bf\noindent The appendix is organized as follows:}
\begin{itemize}[topsep=0pt, leftmargin=12pt]
    \item The background comparison is in~\secref{sec:supp_icomp_table}.
    \item The implementation and experimental details are presented in~\secref{sec:supp_implemet_detail}.
    \item Additional experiment results for \textbf{classification} are reported in~\secref{sec:supp_additional_exp_class}.
    \item Additional experiment results for \textbf{class-conditioned generation} are reported in~\secref{sec:supp_additional_exp_gene}.
    \item Additional experiment results for \textbf{global pruning} are reported in~\secref{sec:supp_global}.
    % \item Limitations of this work are in~\secref{sec:supp_limitation}.
    % \item Broader impacts of this work are in~\secref{sec:supp_impact}.
    
\end{itemize}

\section{Background comparison}
\label{sec:supp_icomp_table}
% \ray{TODO: Compare the background in Tab. 1}
\begin{table}[h]
    \centering
    \small
    \caption{\textbf{Comparison with related works.} Typical pruning methods consist of three stages of sparsity training, pruning, and post-pruning fine-tuning. Prior works have mainly focused on making the fine-tuning stage unlearn-aware.
    % \ray{@Yushan: Let's move this to appendix.}
    }
    % \vspace{-0.2cm}
 {
    \begin{tabular}{lccc}
        \specialrule{.15em}{.05em}{.05em}
        \multirow{2}{*}{\textbf{Related Work}} & \multicolumn{3}{c}{\textbf{Unlearn-aware Stages}} \\
        \cmidrule(lr){2-4}
        & \textbf{Sparsity} & \textbf{Pruning} & \textbf{Fine-tuning} \\
        \hline
        %(Pretrain $\rightarrow$) Reprune & & & \\
        %\hline
        %Reprune$\to$SFRon~\cite{huang2024unified} & & & \checkmark \\
        %\hline
        %SFRon$\to$Reprune~\cite{huang2024unified} & & & \checkmark \\
        %\hline
        $\ell_1$-MU & & & \checkmark\\
        %\hline
        Un-pruning & & & \checkmark \\
        Bilevel & & & \checkmark \\
        LLM-Eraser & & \checkmark & \checkmark \\
        \textbf{PruneForget (Ours)} & \checkmark & \checkmark & \checkmark \\
        \specialrule{.15em}{.05em}{.05em}
    \end{tabular}
    }
    \label{tab:method_comparison}
\end{table}
 As summarized in~\tabref{tab:method_comparison}, prior approaches only partially integrate unlearning into the three stages of pruning (sparse training, pruning, post-pruning fine-tuning). In contrast, our proposed PruneForget makes all three stages unlearn-aware to further improve the unlearning performance of pruned models.

% \ray{TODO: qualitative images in appendix.}
% \input{figs/ddpm_cifar}

\begin{table}[t]
\centering
\small
\caption{\textbf{Model compute reduction after pruning.} We report MACs, parameter counts, and theoretical speedup across pruned models (with different pruning ratios $p$). Here, we follow~\citet{fang2023depgraph} and report the theoretical speedup $\triangleq \frac{\text{MACs}(\text{base})}{\text{MACs}(\text{pruned})}$.}
\label{tab:macs}
\begin{tabular}{llcc|ccc}
\specialrule{.15em}{.05em}{.05em}
\textbf{Model} & \textbf{Dataset} & \textbf{Input} & \textbf{$p$} & \textbf{MACs} & \textbf{Params} & \textbf{Speedup} \\
\hline
\multirow{4}{*}{ResNet-18} & \multirow{4}{*}{\shortstack{CIFAR-10 / \\ CIFAR-100}} & \multirow{4}{*}{$32\times 32$} 
  & 0  & 556.6M  & 11.17M  & $1.00\times$ \\
& & & 0.3 & 269.4M & 5.46M & $2.07\times$ \\
& & & 0.5 & 139.9M & 2.80M  & $3.98\times$ \\
& & & 0.7 & 49.8M  & 1.00M & $11.17\times$ \\
\hline
\multirow{2}{*}{Swin-T} & \multirow{2}{*}{TinyImageNet} & \multirow{2}{*}{$224\times 224$} 
  & 0  & 4.36G  & 27.67M & $1.00\times$ \\
& & & 0.2 & 3.56G  & 20.19M & $1.22\times$ \\
\hline
\multirow{2}{*}{DDPM} & \multirow{2}{*}{CIFAR-10} & \multirow{2}{*}{$32\times 32$} 
  & 0  & 6.06G  & 36.08M & $1.00\times$ \\
& & & 0.3 & 2.07G  & 14.05M & $2.93\times$ \\
\hline
\multirow{2}{*}{DiT-XL/2} & \multirow{2}{*}{ImageNet} & \multirow{2}{*}{$256\times 256$} 
  & 0  & 118.74G & 675.13M & $1.00\times$ \\
& & & 0.3 & 51.81G  & 309.46M & $2.29\times$ \\
\specialrule{.15em}{.05em}{.05em}
\end{tabular}
\end{table}
\section{Implementation and Experiment Details}
\label{sec:supp_implemet_detail} 
 We now provide the implementation details for the pruning methods, PruneForget, and baselines. For any unspecified training hyperparameters, we apply the default settings of the respective pruning methods. We adopt the pruning ratios specified in the original methods and apply comparable values for models not evaluated in the original studies. 
 
 As we apply uniform pruning, where an equal proportion of channels is removed across all layers, all baselines share identical pruned architectures and achieve the same operational speedup. Parameter counts and MACs for each pruned configuration are detailed in \tabref{tab:macs}. We follow the official implementation of SFRon~\cite{huang2024unified} to obtain the pre-trained classifiers utilized to evaluate Unlearned Accuracy (UA) in generative tasks. Consistent with their evaluation protocol, we synthesize 1,000 images per class to compute both the FID and UA metrics for DDPM on CIFAR-10. For DiT on ImageNet-1K, we sample 500 images for the unlearn class and randomly sample 10,000 images from the retain classes. We omit Membership Inference Attack (MIA)~\cite{song2021systematic} evaluation, as recent studies~\cite{seo2025revisiting, xiao2025right, zheng2026designing} demonstrate it is an unreliable metric for assessing data removal. The runtime is measured on a single NVIDIA L40S GPU.

\myparagraph{Details for pruning methods.}
\label{sec:supp_implemet_detail_prune} 
Our method works with various pruning methods. For completeness, we document the details for these pruning methods.

\begin{itemize}[topsep=0pt, leftmargin=12pt]
    \item \textbf{Group-level Pruning~\cite{fang2023depgraph}:} This baseline constructs parameter groups $\gG=\{g\}$, assigning layers $l$ into groups $g$ based on their dependencies in the computational graph. The method employs a standard three-stage pipeline consisting of sparse training, structural pruning, and post-pruning fine-tuning. Group regularization is applied using the weighted $L_2$-norm of the parameter groups, which simultaneously serves as the criterion for the importance weights. Following their implementation, both $T_s$ and $T_f$ are set to 100 epochs. The model is optimized using stochastic gradient descent (SGD) with a learning rate of 0.01, a momentum of 0.9, and no weight decay, decayed via a cosine annealing scheduler.

    \item \textbf{Isomorphic Pruning~\cite{fang2024isomorphic}:} This baseline groups sub-structures into various isomorphic groups and performs importance ranking within these groups. The method employs a two-stage pipeline consisting of structural pruning and post-pruning fine-tuning, utilizing a Taylor-based pruning framework. We set $T_f$ to 50 epochs. Following their implementation, the model is optimized using AdamW with a learning rate of $5 \times 10^{-4}$ and a weight decay of 0.05, decayed via a cosine annealing scheduler.
    
    \item \textbf{Diff-Pruning~\cite{fang2023structural}:} This is a structural pruning method specialized for diffusion models. It employs a two-stage pipeline consisting of structural pruning followed by post-pruning fine-tuning. To calculate importance weights, the method accumulates first-order gradients over critical denoising steps, specifically focusing on larger timesteps. $T_f$ and the learning rate are set to 100k iterations and $2 \times 10^{-4}$ for DDPM, and 30k iterations and $1 \times 10^{-4}$ for DiT. Both models are optimized using Adam with zero weight decay.
\end{itemize}
% \clover{TODO: the 3 pruning methods}

% \clover{TODO: $T^s$, $T^f$}

% \clover{TODO: mention num of stages in pruning method, give brief intro for pruning methods}

\myparagraph{Details for PruneForget.}
\label{sec:supp_implemet_detail_pruneforget} The standardization of the sensitivity score is computed as 
\bea
S'_{l[k]} = \text{Standardize}(S_{l[k]}) = 1 + \frac{S_{l[k]}}{S_{l[k]}^{\text{median}}+\epsilon}, 
\eea
where $\epsilon$ is set to $10^{-8}$.
For the unlearning loss $\gL_{\tt U}$ applied in Stages 1 and 3, we follow~\citet{huang2024unified}
and apply a gradient-based masked parameter update. We set the mask sparsity to 0.5, meaning 50\%
of the weights are updated by $\gL_{\tt U}$.
Considering that directly applying $\gL_{\tt U}$ from the very beginning of a pruned model can
destabilize the recovery of retain-set performance, we apply $\gL_{\tt U}$ during the interval
$0.5T_f < t < 0.9T_f$ in Stage 3 for generative models, incorporating an early stopping mechanism
when the pixel-wise mean squared error for $\gU$ exceeds 0.2.
We choose $f \triangleq f^{\text{base}} \cdot |\gD \setminus \gU|/|\gU|$ and the unlearning loss
weight $\alpha \triangleq |\gU|/|\gD \setminus \gU|$, to counteract the imbalances in the dataset
sizes. We perform a grid search over
$f^{\text{base}} \in \{20, 50, 80, 100\}$ for ResNet-18, with
the detailed search results reported in \tabref{tab:cls_cifar_fbase} and
\tabref{tab:cls_cifar_class_fbase}. Following the same strategy for the other architectures, we set
$f^{\text{base}} = 3$ for Swin-T and $f^{\text{base}} = 5$ for both DDPM and DiT.
The regularization weight $\lambda$ is set to $1 \times 10^{-3}$, and the supervised loss weight
$\eta$ follows the learning rate of the pruning mechanism.
For the updating sensitivity variant evaluated in \tabref{tab:dynamic_S}, sensitivity scores are calculated via a running mean over 20 batches of data and updated every 10 epochs throughout Stage 1.

\begin{table*}[t]
    \setlength{\tabcolsep}{3pt}
    \renewcommand{\arraystretch}{1.15}
\centering
\caption{\textbf{Hyperparameter search of $f_{base}$ on CIFAR-10 (ResNet-18) random unlearning.} We use grid search to determine the most suitable unlearning frequencies and identify $f_{base}=20$ and $f_{base}=80$ as the optimal settings for 10\% and 50\% unlearning, respectively. }
\vspace{-0.2cm}
\resizebox{1.01\textwidth}{!}{
\begin{tabular}{l|ccc | >{\columncolor{avg_highlight}}c | cc | ccc | >{\columncolor{avg_highlight}}c | cc}
\specialrule{.15em}{.05em}{.05em}
\multirow{2}{*}{\textbf{$f_{base}$}} & \multicolumn{6}{c|}{\textbf{Random Unlearn 10\%}} & \multicolumn{6}{c}{\textbf{Random Unlearn 50\%}} \\
 & \textbf{$\Delta$UA $\downarrow$} & \textbf{$\Delta$RA $\downarrow$} & \textbf{$\Delta$TA $\downarrow$} & \textbf{Avg $\Delta$ $\downarrow$} & \textbf{$D_{\text{KL}}$ $\downarrow$} & \textbf{Time} & \textbf{$\Delta$UA $\downarrow$} & \textbf{$\Delta$RA $\downarrow$} & \textbf{$\Delta$TA $\downarrow$} & \textbf{Avg $\Delta$ $\downarrow$} & \textbf{$D_{\text{KL}}$ $\downarrow$} & \textbf{Time} \\
\hline
$20$ & \textbf{\textcolor{colorBest}{0.03}} (6.01$_{\pm 0.28}$) & \textbf{\textcolor{colorBest}{0.00}} (100.00$_{\pm 0.00}$) & \textbf{\textcolor{colorBest}{0.01}} (93.73$_{\pm 0.22}$) & \textbf{\textcolor{colorBest}{0.01}} & \textbf{\textcolor{colorBest}{0.22}} & 998 & 1.43 (9.03$_{\pm 1.45}$) & 0.01 (99.99$_{\pm 0.01}$) & 0.67 (91.54$_{\pm 0.75}$) & 0.70 & 0.30 & 609 \\
$50$ & \textit{\textbf{0.06}} (6.09$_{\pm 0.42}$) & \textbf{\textcolor{colorBest}{0.00}} (100.00$_{\pm 0.00}$) & 0.25 (93.49$_{\pm 0.26}$) & 0.10 & \textbf{\textcolor{colorBest}{0.22}} & 996 & 0.57 (8.17$_{\pm 0.23}$) & \textbf{\textcolor{colorBest}{0.00}} (100.00$_{\pm 0.00}$) & 0.21 (92.00$_{\pm 0.28}$) & 0.26 & 0.27 & 603 \\
$80$ & 0.12 (5.92$_{\pm 0.49}$) & \textbf{\textcolor{colorBest}{0.00}} (100.00$_{\pm 0.00}$) & \textit{\textbf{0.07}} (93.67$_{\pm 0.24}$) & \textit{\textbf{0.06}} & \textbf{\textcolor{colorBest}{0.22}} & 986 & \textbf{\textcolor{colorBest}{0.16}} (7.44$_{\pm 0.32}$) & \textbf{\textcolor{colorBest}{0.00}} (100.00$_{\pm 0.00}$) & \textbf{\textcolor{colorBest}{0.02}} (92.22$_{\pm 0.23}$) & \textbf{\textcolor{colorBest}{0.06}} & \textbf{\textcolor{colorBest}{0.25}} & 594 \\
$100$ & 0.20 (5.84$_{\pm 0.40}$) & \textbf{\textcolor{colorBest}{0.00}} (100.00$_{\pm 0.00}$) & 0.16 (93.59$_{\pm 0.19}$) & 0.12 & \textbf{\textcolor{colorBest}{0.22}} & 973 & \textit{\textbf{0.19}} (7.41$_{\pm 0.36}$) & \textbf{\textcolor{colorBest}{0.00}} (100.00$_{\pm 0.00}$) & \textit{\textbf{0.04}} (92.24$_{\pm 0.22}$) & \textit{\textbf{0.07}} & \textbf{\textcolor{colorBest}{0.25}} & 579 \\
\specialrule{.15em}{.05em}{.05em}
\end{tabular}
}
\label{tab:cls_cifar_fbase}
\vspace{-0.1cm}
\end{table*}

\begin{table*}[t]
    \setlength{\tabcolsep}{3pt}
    \renewcommand{\arraystretch}{1.15}
\centering
\caption{\textbf{Hyperparameter search of $f_{base}$ on CIFAR-10 (ResNet-18) class unlearning.} Grid search results indicate that $f_{base}=100$ achieves the best performance for both settings.}
\vspace{-0.2cm}
\resizebox{1.01\textwidth}{!}{
\begin{tabular}{l|ccc | >{\columncolor{avg_highlight}}c | cc | ccc | >{\columncolor{avg_highlight}}c | cc}
\specialrule{.15em}{.05em}{.05em}
\multirow{2}{*}{\textbf{$f_{base}$}} & \multicolumn{6}{c|}{\textbf{Unlearn 1 Class (10\%)}} & \multicolumn{6}{c}{\textbf{Unlearn 5 Classes (50\%)}} \\
 & \textbf{$\Delta$UA $\downarrow$} & \textbf{$\Delta$RA $\downarrow$} & \textbf{$\Delta$TA $\downarrow$} & \textbf{Avg $\Delta$ $\downarrow$} & \textbf{$D_{\text{KL}}$ $\downarrow$} & \textbf{Time} & \textbf{$\Delta$UA $\downarrow$} & \textbf{$\Delta$RA $\downarrow$} & \textbf{$\Delta$TA $\downarrow$} & \textbf{Avg $\Delta$ $\downarrow$} & \textbf{$D_{\text{KL}}$ $\downarrow$} & \textbf{Time} \\
\hline
$20$ & \textbf{\textcolor{colorBest}{0.00}} (100.00$_{\pm 0.00}$) & \textbf{\textcolor{colorBest}{0.00}} (100.00$_{\pm 0.00}$) & 0.51 (93.76$_{\pm 0.68}$) & 0.17 & \textit{\textbf{0.28}} & 992 & \textbf{\textcolor{colorBest}{0.00}} (100.00$_{\pm 0.00}$) & \textbf{\textcolor{colorBest}{0.00}} (100.00$_{\pm 0.00}$) & 0.42 (95.96$_{\pm 1.23}$) & 0.14 & 0.40 & 602 \\
$50$ & \textbf{\textcolor{colorBest}{0.00}} (100.00$_{\pm 0.00}$) & \textbf{\textcolor{colorBest}{0.00}} (100.00$_{\pm 0.00}$) & \textit{\textbf{0.45}} (93.82$_{\pm 0.83}$) & \textit{\textbf{0.15}} & \textbf{\textcolor{colorBest}{0.28}} & 984 & \textbf{\textcolor{colorBest}{0.00}} (100.00$_{\pm 0.00}$) & \textbf{\textcolor{colorBest}{0.00}} (100.00$_{\pm 0.00}$) & 0.27 (96.10$_{\pm 1.29}$) & 0.09 & \textit{\textbf{0.39}} & 600 \\
$80$ & \textbf{\textcolor{colorBest}{0.00}} (100.00$_{\pm 0.00}$) & \textbf{\textcolor{colorBest}{0.00}} (100.00$_{\pm 0.00}$) & \textbf{\textcolor{colorBest}{0.41}} (93.86$_{\pm 0.71}$) & \textbf{\textcolor{colorBest}{0.14}} & \textbf{\textcolor{colorBest}{0.28}} & 984 & \textbf{\textcolor{colorBest}{0.00}} (100.00$_{\pm 0.00}$) & \textbf{\textcolor{colorBest}{0.00}} (100.00$_{\pm 0.00}$) & \textit{\textbf{0.25}} (96.13$_{\pm 1.28}$) & \textit{\textbf{0.08}} & 0.40 & 600 \\
$100$ & \textbf{\textcolor{colorBest}{0.00}} (100.00$_{\pm 0.00}$) & \textbf{\textcolor{colorBest}{0.00}} (100.00$_{\pm 0.00}$) & \textbf{\textcolor{colorBest}{0.41}} (93.86$_{\pm 0.71}$) & \textbf{\textcolor{colorBest}{0.14}} & \textbf{\textcolor{colorBest}{0.28}} & 984 & \textbf{\textcolor{colorBest}{0.00}} (100.00$_{\pm 0.00}$) & \textbf{\textcolor{colorBest}{0.00}} (100.00$_{\pm 0.00}$) & \textbf{\textcolor{colorBest}{0.20}} (96.17$_{\pm 1.21}$) & \textbf{\textcolor{colorBest}{0.07}} & \textbf{\textcolor{colorBest}{0.38}} & 596 \\
\specialrule{.15em}{.05em}{.05em}
\end{tabular}
}
\label{tab:cls_cifar_class_fbase}
\vspace{-0.1cm}
\end{table*}

\label{sec:supp_implemet_detail_baselines}
\myparagraph{Details for baselines.}
Our experimental framework is based on the officially released code of SFRon~\cite{huang2024unified}. We follow their implementation to obtain the pre-trained models, retrained models, and the core SFRon unlearning algorithm. For the remaining baselines, we refer to their respective papers and properly adapt them to our experimental settings.
\begin{itemize}[topsep=0pt, leftmargin=12pt]
    \item \textbf{$\ell_1$-MU~\cite{jia2023model}:} For ResNet-18 on CIFAR-10, we employ an $\ell_1$ penalty as the regularizer $\gR(\theta)$ during Stage 1, followed by the two standard pruning stages. For Swin-T and DDPM, since the applied pruning methods consist of only the latter two stages, we introduce an additional 10-epoch $\ell_1$ regularization phase before Stages 2 and 3. We apply a linear decay to the regularization coefficient $\lambda$, as detailed in their original paper. We compute the importance weights using an $\ell_1$-based group magnitude metric. 
    \item \textbf{Un-pruning~\cite{xiao2025right}:} The un-pruning process operates through an $N$-step iterative cycle of weight re-initialization, $k$ epochs of unlearning via fine-tuning (FT), and soft pruning, culminating in a final physical pruning step. We utilize FT as the unlearning algorithm due to its superior empirical results. To ensure the total computational runtime remains comparable to our default pruning baselines, we configure the iterative process to 20 steps of 10 epochs for ResNet-18, and 3 steps of 25 epochs for both Swin-T and DDPM. Finally, following the magnitude-based strategy of the original paper, we compute structural importance using an $\ell_2$-norm group magnitude metric.
    \item \textbf{Bilevel~\cite{shirkavand2025efficient}:} This baseline combines an unlearning loss with knowledge distillation from the pre-trained network in the post-pruning training stage. Because structured pruning inherently alters the network's intermediate feature dimensions, we omit feature-level distillation and rely exclusively on logit-based distillation. For the unlearning objective, we employ gradient ascent for classification tasks and random labeling for generative tasks. Following the original authors' configuration, we train the model for 20,000 iterations, executing 20 lower steps between two upper steps.
    \item \textbf{LLM-Eraser~\cite{zhang2025llm}:} This framework alternates between iterative pruning and unlearning steps. We utilize the authors' proposed Taylor score-based method to determine weight importance. For the unlearning phase, we strictly follow their algorithmic design, optimizing a joint objective that comprises generic selective training, KL divergence minimization, and contrastive disentanglement. To ensure the total computational runtime remains comparable, we configure the iterative process to 10 steps of 10 epochs for ResNet-18, 1 step of 25 epochs for Swin-T, 8 steps of 10 epochs for DDPM, and 4 steps of 2 epochs for DiT.
\end{itemize}

% \clover{TODO: Details for Pretrain, retrain.}

% \clover{TODO: explain how to implement the baselines}

\section{Additional Results for Classification}
\label{sec:supp_additional_exp_class}

\begin{table}[t]
\small
    \setlength{\tabcolsep}{3pt}
    \centering
\caption{\textbf{Comparison with pure unlearning.} We compare pure unlearning (baselines include FT~\cite{warnecke2023machine}, GA~\cite{thudi2022unrolling}, SalUn~\cite{fan2024salun}, BT~\cite{chundawat2023can}, SCRUB~\cite{kurmanji2023towards}, and SFRon~\cite{huang2024unified}) against unlearning with pruning on a 10\% random subset of CIFAR-10. Note that these are inherently distinct tasks evaluated against different respective \hl{oracles}. }
{
\begin{tabular}{l|ccc | >{\columncolor{avg_highlight}}c | c}
\specialrule{.15em}{.05em}{.05em}
    \multirow{2}{*}{\textbf{Method}} & \multicolumn{5}{c}{\textbf{Random Unlearn 10\%}} \\
 & \textbf{$\Delta$UA $\downarrow$} & \textbf{$\Delta$RA $\downarrow$} & \textbf{$\Delta$TA $\downarrow$} & \textbf{Avg $\Delta$ $\downarrow$} & \textbf{$D_{\text{KL}}$ $\downarrow$} \\
\hline
\multicolumn{6}{c}{\cellcolor{gray!20}\textbf{Task: Pure Unlearning (No Pruning)}} \\
\hline
\cellcolor{colorOracle}Retrain & \cellcolor{colorOracle} 0.00 (4.38$_{\pm 0.25}$) & \cellcolor{colorOracle} 0.00 (100.00$_{\pm 0.00}$) & \cellcolor{colorOracle} 0.00 (95.34$_{\pm 0.08}$) & \cellcolor{colorOracle} 0.00 & \cellcolor{colorOracle} 0.00 \\
FT & 4.28 (0.10$_{\pm 0.05}$) & \textbf{\textcolor{colorBest}{0.01}} (99.99$_{\pm 0.00}$) & \textbf{\textcolor{colorBest}{0.39}} (94.94$_{\pm 0.15}$) & 1.56 & 0.26 \\
GA & \textit{\textbf{1.71}} (6.09$_{\pm 1.67}$) & 6.24 (93.76$_{\pm 1.89}$) & 8.34 (87.00$_{\pm 1.64}$) & 5.43 & 0.36 \\
SalUn & 4.38 (0.00$_{\pm 0.01}$) & \textbf{\textcolor{colorBest}{0.01}} (99.99$_{\pm 0.01}$) & \textit{\textbf{0.45}} (94.89$_{\pm 0.09}$) & 1.61 & 0.27 \\
BT & 3.26 (1.12$_{\pm 0.00}$) & \textbf{\textcolor{colorBest}{0.01}} (99.99$_{\pm 0.00}$) & 0.71 (94.63$_{\pm 0.06}$) & \textit{\textbf{1.33}} & \textit{\textbf{0.24}} \\
SCRUB & 3.82 (0.56$_{\pm 0.31}$) & \textit{\textbf{0.12}} (99.88$_{\pm 0.08}$) & 1.20 (94.13$_{\pm 0.35}$) & 1.71 & 0.25 \\
SFRon & \textbf{\textcolor{colorBest}{0.96}} (3.42$_{\pm 0.77}$) & \textit{\textbf{0.12}} (99.88$_{\pm 0.16}$) & 1.15 (94.19$_{\pm 0.33}$) & \textbf{\textcolor{colorBest}{0.74}} & \textbf{\textcolor{colorBest}{0.15}} \\
\hline
\multicolumn{6}{c}{\cellcolor{gray!20}\textbf{Task: Unlearning + Pruning}} \\
\hline
\cellcolor{colorOracle}Retrain$\to$3SP & \cellcolor{colorOracle} 0.00 (6.03$_{\pm 0.56}$) & \cellcolor{colorOracle} 0.00 (100.00$_{\pm 0.00}$) & \cellcolor{colorOracle} 0.00 (93.74$_{\pm 0.32}$) & \cellcolor{colorOracle} 0.00 & \cellcolor{colorOracle} 0.00 \\
Pretrain$\to$3SP & 0.89 (5.15$_{\pm 0.38}$) & \textbf{\textcolor{colorBest}{0.00}} (100.00$_{\pm 0.00}$) & 0.41 (94.15$_{\pm 0.20}$) & 0.43 & \textbf{\textcolor{colorBest}{0.20}} \\
3SP$\to$SFRon & 14.84 (20.88$_{\pm 0.77}$) & 1.88 (98.12$_{\pm 0.36}$) & 3.38 (90.37$_{\pm 0.37}$) & 6.70 & 0.24 \\
SFRon$\to$3SP & \textit{\textbf{0.16}} (5.87$_{\pm 0.57}$) & \textbf{\textcolor{colorBest}{0.00}} (100.00$_{\pm 0.01}$) & \textit{\textbf{0.04}} (93.70$_{\pm 0.31}$) & \textit{\textbf{0.07}} & \textit{\textbf{0.22}} \\
PruneForget & \textbf{\textcolor{colorBest}{0.03}} (6.01$_{\pm 0.28}$) & \textbf{\textcolor{colorBest}{0.00}} (100.00$_{\pm 0.00}$) & \textbf{\textcolor{colorBest}{0.01}} (93.73$_{\pm 0.22}$) & \textbf{\textcolor{colorBest}{0.01}} & \textit{\textbf{0.22}} \\
\specialrule{.15em}{.05em}{.05em}
\end{tabular}
}
\label{tab:supp_compare_with_no_prune}
\vspace{-0.2cm}
\end{table}
\myparagraph{Comparison with pure unlearning task (without pruning).}
In this section, we contrast our joint task of unlearning and pruning with pure unlearning paradigms, which update model weights without modifying the original architecture. \tabref{tab:supp_compare_with_no_prune} presents the results for 10\% random unlearning on CIFAR-10. Note that these two tasks require different oracle models for evaluation. Pure unlearning targets a standard retrained model. Differently, our oracle for the joint task is a model trained from scratch and then pruned. We observe that sequentially applying pruning to the best-performing pure unlearning baseline (3SP$\to$SFRon and SFRon$\to$3SP) yields suboptimal results. In contrast, our proposed method achieves significantly better performance in this combined setting. 

%As a result, the two oracles exhibit different performance reflecting their architectural disparities.

\begin{table}[t]
\small
    \setlength{\tabcolsep}{3pt}
    \centering
\caption{\textbf{Comparison with pruning then retraining from scratch.} We evaluate the Prune$\to$retrain baseline on CIFAR-10 50\% random unlearning. The results demonstrate that pruned models are inherently difficult to optimize from a random initialization.}
{
\begin{tabular}{l|ccc | >{\columncolor{avg_highlight}}c | cc}
\specialrule{.15em}{.05em}{.05em}
\multirow{2}{*}{\textbf{Method}} & \multicolumn{6}{c}{\textbf{Random Unlearn 50\%}} \\
 & \textbf{$\Delta$UA $\downarrow$} & \textbf{$\Delta$RA $\downarrow$} & \textbf{$\Delta$TA $\downarrow$} & \textbf{Avg $\Delta$ $\downarrow$} & \textbf{$D_{\text{KL}}$ $\downarrow$} & \textbf{Time (s)} \\
\hline
\cellcolor{colorOracle}Retrain$\to$3SP & \cellcolor{colorOracle} 0.00 (7.60$_{\pm 0.22}$) & \cellcolor{colorOracle} 0.00 (100.00$_{\pm 0.00}$) & \cellcolor{colorOracle} 0.00 (92.20$_{\pm 0.30}$) & \cellcolor{colorOracle} 0.00 & \cellcolor{colorOracle} 0.00 & \cellcolor{colorOracle} 1518 \\
Pretrain$\to$3SP & 1.54 (6.06$_{\pm 0.17}$) & \textbf{\textcolor{colorBest}{0.00}} (100.00$_{\pm 0.00}$) & \textit{\textbf{0.81}} (93.01$_{\pm 0.27}$) & \textit{\textbf{0.78}} & \textbf{\textcolor{colorBest}{0.23}} & 556 \\
Prune$\to$retrain & 2.05 (9.65$_{\pm 0.19}$) & 0.01 (99.99$_{\pm 0.01}$) & 3.05 (89.15$_{\pm 0.24}$) & 1.70 & 0.35 & 598 \\
\hline
PruneForget & \textbf{\textcolor{colorBest}{0.16}} (7.44$_{\pm 0.32}$) & \textbf{\textcolor{colorBest}{0.00}} (100.00$_{\pm 0.00}$) & \textbf{\textcolor{colorBest}{0.02}} (92.22$_{\pm 0.23}$) & \textbf{\textcolor{colorBest}{0.06}} & \textit{\textbf{0.25}} & 594 \\
\specialrule{.15em}{.05em}{.05em}
\end{tabular}
}
\label{tab:supp_cifar_compare_scratch_50}
\vspace{-0.2cm}
\end{table}
\myparagraph{Comparison with prune-then-retrain baseline.}
While we treat the retrain-then-prune baseline as our oracle, we evaluate an additional baseline, {\it prune-then-retrain}, for random 50\% unlearning on CIFAR-10 in~\tabref{tab:supp_cifar_compare_scratch_50}.
Specifically, we retrain a pruned model from scratch. Thus, the primary difference between \texttt{Pretrain$\to$3SP} and \texttt{Prune$\to$retrain} is that the latter reinitializes the weight parameters before the fine-tuning stage. In~\tabref{tab:supp_cifar_compare_scratch_50}, we observe that retraining the pruned models results in lower utility, as the TA drops by 3.86\% (89.15\% \textit{vs.} 93.01\%) compared to the model fine-tuned without reinitialization. This observation aligns with the Lottery Ticket Hypothesis~\cite{frankle2018lottery}, which describes that pruned models are difficult to train
from scratch.

\begin{table*}[t]
    \setlength{\tabcolsep}{3pt}
    \renewcommand{\arraystretch}{1.15}
\centering
\caption{\textbf{CIFAR-10 (ResNet-18) random unlearning on classification with $p=0.3$.} Among all baselines, PruneForget achieves the lowest \hlavg{$\text{Avg }\Delta$} under a low pruning ratio.}
\vspace{-0.2cm}
\resizebox{1.01\textwidth}{!}{
\begin{tabular}{l|ccc | >{\columncolor{avg_highlight}}c | cc | ccc | >{\columncolor{avg_highlight}}c | cc}
\specialrule{.15em}{.05em}{.05em}
\multirow{2}{*}{\textbf{Method}} & \multicolumn{6}{c|}{\textbf{Random Unlearn 10\%}} & \multicolumn{6}{c}{\textbf{Random Unlearn 50\%}} \\
 & \textbf{$\Delta$UA $\downarrow$} & \textbf{$\Delta$RA $\downarrow$} & \textbf{$\Delta$TA $\downarrow$} & \textbf{Avg $\Delta$ $\downarrow$} & \textbf{$D_{\text{KL}}$ $\downarrow$} & \textbf{Time} & \textbf{$\Delta$UA $\downarrow$} & \textbf{$\Delta$RA $\downarrow$} & \textbf{$\Delta$TA $\downarrow$} & \textbf{Avg $\Delta$ $\downarrow$} & \textbf{$D_{\text{KL}}$ $\downarrow$} & \textbf{Time} \\
\hline
\cellcolor{colorOracle}Retrain$\to$3SP & \cellcolor{colorOracle} 0.00 (5.15$_{\pm 0.31}$) & \cellcolor{colorOracle} 0.00 (100.00$_{\pm 0.00}$) & \cellcolor{colorOracle} 0.00 (94.75$_{\pm 0.18}$) & \cellcolor{colorOracle} 0.00 & \cellcolor{colorOracle} 0.00 & \cellcolor{colorOracle} 2212 & \cellcolor{colorOracle} 0.00 (6.98$_{\pm 0.20}$) & \cellcolor{colorOracle} 0.00 (100.00$_{\pm 0.00}$) & \cellcolor{colorOracle} 0.00 (92.78$_{\pm 0.20}$) & \cellcolor{colorOracle} 0.00 & \cellcolor{colorOracle} 0.00 & \cellcolor{colorOracle} 1494 \\
Pretrain$\to$3SP & 1.14 (4.01$_{\pm 0.10}$) & \textbf{\textcolor{colorBest}{0.00}} (100.00$_{\pm 0.00}$) & \textbf{\textcolor{colorBest}{0.05}} (94.80$_{\pm 0.14}$) & \textit{\textbf{0.40}} & \textbf{\textcolor{colorBest}{0.14}} & 1057 & 2.40 (4.58$_{\pm 0.12}$) & \textbf{\textcolor{colorBest}{0.00}} (100.00$_{\pm 0.00}$) & 1.04 (93.82$_{\pm 0.16}$) & 1.15 & \textbf{\textcolor{colorBest}{0.20}} & 638 \\
3SP$\to$SFRon & 15.17 (20.32$_{\pm 0.61}$) & 1.78 (98.22$_{\pm 0.13}$) & 4.10 (90.65$_{\pm 0.23}$) & 7.02 & 0.21 & 1171 & 5.30 (12.29$_{\pm 0.29}$) & 0.41 (99.59$_{\pm 0.05}$) & 2.90 (89.88$_{\pm 0.15}$) & 2.87 & 0.25 & 698 \\
SFRon$\to$3SP & \textit{\textbf{1.10}} (6.25$_{\pm 0.41}$) & \textbf{\textcolor{colorBest}{0.00}} (100.00$_{\pm 0.00}$) & 1.35 (93.40$_{\pm 0.23}$) & 0.82 & 0.20 & 1167 & \textit{\textbf{0.14}} (6.84$_{\pm 0.16}$) & \textbf{\textcolor{colorBest}{0.00}} (100.00$_{\pm 0.00}$) & \textit{\textbf{0.22}} (92.55$_{\pm 0.29}$) & \textit{\textbf{0.12}} & 0.25 & 707 \\
$\ell_1$-MU$\to$2SP & 5.94 (11.09$_{\pm 0.26}$) & 5.20 (94.80$_{\pm 0.29}$) & 5.95 (88.80$_{\pm 0.31}$) & 5.69 & 0.27 & 931 & 5.46 (12.44$_{\pm 0.18}$) & 3.25 (96.75$_{\pm 0.26}$) & 5.41 (87.37$_{\pm 0.21}$) & 4.71 & 0.33 & 545 \\
Un-pruning & 1.92 (7.07$_{\pm 0.33}$) & \textit{\textbf{0.02}} (99.98$_{\pm 0.01}$) & 2.12 (92.63$_{\pm 0.11}$) & 1.35 & 0.35 & 914 & 2.54 (9.52$_{\pm 0.12}$) & \textit{\textbf{0.03}} (99.97$_{\pm 0.01}$) & 2.71 (90.06$_{\pm 0.27}$) & 1.76 & 0.43 & 553 \\
Bilevel & 8.49 (13.64$_{\pm 0.75}$) & 6.98 (93.02$_{\pm 0.96}$) & 7.97 (86.78$_{\pm 0.86}$) & 7.81 & 0.54 & 1023 & 6.73 (13.72$_{\pm 1.12}$) & 2.94 (97.06$_{\pm 0.91}$) & 6.70 (86.08$_{\pm 1.04}$) & 5.46 & 0.61 & 722 \\
LLM-Eraser & 13.51 (18.66$_{\pm 2.64}$) & 16.53 (83.47$_{\pm 2.39}$) & 13.76 (80.99$_{\pm 2.28}$) & 14.60 & 0.53 & 1589 & 10.30 (17.29$_{\pm 1.51}$) & 13.02 (86.98$_{\pm 1.68}$) & 10.29 (82.49$_{\pm 1.62}$) & 11.20 & 0.45 & 853 \\
\hline
PruneForget & \textbf{\textcolor{colorBest}{0.01}} (5.14$_{\pm 0.29}$) & \textbf{\textcolor{colorBest}{0.00}} (100.00$_{\pm 0.00}$) & \textit{\textbf{0.32}} (94.43$_{\pm 0.15}$) & \textbf{\textcolor{colorBest}{0.11}} & \textit{\textbf{0.16}} & 1103 & \textbf{\textcolor{colorBest}{0.07}} (6.92$_{\pm 0.14}$) & \textbf{\textcolor{colorBest}{0.00}} (100.00$_{\pm 0.00}$) & \textbf{\textcolor{colorBest}{0.16}} (92.94$_{\pm 0.15}$) & \textbf{\textcolor{colorBest}{0.07}} & \textit{\textbf{0.22}} & 651 \\
\specialrule{.15em}{.05em}{.05em}
\end{tabular}
}
\label{tab:cls_cifar_rand_pr30}
\vspace{-0.1cm}
\end{table*}

\begin{table*}[t]
    \setlength{\tabcolsep}{3pt}
    \renewcommand{\arraystretch}{1.15}
\centering
\caption{\textbf{CIFAR-10 (ResNet-18) random unlearning on classification with $p=0.7$.} PruneForget achieves a small gap to the \hl{oracle} under a high pruning ratio.}
\vspace{-0.2cm}
\resizebox{1.01\textwidth}{!}{
\begin{tabular}{l|ccc | >{\columncolor{avg_highlight}}c | cc | ccc | >{\columncolor{avg_highlight}}c | cc}
\specialrule{.15em}{.05em}{.05em}
\multirow{2}{*}{\textbf{Method}} & \multicolumn{6}{c|}{\textbf{Random Unlearn 10\%}} & \multicolumn{6}{c}{\textbf{Random Unlearn 50\%}} \\
 & \textbf{$\Delta$UA $\downarrow$} & \textbf{$\Delta$RA $\downarrow$} & \textbf{$\Delta$TA $\downarrow$} & \textbf{Avg $\Delta$ $\downarrow$} & \textbf{$D_{\text{KL}}$ $\downarrow$} & \textbf{Time} & \textbf{$\Delta$UA $\downarrow$} & \textbf{$\Delta$RA $\downarrow$} & \textbf{$\Delta$TA $\downarrow$} & \textbf{Avg $\Delta$ $\downarrow$} & \textbf{$D_{\text{KL}}$ $\downarrow$} & \textbf{Time} \\
\hline
\cellcolor{colorOracle}Retrain$\to$3SP & \cellcolor{colorOracle} 0.00 (8.39$_{\pm 0.31}$) & \cellcolor{colorOracle} 0.00 (99.89$_{\pm 0.00}$) & \cellcolor{colorOracle} 0.00 (91.34$_{\pm 0.18}$) & \cellcolor{colorOracle} 0.00 & \cellcolor{colorOracle} 0.00 & \cellcolor{colorOracle} 2131 & \cellcolor{colorOracle} 0.00 (9.71$_{\pm 0.20}$) & \cellcolor{colorOracle} 0.00 (99.98$_{\pm 0.00}$) & \cellcolor{colorOracle} 0.00 (90.09$_{\pm 0.20}$) & \cellcolor{colorOracle} 0.00 & \cellcolor{colorOracle} 0.00 & \cellcolor{colorOracle} 1454 \\
Pretrain$\to$3SP & 0.49 (7.90$_{\pm 0.38}$) & \textbf{\textcolor{colorBest}{0.03}} (99.92$_{\pm 0.02}$) & 0.40 (91.74$_{\pm 0.23}$) & \textit{\textbf{0.31}} & \textbf{\textcolor{colorBest}{0.14}} & 991 & \textbf{\textcolor{colorBest}{0.05}} (9.65$_{\pm 0.32}$) & \textbf{\textcolor{colorBest}{0.01}} (99.97$_{\pm 0.01}$) & \textbf{\textcolor{colorBest}{0.19}} (89.90$_{\pm 0.31}$) & \textbf{\textcolor{colorBest}{0.08}} & \textbf{\textcolor{colorBest}{0.20}} & 597 \\
3SP$\to$SFRon & 13.88 (22.26$_{\pm 0.94}$) & 7.48 (92.41$_{\pm 1.08}$) & 4.91 (86.43$_{\pm 0.93}$) & 8.76 & 0.21 & 1097 & 6.47 (16.18$_{\pm 0.58}$) & 5.39 (94.59$_{\pm 0.97}$) & 5.09 (85.00$_{\pm 0.46}$) & 5.65 & 0.25 & 618 \\
SFRon$\to$3SP & 1.28 (9.66$_{\pm 0.74}$) & 0.50 (99.39$_{\pm 0.37}$) & 1.11 (90.23$_{\pm 0.70}$) & 0.96 & 0.20 & 1120 & 2.43 (12.14$_{\pm 0.42}$) & 0.33 (99.65$_{\pm 0.09}$) & 2.62 (87.47$_{\pm 0.46}$) & 1.79 & 0.25 & 668 \\
$\ell_1$-MU$\to$2SP & 3.17 (11.56$_{\pm 0.46}$) & 5.76 (94.13$_{\pm 0.33}$) & 3.05 (88.29$_{\pm 0.32}$) & 3.99 & 0.27 & 859 & 3.38 (13.09$_{\pm 0.25}$) & 4.17 (95.81$_{\pm 0.34}$) & 3.42 (86.68$_{\pm 0.30}$) & 3.66 & 0.33 & 505 \\
Un-pruning & 1.73 (10.12$_{\pm 0.41}$) & 1.26 (98.63$_{\pm 0.46}$) & 1.73 (89.62$_{\pm 0.41}$) & 1.57 & 0.35 & 918 & 2.94 (12.65$_{\pm 0.40}$) & 1.67 (98.30$_{\pm 0.40}$) & 3.19 (86.90$_{\pm 0.52}$) & 2.60 & 0.43 & 536 \\
Bilevel & 7.33 (15.72$_{\pm 1.43}$) & 10.21 (89.68$_{\pm 1.74}$) & 6.66 (84.69$_{\pm 1.58}$) & 8.06 & 0.54 & 965 & 5.78 (15.48$_{\pm 0.82}$) & 6.17 (93.81$_{\pm 0.96}$) & 5.76 (84.33$_{\pm 0.96}$) & 5.90 & 0.61 & 671 \\
LLM-Eraser & 11.92 (20.30$_{\pm 3.24}$) & 18.20 (81.69$_{\pm 3.08}$) & 11.92 (79.42$_{\pm 2.75}$) & 14.01 & 0.53 & 1415 & 9.63 (19.33$_{\pm 1.33}$) & 16.02 (83.96$_{\pm 1.23}$) & 9.85 (80.24$_{\pm 1.25}$) & 11.83 & 0.45 & 750 \\
\hline
PruneForget & \textbf{\textcolor{colorBest}{0.01}} (8.38$_{\pm 0.38}$) & \textbf{\textcolor{colorBest}{0.02}} (99.87$_{\pm 0.03}$) & \textbf{\textcolor{colorBest}{0.01}} (91.35$_{\pm 0.35}$) & \textbf{\textcolor{colorBest}{0.01}} & \textit{\textbf{0.16}} & 1008 & \textit{\textbf{0.51}} (10.21$_{\pm 0.32}$) & \textit{\textbf{0.04}} (99.94$_{\pm 0.03}$) & \textit{\textbf{0.78}} (89.31$_{\pm 0.35}$) & \textit{\textbf{0.44}} & \textit{\textbf{0.22}} & 606 \\
\specialrule{.15em}{.05em}{.05em}
\end{tabular}
}
\label{tab:cls_cifar_rand_pr70}
\end{table*}
\myparagraph{Comparison under different pruning ratios.}
To evaluate the robustness of our method across varying degrees of sparsity, we compare PruneForget against baselines at both lower ($p=0.3$) and higher ($p=0.7$) pruning ratios for random unlearning on CIFAR-10 with ResNet-18. As shown in~\tabref{tab:cls_cifar_rand_pr30}, PruneForget consistently achieves the lowest Avg~$\Delta$ under a low pruning ratio, recording an Avg~$\Delta$ of 0.11 and 0.07 for the 10\% and 50\% unlearning scenarios, respectively. Furthermore, \tabref{tab:cls_cifar_rand_pr70} demonstrates that PruneForget maintains a small gap to the oracle even under a higher pruning ratio. It achieves an Avg~$\Delta$ of 0.01 in the 10\% unlearning setting, while keeping the gap highly competitive (Avg~$\Delta = 0.44$) when unlearning 50\% of the data. These results highlight the stability and generalizability of PruneForget, proving its effectiveness across different compression regimes.

\begin{table*}[t]
    \setlength{\tabcolsep}{3pt}
\centering
\small
\caption{\textbf{CIFAR-100 (ResNet-18) random unlearning on classification.} PruneForget achieves performance comparable to the \hl{oracle} (Retrain$\to$3SP), demonstrating its generalizability across more complex datasets.}
\resizebox{0.9\columnwidth}{!}{
\begin{tabular}{l|ccc | >{\columncolor{avg_highlight}}c | cc}
\specialrule{.15em}{.05em}{.05em}
\multirow{2}{*}{\textbf{Method}} & \multicolumn{6}{c}{\textbf{Random Unlearn 50\%}} \\
 & \textbf{$\Delta$UA $\downarrow$} & \textbf{$\Delta$RA $\downarrow$} & \textbf{$\Delta$TA $\downarrow$} & \textbf{Avg $\Delta$ $\downarrow$} & \textbf{$D_{\text{KL}}$ $\downarrow$} & \textbf{Time} \\
\hline
\cellcolor{colorOracle}Retrain$\to$3SP & \cellcolor{colorOracle} 0.00 (32.48$_{\pm 0.31}$) & \cellcolor{colorOracle} 0.00 (99.99$_{\pm 0.01}$) & \cellcolor{colorOracle} 0.00 (67.40$_{\pm 0.18}$) & \cellcolor{colorOracle} 0.00 & \cellcolor{colorOracle} 0.00 & \cellcolor{colorOracle} 2551 \\
Pretrain$\to$3SP & 3.22 (29.27$_{\pm 0.71}$) & \textbf{\textcolor{colorBest}{0.00}} (99.99$_{\pm 0.01}$) & 1.50 (68.90$_{\pm 0.69}$) & 1.57 & \textbf{\textcolor{colorBest}{0.20}} & 567 \\
3SP$\to$SFRon& 5.39 (37.87$_{\pm 0.31}$) & 0.44 (99.55$_{\pm 0.05}$) & 1.67 (65.72$_{\pm 0.37}$) & 2.50 & 0.25 & 631 \\
SFRon$\to$3SP & \textit{\textbf{2.12}} (30.37$_{\pm 0.28}$) & \textbf{\textcolor{colorBest}{0.00}} (99.99$_{\pm 0.01}$) & \textit{\textbf{1.48}} (68.88$_{\pm 0.30}$) & \textit{\textbf{1.20}} & 0.25 & 635 \\
$\ell_1$-MU$\to$2SP & 6.28 (38.76$_{\pm 0.36}$) & 7.59 (92.40$_{\pm 0.74}$) & 6.13 (61.27$_{\pm 0.53}$) & 6.67 & 0.33 & 467 \\
Un-pruning & 5.19 (37.67$_{\pm 0.35}$) & 0.08 (99.90$_{\pm 0.03}$) & 5.08 (62.32$_{\pm 0.55}$) & 3.45 & 0.43 & 543 \\
Bilevel & 6.00 (38.48$_{\pm 1.03}$) & 4.07 (95.92$_{\pm 0.71}$) & 5.05 (62.35$_{\pm 0.87}$) & 5.04 & 0.61 & 625 \\
LLM-Eraser & 14.43 (46.91$_{\pm 1.45}$) & 33.41 (66.58$_{\pm 2.34}$) & 14.45 (52.95$_{\pm 1.68}$) & 20.76 & 0.45 & 793 \\
\hline
PruneForget & \textbf{\textcolor{colorBest}{1.46}} (33.94$_{\pm 1.24}$) & \textbf{\textcolor{colorBest}{0.00}} (99.99$_{\pm 0.01}$) & \textbf{\textcolor{colorBest}{0.69}} (66.70$_{\pm 0.76}$) & \textbf{\textcolor{colorBest}{0.72}} & \textit{\textbf{0.22}} & 581 \\
\specialrule{.15em}{.05em}{.05em}
\end{tabular}
}
\label{tab:cls_cifar100_rand_pr50}
\end{table*}
\myparagraph{Experiment results on CIFAR-100 (ResNet-18).}
While we have demonstrated the effectiveness of PruneForget for ResNet-18 on CIFAR-10, we further explore its generalizability across more complex datasets. \tabref{tab:cls_cifar100_rand_pr50} presents the results for 50\% random unlearning on the CIFAR-100 dataset. PruneForget outperforms all baselines, achieving the lowest Avg~$\Delta$ of just 0.72. It optimally balances unlearning and utility, recording the lowest gaps in unlearn ($\Delta$UA $= 1.46$) and test accuracy ($\Delta$TA $= 0.69$) while perfectly preserving retain accuracy ($\Delta$RA $= 0.00$). In contrast, the closest baseline (SFRon$\to$3SP) trails at Avg~$\Delta = 1.20$, and other baselines suffer severe degradation. Furthermore, PruneForget achieves these near-oracle results highly efficiently, operating over $4.3\times$ faster than the retrain oracle.

\section{Additional Results for Class-conditioned Generation}
\label{sec:supp_additional_exp_gene}
\myparagraph{Results on class-wise unlearning on CIFAR-10 (DDPM).}
 In~\tabref{tab:ddpm_cifar}, we show the class-wise unlearning results over 5 classes (\textit{Automobile}, \textit{Cat}, \textit{Dog}, \textit{Horse}, and \textit{Truck}). We observe PruneForget effectively matches the oracle model across all tasks. It yields the lowest $\Delta$UA in most cases alongside highly competitive $\Delta$FID scores, demonstrating its ability to effectively unlearn targeted concepts while maintaining overall image quality. Importantly, PruneForget achieves these near-zero performance gaps while accelerating runtime by over 10$\times$ relative to the oracle model.
 
\begin{table*}[t]
\small
    \setlength{\tabcolsep}{3pt}
    \centering
\caption{\textbf{CIFAR-10 (DDPM) class-wise unlearning on generation.} PruneForget consistently achieves close performance to the \hl{oracle}.}
% \vspace{-0.1cm}
\resizebox{\textwidth}{!}{ % Resize the table to fit within text width
\begin{tabular}{l|cc|cc|cc|cc|cc|c}
\specialrule{.15em}{.05em}{.05em}
    \multirow{2}{*}{\textbf{Method}} & \multicolumn{2}{c|}{{Automobile}} & \multicolumn{2}{c|}{{Cat}} & \multicolumn{2}{c|}{{Dog}} & \multicolumn{2}{c|}{{Horse}} & \multicolumn{2}{c|}{{Truck}} & \multirow{2}{*}{\textbf{Time}} \\
 & \textbf{$\Delta$UA $\downarrow$} & \textbf{$\Delta$FID $\downarrow$} & \textbf{$\Delta$UA $\downarrow$} & \textbf{$\Delta$FID $\downarrow$} & \textbf{$\Delta$UA $\downarrow$} & \textbf{$\Delta$FID $\downarrow$} & \textbf{$\Delta$UA $\downarrow$} & \textbf{$\Delta$FID $\downarrow$} & \textbf{$\Delta$UA $\downarrow$} & \textbf{$\Delta$FID $\downarrow$} & \\
\hline
\cellcolor{colorOracle}Retrain$\to$2SP & \cellcolor{colorOracle} 0.00 (98.87) & \cellcolor{colorOracle} 0.00 (17.10) & \cellcolor{colorOracle} 0.00 (95.50) & \cellcolor{colorOracle} 0.00 (15.96) & \cellcolor{colorOracle} 0.00 (98.79) & \cellcolor{colorOracle} 0.00 (16.26) & \cellcolor{colorOracle} 0.00 (99.12) & \cellcolor{colorOracle} 0.00 (17.54) & \cellcolor{colorOracle} 0.00 (99.66) & \cellcolor{colorOracle} 0.00 (16.78) & \cellcolor{colorOracle} 398622 \\
Pretrain$\to$2SP & 0.65 (98.22) & \textbf{\textcolor{colorBest}{0.10}} (17.20) & 0.55 (94.95) & 0.61 (16.57) & 0.38 (98.41) & \textit{\textbf{0.07}} (16.33) & 0.13 (98.99) & 0.21 (17.33) & 0.55 (99.11) & \textit{\textbf{0.52}} (17.30) & 21304 \\
2SP$\to$SFRon& 0.66 (98.21) & 0.34 (17.44) & \textit{\textbf{0.15}} (95.35) & \textit{\textbf{0.13}} (16.09) & 0.36 (98.43) & \textit{\textbf{0.07}} (16.19) & 0.19 (98.93) & \textbf{\textcolor{colorBest}{0.02}} (17.56) & 0.59 (99.07) & 0.53 (17.31) & 25652 \\
SFRon$\to$2SP & 0.81 (98.06) & \textit{\textbf{0.30}} (17.40) & 0.78 (94.72) & \textbf{\textcolor{colorBest}{0.11}} (16.07) & 0.25 (98.54) & \textbf{\textcolor{colorBest}{0.05}} (16.31) & 0.16 (98.96) & 0.21 (17.75) & 0.77 (98.89) & \textbf{\textcolor{colorBest}{0.47}} (17.25) & 26674 \\
$\ell_1$-MU$\to$2SP & \textit{\textbf{0.21}} (98.66) & 1.03 (18.13) & 0.39 (95.11) & 0.55 (16.51) & \textit{\textbf{0.16}} (98.63) & 0.63 (16.89) & \textit{\textbf{0.07}} (99.05) & 0.49 (18.03) & \textbf{\textcolor{colorBest}{0.15}} (99.51) & 1.11 (17.89) & 22675 \\
Un-pruning& 1.02 (99.89) & 371.14 (388.24) & 0.60 (94.90) & 286.38 (302.34) & 1.08 (99.87) & 315.46 (331.72) & 0.88 (100.00) & 331.98 (349.52) & \textbf{\textcolor{colorBest}{0.15}} (99.81) & 326.22 (343.00) &  36586\\
% Un-pruning~\cite{xiao2025right} & 1.13 (100.00) & 306.05 (323.15) & 26.63 (68.87) & 286.38 (302.34) & 1.18 (99.97) & 312.05 (328.31) & 0.88 (100.00) & 299.01 (316.55) & 0.61 (99.05) & 308.58 (325.36) & 4632 \\
Bilevel & 6.34 (92.53) & 1.01 (18.11) & 5.47 (90.03) & 1.38 (17.34) & 2.56 (96.23) & 1.04 (17.30) & 3.18 (95.94) & 0.22 (17.76) & 3.95 (95.71) & 1.26 (18.04) & 33863 \\
% LLM-Eraser & 1.17 (97.70) & 45.63 (62.73) & 10.59 (84.91) & 54.88 (70.84) & 5.28 (93.51) & 48.40 (64.66) & 0.56 (98.56) & 76.36 (93.90) & 0.70 (98.96) & 84.79 (101.57) & 1835 \\
LLM-Eraser & 1.05 (97.82) & 7.42 (24.52) & 3.49 (92.01) & 6.46 (22.42) & 9.74 (89.05) & 9.91 (26.17) & 0.19 (98.93) & 9.24 (26.78) & 0.93 (98.73) & 4.00 (20.78) & 27570 \\
\hline
PruneForget & \textbf{\textcolor{colorBest}{0.17}} (98.70) & 0.38 (17.48) & \textbf{\textcolor{colorBest}{0.13}} (95.37) & \textit{\textbf{0.13}} (15.83) & \textbf{\textcolor{colorBest}{0.04}} (98.75) & 0.10 (16.36) & \textbf{\textcolor{colorBest}{0.04}} (99.08) & \textit{\textbf{0.04}} (17.50) & \textit{\textbf{0.26}} (99.40) & 0.63 (17.41) & 26254 \\
\specialrule{.15em}{.05em}{.05em}
\end{tabular}
}
\label{tab:ddpm_cifar}
\vspace{-.05cm}
\end{table*}

\myparagraph{Results on class-wise unlearning on ImageNet (DiT-XL/2).}
\begin{table*}[h]
\small
    \setlength{\tabcolsep}{3pt}
    \renewcommand{\arraystretch}{1.15}
    \centering
\caption{\textbf{ImageNet-1K (DiT-XL/2) class-wise unlearning on generation.} PruneForget uniquely achieves perfect forgetting while maintaining generative quality. \textbf{Averages} (shaded) across the 5 classes are reported in the summary section. We omit the \hl{oracle} (Retrain$\to$2SP) due to computational constraints. Best results are in \textbf{\textcolor{colorBest}{blue}}, while second best are \textit{\textbf{italicized}}.}
\resizebox{\textwidth}{!}{ % Resize the table to fit within text width
\begin{tabular}{l|cc|cc|cc|cc|cc|>{\columncolor{avg_highlight}}c >{\columncolor{avg_highlight}}c|c}
\specialrule{.15em}{.05em}{.05em}
    \multirow{2}{*}{\textbf{Method}} & \multicolumn{2}{c|}{Cacatua galerita} & \multicolumn{2}{c|}{Golden retriever} & \multicolumn{2}{c|}{White wolf} & \multicolumn{2}{c|}{Arctic fox} & \multicolumn{2}{c|}{Otter} & \multicolumn{2}{c|}{\cellcolor{avg_highlight}\textbf{Average}} & \multirow{2}{*}{\textbf{Time}} \\
 & \textbf{UA $\uparrow$} & \textbf{FID $\downarrow$} & \textbf{UA $\uparrow$} & \textbf{FID $\downarrow$} & \textbf{UA $\uparrow$} & \textbf{FID $\downarrow$} & \textbf{UA $\uparrow$} & \textbf{FID $\downarrow$} & \textbf{UA $\uparrow$} & \textbf{FID $\downarrow$} & \textbf{UA $\uparrow$} & \textbf{FID $\downarrow$} & \\
\hline
Pretrain$\to$2SP & 4.2 & \textit{\textbf{12.51}} & 2.0 & \textbf{\textcolor{colorBest}{12.65}} & 2.6 & \textbf{\textcolor{colorBest}{12.74}} & \textit{\textbf{9.8}} & \textbf{\textcolor{colorBest}{12.60}} & 3.2 & \textit{\textbf{12.82}} & 4.36$_{\pm 3.15}$ & \textbf{\textcolor{colorBest}{12.66$_{\pm 0.12}$}} & 35296.4 \\
2SP$\to$SFRon    & \textbf{\textcolor{colorBest}{100.0}} & 115.01 & \textbf{\textcolor{colorBest}{100.0}} & 115.68 & \textbf{\textcolor{colorBest}{100.0}} & 21.62 & \textbf{\textcolor{colorBest}{100.0}} & 316.17 & \textbf{\textcolor{colorBest}{100.0}} & 22.25 & \textbf{\textcolor{colorBest}{100.0$_{\pm 0.00}$}} & 118.15$_{\pm 120.15}$ & 35930.8 \\
SFRon$\to$2SP    & \textit{\textbf{5.4}} & \textbf{\textcolor{colorBest}{12.36}} & \textit{\textbf{2.8}} & \textit{\textbf{12.73}} & \textit{\textbf{8.8}} & \textit{\textbf{13.18}} & 5.0 & \textit{\textbf{12.87}} & \textit{\textbf{10.8}} & \textbf{\textcolor{colorBest}{12.80}} & \textit{\textbf{6.56$_{\pm 3.20}$}} & \textit{\textbf{12.79$_{\pm 0.29}$}} & 37440.7 \\
LLM-Eraser      & \textbf{\textcolor{colorBest}{100.0}} & 564.48 & \textbf{\textcolor{colorBest}{100.0}} & 359.76 & \textbf{\textcolor{colorBest}{100.0}} & 564.48 & \textbf{\textcolor{colorBest}{100.0}} & 357.60 & \textbf{\textcolor{colorBest}{100.0}} & 315.15 & \textbf{\textcolor{colorBest}{100.0$_{\pm 0.00}$}} & 432.29$_{\pm 121.97}$ & 824133.9 \\
\hline
PruneForget     & \textbf{\textcolor{colorBest}{100.0}} & 12.95 & \textbf{\textcolor{colorBest}{100.0}} & 13.68 & \textbf{\textcolor{colorBest}{100.0}} & 13.40 & \textbf{\textcolor{colorBest}{100.0}} & 13.48 & \textbf{\textcolor{colorBest}{100.0}} & 13.84 & \textbf{\textcolor{colorBest}{100.0$_{\pm 0.00}$}} & 13.47$_{\pm 0.34}$ & 35765.0 \\
\specialrule{.15em}{.05em}{.05em}
\end{tabular}
}
\label{tab:dit_imagenet}
\end{table*}
In~\tabref{tab:dit_imagenet}, we present the class-wise unlearning results. Following~\citet{huang2024unified}, we omit the oracle (Retrain$\to$2SP) due to computational constraints and use the raw values of UA and FID for comparison. We observe that PruneForget effectively unlearns the target class (achieving 100 UA in all cases) while maintaining high generation fidelity on the retain classes (average FID is 13.47). In contrast, baseline methods either fail to forget the specified concept (UA $<10$) or suffer catastrophic degradation in their generative capabilities (FID $>100$).
\begin{table*}[h!]
\small
    \setlength{\tabcolsep}{3pt}
    \centering
\caption{\textbf{Ablation of early stopping on generation.} We evaluate the impact of the early stopping mechanism on DDPM for CIFAR-10 class-wise unlearning. PruneForget consistently maintains tighter $\Delta$UA and $\Delta$FID gaps relative to the \hl{oracle}.}
\resizebox{\textwidth}{!}{ 
\begin{tabular}{l|cc|cc|cc|cc|cc|c}
\specialrule{.15em}{.05em}{.05em}
    \multirow{2}{*}{\textbf{Method}} & \multicolumn{2}{c|}{\textbf{Automobile}} & \multicolumn{2}{c|}{\textbf{Cat}} & \multicolumn{2}{c|}{\textbf{Dog}} & \multicolumn{2}{c|}{\textbf{Horse}} & \multicolumn{2}{c|}{\textbf{Truck}} & \multirow{2}{*}{\textbf{Time}} \\
 & \textbf{$\Delta$UA $\downarrow$} & \textbf{$\Delta$FID $\downarrow$} & \textbf{$\Delta$UA $\downarrow$} & \textbf{$\Delta$FID $\downarrow$} & \textbf{$\Delta$UA $\downarrow$} & \textbf{$\Delta$FID $\downarrow$} & \textbf{$\Delta$UA $\downarrow$} & \textbf{$\Delta$FID $\downarrow$} & \textbf{$\Delta$UA $\downarrow$} & \textbf{$\Delta$FID $\downarrow$} & \\
\hline
\cellcolor{colorOracle}Retrain$\to$2SP & \cellcolor{colorOracle} 0.00 (98.87) & \cellcolor{colorOracle} 0.00 (17.10) & \cellcolor{colorOracle} 0.00 (95.50) & \cellcolor{colorOracle} 0.00 (15.96) & \cellcolor{colorOracle} 0.00 (98.79) & \cellcolor{colorOracle} 0.00 (16.26) & \cellcolor{colorOracle} 0.00 (99.12) & \cellcolor{colorOracle} 0.00 (17.54) & \cellcolor{colorOracle} 0.00 (99.66) & \cellcolor{colorOracle} 0.00 (16.78) & \cellcolor{colorOracle} 398622 \\
w/o Early Stop & {{5.60}} (93.27) & {{3.21}} (20.31) & {{1.00}} (96.50) & {{6.64}} (22.60) & {{0.15}} (98.64) & {{4.82}} (21.08) & {{0.20}} (99.32) & {{8.41}} (25.95) & \textbf{\textcolor{colorBest}{0.04}} (99.62) & {{1.19}} (17.97) & {{26396}} \\
PruneForget & \textbf{\textcolor{colorBest}{0.17}} (98.70) & \textbf{\textcolor{colorBest}{0.38}} (17.48) & \textbf{\textcolor{colorBest}{0.13}} (95.37) & \textbf{\textcolor{colorBest}{0.13}} (15.83) & \textbf{\textcolor{colorBest}{0.04}} (98.75) & \textbf{\textcolor{colorBest}{0.10}} (16.36) & \textbf{\textcolor{colorBest}{0.04}} (99.08) & \textbf{\textcolor{colorBest}{0.04}} (17.50) & {{0.26}} (99.40) & \textbf{\textcolor{colorBest}{0.63}} (17.41) & 26254 \\
\specialrule{.15em}{.05em}{.05em}
\end{tabular}
}
\label{tab:supp_early_stop_ddpm}
\end{table*}

\myparagraph{Ablation of early stopping.}
\tabref{tab:supp_early_stop_ddpm} shows the ablation of the early stopping mechanism mentioned in~\secref{sec:additional_details} for generative tasks, evaluating DDPM across five individual CIFAR-10 classes. The results highlight that without early stopping, the model is highly vulnerable to the destructive effects of an unconstrained unlearning loss. This aggressive unlearning catastrophically degrades overall image synthesis quality, evidenced by sharp increases in $\Delta$FID (\eg, from 0.13 to 6.64 for \textit{Cat} and from 0.04 to 8.41 for \textit{Horse}). This demonstrates that early stopping is important for diffusion unlearning, allowing the network to preserve high-fidelity generation capabilities while successfully eradicating targeted concepts.

\begin{figure*}[t]
\centering
\caption{\textbf{Qualitative results of CIFAR-10 (DDPM) class-wise unlearning.} We show generated samples for the unlearn classes (\textit{Automobile}, \textit{Cat}, \textit{Dog}, \textit{Horse}, and \textit{Truck}) alongside several retain classes. We observe that PruneForget successfully unlearns the class concept, \eg, when the \hlblue{pre-trained} model successfully generated a car, PruneForget and the \hl{oracle} instead generate an ambiguous object. Overall, we observe that PruneForget consistently generates images that closely match the \hl{oracle} (Retrain$\to$2SP). These observations are consistent with the quantitative metrics.
}
\resizebox{\textwidth}{!}{
\begin{tabular}{l|c*5{w{c}{0.06\textwidth}}|*9{w{c}{0.06\textwidth}}}
\specialrule{.15em}{.05em}{.05em}
\multirow{2}{*}{\textbf{Method}} & \multicolumn{6}{c|}{\textbf{Unlearn class}} & \multicolumn{9}{c}{\textbf{Retain classes}} \\ 
& & \textbf{I1} & \textbf{I2} & \textbf{I3} & \textbf{I4} & \textbf{I5} & \textbf{C1} & \textbf{C2} & \textbf{C3} & \textbf{C4} & \textbf{C5} & \textbf{C6} & \textbf{C7} & \textbf{C8} & \textbf{C9} \\
\hline

\cellcolor{colorPretrain}Pretrain & &
\includegraphics[width=0.056\textwidth, valign=c, margin=0pt 3pt 0pt 3pt]{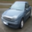} & \includegraphics[width=0.056\textwidth, valign=c, margin=0pt 3pt 0pt 3pt]{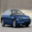} & \includegraphics[width=0.056\textwidth, valign=c, margin=0pt 3pt 0pt 3pt]{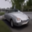} & \includegraphics[width=0.056\textwidth, valign=c, margin=0pt 3pt 0pt 3pt]{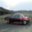} & \includegraphics[width=0.056\textwidth, valign=c, margin=0pt 3pt 0pt 3pt]{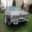} & 
\includegraphics[width=0.056\textwidth, valign=c, margin=0pt 3pt 0pt 3pt]{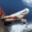} & \includegraphics[width=0.056\textwidth, valign=c, margin=0pt 3pt 0pt 3pt]{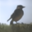} & \includegraphics[width=0.056\textwidth, valign=c, margin=0pt 3pt 0pt 3pt]{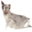} & \includegraphics[width=0.056\textwidth, valign=c, margin=0pt 3pt 0pt 3pt]{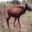} & \includegraphics[width=0.056\textwidth, valign=c, margin=0pt 3pt 0pt 3pt]{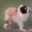} & \includegraphics[width=0.056\textwidth, valign=c, margin=0pt 3pt 0pt 3pt]{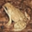} & \includegraphics[width=0.056\textwidth, valign=c, margin=0pt 3pt 0pt 3pt]{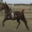} & \includegraphics[width=0.056\textwidth, valign=c, margin=0pt 3pt 0pt 3pt]{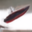} & \includegraphics[width=0.056\textwidth, valign=c, margin=0pt 3pt 0pt 3pt]{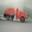} \\

\cellcolor{colorOracle}Retrain$\to$2SP & \smash{\rotatebox[origin=c]{90}{Automobile}} & 
\includegraphics[width=0.056\textwidth, valign=c, margin=0pt 3pt 0pt 3pt]{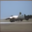} & \includegraphics[width=0.056\textwidth, valign=c, margin=0pt 3pt 0pt 3pt]{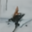} & \includegraphics[width=0.056\textwidth, valign=c, margin=0pt 3pt 0pt 3pt]{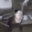} & \includegraphics[width=0.056\textwidth, valign=c, margin=0pt 3pt 0pt 3pt]{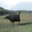} & \includegraphics[width=0.056\textwidth, valign=c, margin=0pt 3pt 0pt 3pt]{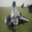} & 
\includegraphics[width=0.056\textwidth, valign=c, margin=0pt 3pt 0pt 3pt]{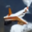} & \includegraphics[width=0.056\textwidth, valign=c, margin=0pt 3pt 0pt 3pt]{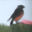} & \includegraphics[width=0.056\textwidth, valign=c, margin=0pt 3pt 0pt 3pt]{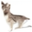} & \includegraphics[width=0.056\textwidth, valign=c, margin=0pt 3pt 0pt 3pt]{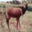} & \includegraphics[width=0.056\textwidth, valign=c, margin=0pt 3pt 0pt 3pt]{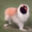} & \includegraphics[width=0.056\textwidth, valign=c, margin=0pt 3pt 0pt 3pt]{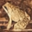} & \includegraphics[width=0.056\textwidth, valign=c, margin=0pt 3pt 0pt 3pt]{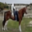} & \includegraphics[width=0.056\textwidth, valign=c, margin=0pt 3pt 0pt 3pt]{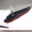} & \includegraphics[width=0.056\textwidth, valign=c, margin=0pt 3pt 0pt 3pt]{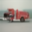} \\

PruneForget & & 
\includegraphics[width=0.056\textwidth, valign=c, margin=0pt 3pt 0pt 3pt]{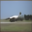} & \includegraphics[width=0.056\textwidth, valign=c, margin=0pt 3pt 0pt 3pt]{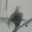} & \includegraphics[width=0.056\textwidth, valign=c, margin=0pt 3pt 0pt 3pt]{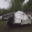} & \includegraphics[width=0.056\textwidth, valign=c, margin=0pt 3pt 0pt 3pt]{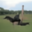} & \includegraphics[width=0.056\textwidth, valign=c, margin=0pt 3pt 0pt 3pt]{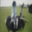} & 
\includegraphics[width=0.056\textwidth, valign=c, margin=0pt 3pt 0pt 3pt]{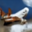} & \includegraphics[width=0.056\textwidth, valign=c, margin=0pt 3pt 0pt 3pt]{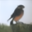} & \includegraphics[width=0.056\textwidth, valign=c, margin=0pt 3pt 0pt 3pt]{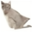} & \includegraphics[width=0.056\textwidth, valign=c, margin=0pt 3pt 0pt 3pt]{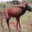} & \includegraphics[width=0.056\textwidth, valign=c, margin=0pt 3pt 0pt 3pt]{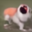} & \includegraphics[width=0.056\textwidth, valign=c, margin=0pt 3pt 0pt 3pt]{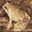} & \includegraphics[width=0.056\textwidth, valign=c, margin=0pt 3pt 0pt 3pt]{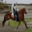} & \includegraphics[width=0.056\textwidth, valign=c, margin=0pt 3pt 0pt 3pt]{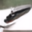} & \includegraphics[width=0.056\textwidth, valign=c, margin=0pt 3pt 0pt 3pt]{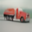} \\

\hline

\cellcolor{colorPretrain}Pretrain & & 
\includegraphics[width=0.056\textwidth, valign=c, margin=0pt 3pt 0pt 3pt]{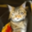} & \includegraphics[width=0.056\textwidth, valign=c, margin=0pt 3pt 0pt 3pt]{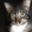} & \includegraphics[width=0.056\textwidth, valign=c, margin=0pt 3pt 0pt 3pt]{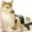} & \includegraphics[width=0.056\textwidth, valign=c, margin=0pt 3pt 0pt 3pt]{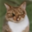} & \includegraphics[width=0.056\textwidth, valign=c, margin=0pt 3pt 0pt 3pt]{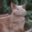} & 
\includegraphics[width=0.056\textwidth, valign=c, margin=0pt 3pt 0pt 3pt]{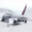} & \includegraphics[width=0.056\textwidth, valign=c, margin=0pt 3pt 0pt 3pt]{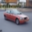} & \includegraphics[width=0.056\textwidth, valign=c, margin=0pt 3pt 0pt 3pt]{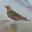} & \includegraphics[width=0.056\textwidth, valign=c, margin=0pt 3pt 0pt 3pt]{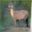} & \includegraphics[width=0.056\textwidth, valign=c, margin=0pt 3pt 0pt 3pt]{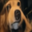} & \includegraphics[width=0.056\textwidth, valign=c, margin=0pt 3pt 0pt 3pt]{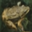} & \includegraphics[width=0.056\textwidth, valign=c, margin=0pt 3pt 0pt 3pt]{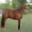} & \includegraphics[width=0.056\textwidth, valign=c, margin=0pt 3pt 0pt 3pt]{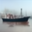} & \includegraphics[width=0.056\textwidth, valign=c, margin=0pt 3pt 0pt 3pt]{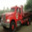} \\

\cellcolor{colorOracle}Retrain$\to$2SP & \smash{\rotatebox[origin=c]{90}{Cat}} & 
\includegraphics[width=0.056\textwidth, valign=c, margin=0pt 3pt 0pt 3pt]{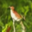} & \includegraphics[width=0.056\textwidth, valign=c, margin=0pt 3pt 0pt 3pt]{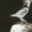} & \includegraphics[width=0.056\textwidth, valign=c, margin=0pt 3pt 0pt 3pt]{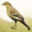} & \includegraphics[width=0.056\textwidth, valign=c, margin=0pt 3pt 0pt 3pt]{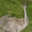} & \includegraphics[width=0.056\textwidth, valign=c, margin=0pt 3pt 0pt 3pt]{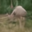} & 
\includegraphics[width=0.056\textwidth, valign=c, margin=0pt 3pt 0pt 3pt]{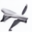} & \includegraphics[width=0.056\textwidth, valign=c, margin=0pt 3pt 0pt 3pt]{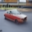} & \includegraphics[width=0.056\textwidth, valign=c, margin=0pt 3pt 0pt 3pt]{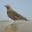} & \includegraphics[width=0.056\textwidth, valign=c, margin=0pt 3pt 0pt 3pt]{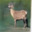} & \includegraphics[width=0.056\textwidth, valign=c, margin=0pt 3pt 0pt 3pt]{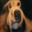} & \includegraphics[width=0.056\textwidth, valign=c, margin=0pt 3pt 0pt 3pt]{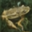} & \includegraphics[width=0.056\textwidth, valign=c, margin=0pt 3pt 0pt 3pt]{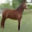} & \includegraphics[width=0.056\textwidth, valign=c, margin=0pt 3pt 0pt 3pt]{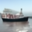} & \includegraphics[width=0.056\textwidth, valign=c, margin=0pt 3pt 0pt 3pt]{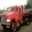} \\

PruneForget & & 
\includegraphics[width=0.056\textwidth, valign=c, margin=0pt 3pt 0pt 3pt]{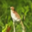} & \includegraphics[width=0.056\textwidth, valign=c, margin=0pt 3pt 0pt 3pt]{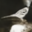} & \includegraphics[width=0.056\textwidth, valign=c, margin=0pt 3pt 0pt 3pt]{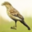} & \includegraphics[width=0.056\textwidth, valign=c, margin=0pt 3pt 0pt 3pt]{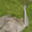} & \includegraphics[width=0.056\textwidth, valign=c, margin=0pt 3pt 0pt 3pt]{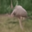} & 
\includegraphics[width=0.056\textwidth, valign=c, margin=0pt 3pt 0pt 3pt]{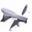} & \includegraphics[width=0.056\textwidth, valign=c, margin=0pt 3pt 0pt 3pt]{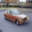} & \includegraphics[width=0.056\textwidth, valign=c, margin=0pt 3pt 0pt 3pt]{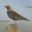} & \includegraphics[width=0.056\textwidth, valign=c, margin=0pt 3pt 0pt 3pt]{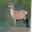} & \includegraphics[width=0.056\textwidth, valign=c, margin=0pt 3pt 0pt 3pt]{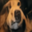} & \includegraphics[width=0.056\textwidth, valign=c, margin=0pt 3pt 0pt 3pt]{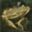} & \includegraphics[width=0.056\textwidth, valign=c, margin=0pt 3pt 0pt 3pt]{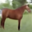} & \includegraphics[width=0.056\textwidth, valign=c, margin=0pt 3pt 0pt 3pt]{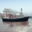} & \includegraphics[width=0.056\textwidth, valign=c, margin=0pt 3pt 0pt 3pt]{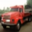} \\

\hline

\cellcolor{colorPretrain}Pretrain & & 
\includegraphics[width=0.056\textwidth, valign=c, margin=0pt 3pt 0pt 3pt]{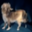} & \includegraphics[width=0.056\textwidth, valign=c, margin=0pt 3pt 0pt 3pt]{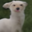} & \includegraphics[width=0.056\textwidth, valign=c, margin=0pt 3pt 0pt 3pt]{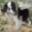} & \includegraphics[width=0.056\textwidth, valign=c, margin=0pt 3pt 0pt 3pt]{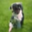} & \includegraphics[width=0.056\textwidth, valign=c, margin=0pt 3pt 0pt 3pt]{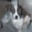} & 
\includegraphics[width=0.056\textwidth, valign=c, margin=0pt 3pt 0pt 3pt]{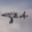} & \includegraphics[width=0.056\textwidth, valign=c, margin=0pt 3pt 0pt 3pt]{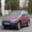} & \includegraphics[width=0.056\textwidth, valign=c, margin=0pt 3pt 0pt 3pt]{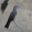} & \includegraphics[width=0.056\textwidth, valign=c, margin=0pt 3pt 0pt 3pt]{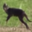} & \includegraphics[width=0.056\textwidth, valign=c, margin=0pt 3pt 0pt 3pt]{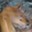} & \includegraphics[width=0.056\textwidth, valign=c, margin=0pt 3pt 0pt 3pt]{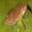} & \includegraphics[width=0.056\textwidth, valign=c, margin=0pt 3pt 0pt 3pt]{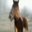} & \includegraphics[width=0.056\textwidth, valign=c, margin=0pt 3pt 0pt 3pt]{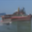} & \includegraphics[width=0.056\textwidth, valign=c, margin=0pt 3pt 0pt 3pt]{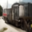} \\

\cellcolor{colorOracle}Retrain$\to$2SP & \smash{\rotatebox[origin=c]{90}{Dog}} & 
\includegraphics[width=0.056\textwidth, valign=c, margin=0pt 3pt 0pt 3pt]{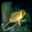} & \includegraphics[width=0.056\textwidth, valign=c, margin=0pt 3pt 0pt 3pt]{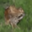} & \includegraphics[width=0.056\textwidth, valign=c, margin=0pt 3pt 0pt 3pt]{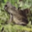} & \includegraphics[width=0.056\textwidth, valign=c, margin=0pt 3pt 0pt 3pt]{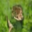} & \includegraphics[width=0.056\textwidth, valign=c, margin=0pt 3pt 0pt 3pt]{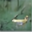} & 
\includegraphics[width=0.056\textwidth, valign=c, margin=0pt 3pt 0pt 3pt]{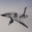} & \includegraphics[width=0.056\textwidth, valign=c, margin=0pt 3pt 0pt 3pt]{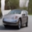} & \includegraphics[width=0.056\textwidth, valign=c, margin=0pt 3pt 0pt 3pt]{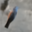} & \includegraphics[width=0.056\textwidth, valign=c, margin=0pt 3pt 0pt 3pt]{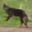} & \includegraphics[width=0.056\textwidth, valign=c, margin=0pt 3pt 0pt 3pt]{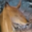} & \includegraphics[width=0.056\textwidth, valign=c, margin=0pt 3pt 0pt 3pt]{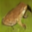} & \includegraphics[width=0.056\textwidth, valign=c, margin=0pt 3pt 0pt 3pt]{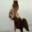} & \includegraphics[width=0.056\textwidth, valign=c, margin=0pt 3pt 0pt 3pt]{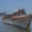} & \includegraphics[width=0.056\textwidth, valign=c, margin=0pt 3pt 0pt 3pt]{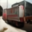} \\

PruneForget & & 
\includegraphics[width=0.056\textwidth, valign=c, margin=0pt 3pt 0pt 3pt]{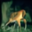} & \includegraphics[width=0.056\textwidth, valign=c, margin=0pt 3pt 0pt 3pt]{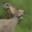} & \includegraphics[width=0.056\textwidth, valign=c, margin=0pt 3pt 0pt 3pt]{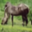} & \includegraphics[width=0.056\textwidth, valign=c, margin=0pt 3pt 0pt 3pt]{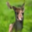} & \includegraphics[width=0.056\textwidth, valign=c, margin=0pt 3pt 0pt 3pt]{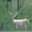} & 
\includegraphics[width=0.056\textwidth, valign=c, margin=0pt 3pt 0pt 3pt]{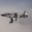} & \includegraphics[width=0.056\textwidth, valign=c, margin=0pt 3pt 0pt 3pt]{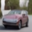} & \includegraphics[width=0.056\textwidth, valign=c, margin=0pt 3pt 0pt 3pt]{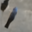} & \includegraphics[width=0.056\textwidth, valign=c, margin=0pt 3pt 0pt 3pt]{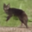} & \includegraphics[width=0.056\textwidth, valign=c, margin=0pt 3pt 0pt 3pt]{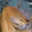} & \includegraphics[width=0.056\textwidth, valign=c, margin=0pt 3pt 0pt 3pt]{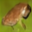} & \includegraphics[width=0.056\textwidth, valign=c, margin=0pt 3pt 0pt 3pt]{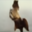} & \includegraphics[width=0.056\textwidth, valign=c, margin=0pt 3pt 0pt 3pt]{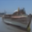} & \includegraphics[width=0.056\textwidth, valign=c, margin=0pt 3pt 0pt 3pt]{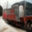} \\

\hline

\cellcolor{colorPretrain}Pretrain & & 
\includegraphics[width=0.056\textwidth, valign=c, margin=0pt 3pt 0pt 3pt]{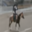} & \includegraphics[width=0.056\textwidth, valign=c, margin=0pt 3pt 0pt 3pt]{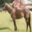} & \includegraphics[width=0.056\textwidth, valign=c, margin=0pt 3pt 0pt 3pt]{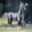} & \includegraphics[width=0.056\textwidth, valign=c, margin=0pt 3pt 0pt 3pt]{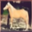} & \includegraphics[width=0.056\textwidth, valign=c, margin=0pt 3pt 0pt 3pt]{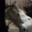} & 
\includegraphics[width=0.056\textwidth, valign=c, margin=0pt 3pt 0pt 3pt]{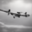} & \includegraphics[width=0.056\textwidth, valign=c, margin=0pt 3pt 0pt 3pt]{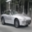} & \includegraphics[width=0.056\textwidth, valign=c, margin=0pt 3pt 0pt 3pt]{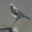} & \includegraphics[width=0.056\textwidth, valign=c, margin=0pt 3pt 0pt 3pt]{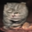} & \includegraphics[width=0.056\textwidth, valign=c, margin=0pt 3pt 0pt 3pt]{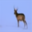} & \includegraphics[width=0.056\textwidth, valign=c, margin=0pt 3pt 0pt 3pt]{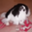} & \includegraphics[width=0.056\textwidth, valign=c, margin=0pt 3pt 0pt 3pt]{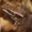} & \includegraphics[width=0.056\textwidth, valign=c, margin=0pt 3pt 0pt 3pt]{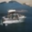} & \includegraphics[width=0.056\textwidth, valign=c, margin=0pt 3pt 0pt 3pt]{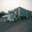} \\

\cellcolor{colorOracle}Retrain$\to$2SP & \smash{\rotatebox[origin=c]{90}{Horse}} & 
\includegraphics[width=0.056\textwidth, valign=c, margin=0pt 3pt 0pt 3pt]{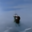} & \includegraphics[width=0.056\textwidth, valign=c, margin=0pt 3pt 0pt 3pt]{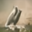} & \includegraphics[width=0.056\textwidth, valign=c, margin=0pt 3pt 0pt 3pt]{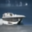} & \includegraphics[width=0.056\textwidth, valign=c, margin=0pt 3pt 0pt 3pt]{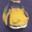} & \includegraphics[width=0.056\textwidth, valign=c, margin=0pt 3pt 0pt 3pt]{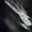} & 
\includegraphics[width=0.056\textwidth, valign=c, margin=0pt 3pt 0pt 3pt]{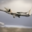} & \includegraphics[width=0.056\textwidth, valign=c, margin=0pt 3pt 0pt 3pt]{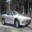} & \includegraphics[width=0.056\textwidth, valign=c, margin=0pt 3pt 0pt 3pt]{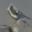} & \includegraphics[width=0.056\textwidth, valign=c, margin=0pt 3pt 0pt 3pt]{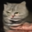} & \includegraphics[width=0.056\textwidth, valign=c, margin=0pt 3pt 0pt 3pt]{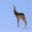} & \includegraphics[width=0.056\textwidth, valign=c, margin=0pt 3pt 0pt 3pt]{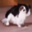} & \includegraphics[width=0.056\textwidth, valign=c, margin=0pt 3pt 0pt 3pt]{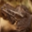} & \includegraphics[width=0.056\textwidth, valign=c, margin=0pt 3pt 0pt 3pt]{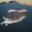} & \includegraphics[width=0.056\textwidth, valign=c, margin=0pt 3pt 0pt 3pt]{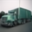} \\

PruneForget & & 
\includegraphics[width=0.056\textwidth, valign=c, margin=0pt 3pt 0pt 3pt]{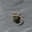} & \includegraphics[width=0.056\textwidth, valign=c, margin=0pt 3pt 0pt 3pt]{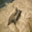} & \includegraphics[width=0.056\textwidth, valign=c, margin=0pt 3pt 0pt 3pt]{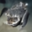} & \includegraphics[width=0.056\textwidth, valign=c, margin=0pt 3pt 0pt 3pt]{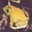} & \includegraphics[width=0.056\textwidth, valign=c, margin=0pt 3pt 0pt 3pt]{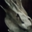} & 
\includegraphics[width=0.056\textwidth, valign=c, margin=0pt 3pt 0pt 3pt]{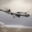} & \includegraphics[width=0.056\textwidth, valign=c, margin=0pt 3pt 0pt 3pt]{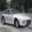} & \includegraphics[width=0.056\textwidth, valign=c, margin=0pt 3pt 0pt 3pt]{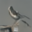} & \includegraphics[width=0.056\textwidth, valign=c, margin=0pt 3pt 0pt 3pt]{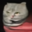} & \includegraphics[width=0.056\textwidth, valign=c, margin=0pt 3pt 0pt 3pt]{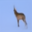} & \includegraphics[width=0.056\textwidth, valign=c, margin=0pt 3pt 0pt 3pt]{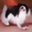} & \includegraphics[width=0.056\textwidth, valign=c, margin=0pt 3pt 0pt 3pt]{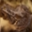} & \includegraphics[width=0.056\textwidth, valign=c, margin=0pt 3pt 0pt 3pt]{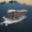} & \includegraphics[width=0.056\textwidth, valign=c, margin=0pt 3pt 0pt 3pt]{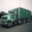} \\

\hline

\cellcolor{colorPretrain}Pretrain & & 
\includegraphics[width=0.056\textwidth, valign=c, margin=0pt 3pt 0pt 3pt]{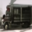} & \includegraphics[width=0.056\textwidth, valign=c, margin=0pt 3pt 0pt 3pt]{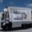} & \includegraphics[width=0.056\textwidth, valign=c, margin=0pt 3pt 0pt 3pt]{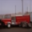} & \includegraphics[width=0.056\textwidth, valign=c, margin=0pt 3pt 0pt 3pt]{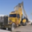} & \includegraphics[width=0.056\textwidth, valign=c, margin=0pt 3pt 0pt 3pt]{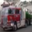} & 
\includegraphics[width=0.056\textwidth, valign=c, margin=0pt 3pt 0pt 3pt]{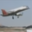} & \includegraphics[width=0.056\textwidth, valign=c, margin=0pt 3pt 0pt 3pt]{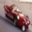} & \includegraphics[width=0.056\textwidth, valign=c, margin=0pt 3pt 0pt 3pt]{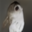} & \includegraphics[width=0.056\textwidth, valign=c, margin=0pt 3pt 0pt 3pt]{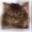} & \includegraphics[width=0.056\textwidth, valign=c, margin=0pt 3pt 0pt 3pt]{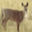} & \includegraphics[width=0.056\textwidth, valign=c, margin=0pt 3pt 0pt 3pt]{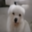} & \includegraphics[width=0.056\textwidth, valign=c, margin=0pt 3pt 0pt 3pt]{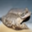} & \includegraphics[width=0.056\textwidth, valign=c, margin=0pt 3pt 0pt 3pt]{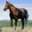} & \includegraphics[width=0.056\textwidth, valign=c, margin=0pt 3pt 0pt 3pt]{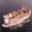} \\

\cellcolor{colorOracle}Retrain$\to$2SP & \smash{\rotatebox[origin=c]{90}{Truck}} & 
\includegraphics[width=0.056\textwidth, valign=c, margin=0pt 3pt 0pt 3pt]{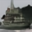} & \includegraphics[width=0.056\textwidth, valign=c, margin=0pt 3pt 0pt 3pt]{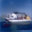} & \includegraphics[width=0.056\textwidth, valign=c, margin=0pt 3pt 0pt 3pt]{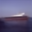} & \includegraphics[width=0.056\textwidth, valign=c, margin=0pt 3pt 0pt 3pt]{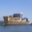} & \includegraphics[width=0.056\textwidth, valign=c, margin=0pt 3pt 0pt 3pt]{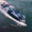} & 
\includegraphics[width=0.056\textwidth, valign=c, margin=0pt 3pt 0pt 3pt]{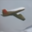} & \includegraphics[width=0.056\textwidth, valign=c, margin=0pt 3pt 0pt 3pt]{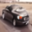} & \includegraphics[width=0.056\textwidth, valign=c, margin=0pt 3pt 0pt 3pt]{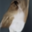} & \includegraphics[width=0.056\textwidth, valign=c, margin=0pt 3pt 0pt 3pt]{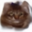} & \includegraphics[width=0.056\textwidth, valign=c, margin=0pt 3pt 0pt 3pt]{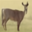} & \includegraphics[width=0.056\textwidth, valign=c, margin=0pt 3pt 0pt 3pt]{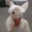} & \includegraphics[width=0.056\textwidth, valign=c, margin=0pt 3pt 0pt 3pt]{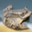} & \includegraphics[width=0.056\textwidth, valign=c, margin=0pt 3pt 0pt 3pt]{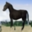} & \includegraphics[width=0.056\textwidth, valign=c, margin=0pt 3pt 0pt 3pt]{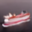} \\

PruneForget & & 
\includegraphics[width=0.056\textwidth, valign=c, margin=0pt 3pt 0pt 3pt]{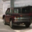} & \includegraphics[width=0.056\textwidth, valign=c, margin=0pt 3pt 0pt 3pt]{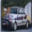} & \includegraphics[width=0.056\textwidth, valign=c, margin=0pt 3pt 0pt 3pt]{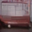} & \includegraphics[width=0.056\textwidth, valign=c, margin=0pt 3pt 0pt 3pt]{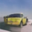} & \includegraphics[width=0.056\textwidth, valign=c, margin=0pt 3pt 0pt 3pt]{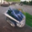} & 
\includegraphics[width=0.056\textwidth, valign=c, margin=0pt 3pt 0pt 3pt]{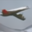} & \includegraphics[width=0.056\textwidth, valign=c, margin=0pt 3pt 0pt 3pt]{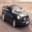} & \includegraphics[width=0.056\textwidth, valign=c, margin=0pt 3pt 0pt 3pt]{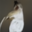} & \includegraphics[width=0.056\textwidth, valign=c, margin=0pt 3pt 0pt 3pt]{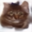} & \includegraphics[width=0.056\textwidth, valign=c, margin=0pt 3pt 0pt 3pt]{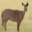} & \includegraphics[width=0.056\textwidth, valign=c, margin=0pt 3pt 0pt 3pt]{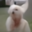} & \includegraphics[width=0.056\textwidth, valign=c, margin=0pt 3pt 0pt 3pt]{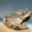} & \includegraphics[width=0.056\textwidth, valign=c, margin=0pt 3pt 0pt 3pt]{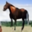} & \includegraphics[width=0.056\textwidth, valign=c, margin=0pt 3pt 0pt 3pt]{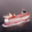} \\

\specialrule{.15em}{.05em}{.05em}
\end{tabular}
}
\label{fig:supp_qualitative_gen}
\end{figure*}

\myparagraph{Qualitative results on CIFAR-10 (DDPM).}
% \figref{fig:supp_qualitative_gen} presents additional qualitative results for class-wise unlearning of DDPM on CIFAR-10. While the pre-trained model generates recognizable images across all categories, both PruneForget and the oracle (Retrain$\to$2SP) entirely dismantle the generative capability for the targeted \textit{Horse} and \textit{Truck} categories. Specifically, for the unlearned classes (I1–I5), PruneForget's outputs lose their original semantics, either becoming ambiguous or shifting to resemble retained classes, closely mirroring the oracle's behavior. Conversely, image generation for the remaining categories (C1–C9) exhibits no loss in visual quality, confirming it can effectively and selectively unlearn class information while maintaining alignment with the oracle.
In~\figref{fig:supp_qualitative_gen}, we show the images generated from the unlearned models on the unlearned classes: \textit{Automobile}, \textit{Cat}, \textit{Dog}, \textit{Horse}, and \textit{Truck}. 
We observe that the pre-trained model generates recognizable images for all categories, while the oracle (Retrain$\rightarrow$2SP) and PruneForget successfully erase the forgotten concepts. Specifically, for the unlearn classes (columns I1–I5), the outputs from PruneForget lose their original class semantics. Observe that the images become ambiguous or belong to another class in the retain classes, which closely mirrors the behavior of the oracle. Next, for the retain classes (columns C1–C9), PruneForget preserves high visual fidelity, producing images that are very similar to those generated by the oracle model. These visual results show that PruneForget effectively unlearns the class information while maintaining alignment with the oracle.

\begin{figure*}[t]
\centering
\caption{\textbf{Qualitative results of ImageNet (DiT-XL/2) class-wise unlearning.} We show generated samples for the 5 evaluated unlearn classes (\textit{Cacatua galerita}, \textit{Golden retriever}, \textit{White wolf}, \textit{Arctic fox}, and \textit{Otter}) alongside several retain classes. PruneForget achieves perfect forgetting (by generating noise for the unlearn class) while maintaining generative quality in the retain classes (matching the \hlblue{pre-trained} model). We omit the \hl{oracle} due to computational constraints.}
\small
\resizebox{\textwidth}{!}{
\begin{tabular}{l|c*2{w{c}{0.11\textwidth}}|*5{w{c}{0.11\textwidth}}}
\specialrule{.15em}{.05em}{.05em}
\multirow{2}{*}{\textbf{Method}} & \multicolumn{3}{c|}{\textbf{Unlearn class}} & \multicolumn{5}{c}{\textbf{Retain classes}} \\ 
& & \textbf{I1} & \textbf{I2} & \textbf{C1} & \textbf{C2} & \textbf{C3} & \textbf{C4} & \textbf{C5} \\
\hline

% --- Cacatua galerita Section ---
\cellcolor{colorPretrain}Pretrain & & 
\includegraphics[width=0.106\textwidth, valign=c, margin=0pt 3pt 0pt 3pt]{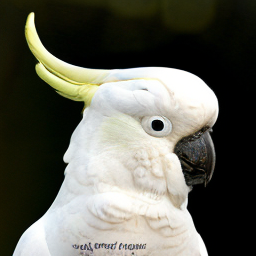} & 
\includegraphics[width=0.106\textwidth, valign=c, margin=0pt 3pt 0pt 3pt]{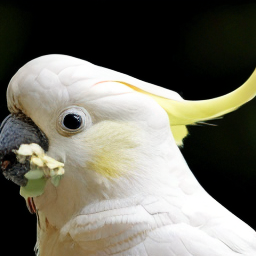} & 
\includegraphics[width=0.106\textwidth, valign=c, margin=0pt 3pt 0pt 3pt]{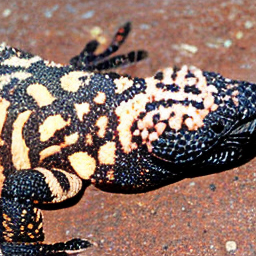} & 
\includegraphics[width=0.106\textwidth, valign=c, margin=0pt 3pt 0pt 3pt]{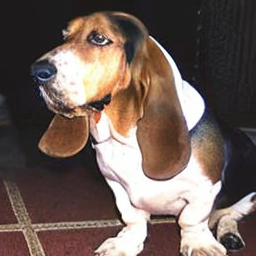} & 
\includegraphics[width=0.106\textwidth, valign=c, margin=0pt 3pt 0pt 3pt]{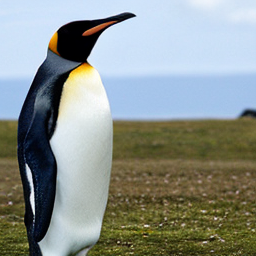} & 
\includegraphics[width=0.106\textwidth, valign=c, margin=0pt 3pt 0pt 3pt]{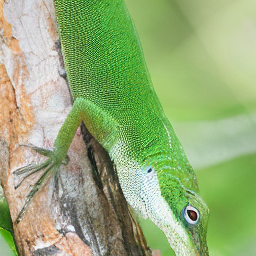} & 
\includegraphics[width=0.106\textwidth, valign=c, margin=0pt 3pt 0pt 3pt]{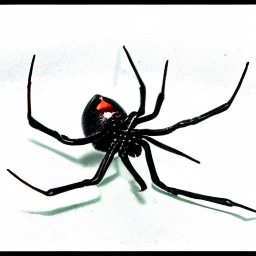} \\

PruneForget & \smash{\raisebox{-0.0\height}{\rotatebox{90}{\scriptsize Cacatua galerita}}} & 
\includegraphics[width=0.106\textwidth, valign=c, margin=0pt 3pt 0pt 3pt]{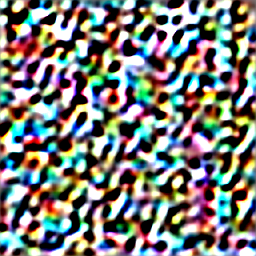} & 
\includegraphics[width=0.106\textwidth, valign=c, margin=0pt 3pt 0pt 3pt]{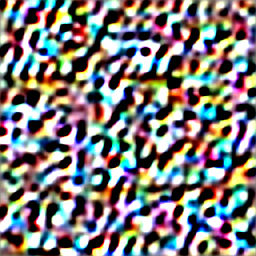} & 
\includegraphics[width=0.106\textwidth, valign=c, margin=0pt 3pt 0pt 3pt]{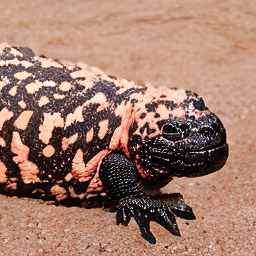} & 
\includegraphics[width=0.106\textwidth, valign=c, margin=0pt 3pt 0pt 3pt]{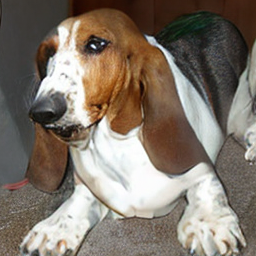} & 
\includegraphics[width=0.106\textwidth, valign=c, margin=0pt 3pt 0pt 3pt]{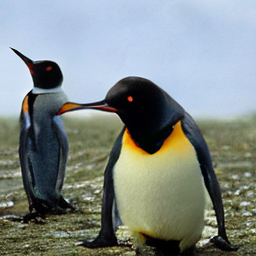} & 
\includegraphics[width=0.106\textwidth, valign=c, margin=0pt 3pt 0pt 3pt]{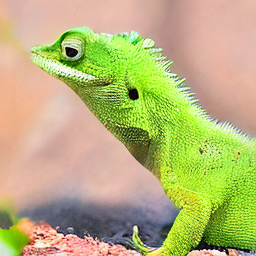} & 
\includegraphics[width=0.106\textwidth, valign=c, margin=0pt 3pt 0pt 3pt]{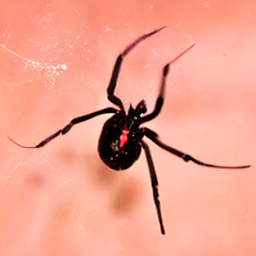} \\

\hline

% --- Golden Retriever Section ---
\cellcolor{colorPretrain}Pretrain & & 
\includegraphics[width=0.106\textwidth, valign=c, margin=0pt 3pt 0pt 3pt]{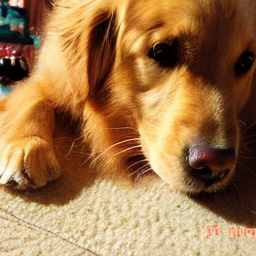} & 
\includegraphics[width=0.106\textwidth, valign=c, margin=0pt 3pt 0pt 3pt]{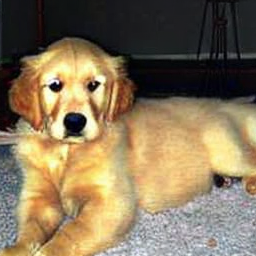} & 
\includegraphics[width=0.106\textwidth, valign=c, margin=0pt 3pt 0pt 3pt]{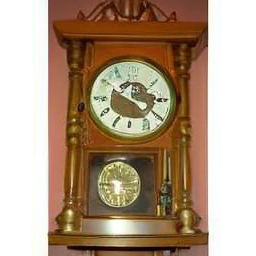} & 
\includegraphics[width=0.106\textwidth, valign=c, margin=0pt 3pt 0pt 3pt]{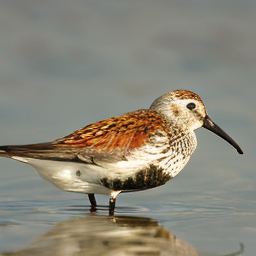} & 
\includegraphics[width=0.106\textwidth, valign=c, margin=0pt 3pt 0pt 3pt]{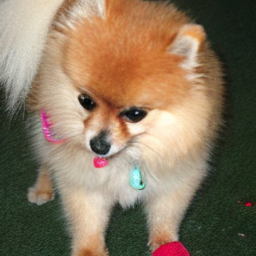} & 
\includegraphics[width=0.106\textwidth, valign=c, margin=0pt 3pt 0pt 3pt]{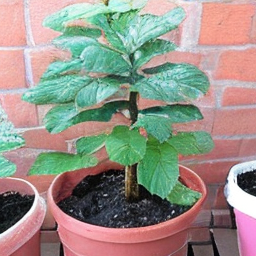} & 
\includegraphics[width=0.106\textwidth, valign=c, margin=0pt 3pt 0pt 3pt]{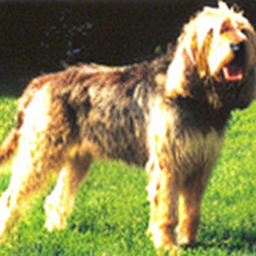} \\

PruneForget & \smash{\raisebox{-0.0\height}{\rotatebox{90}{\scriptsize Golden retriever}}} & 
\includegraphics[width=0.106\textwidth, valign=c, margin=0pt 3pt 0pt 3pt]{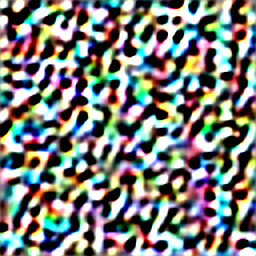} & 
\includegraphics[width=0.106\textwidth, valign=c, margin=0pt 3pt 0pt 3pt]{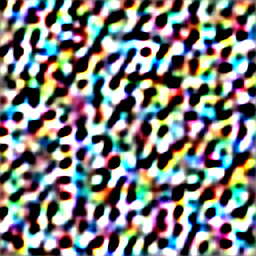} & 
\includegraphics[width=0.106\textwidth, valign=c, margin=0pt 3pt 0pt 3pt]{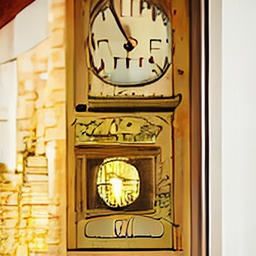} & 
\includegraphics[width=0.106\textwidth, valign=c, margin=0pt 3pt 0pt 3pt]{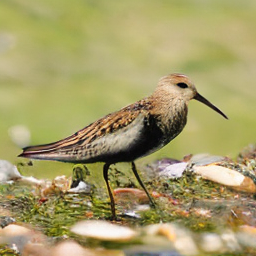} & 
\includegraphics[width=0.106\textwidth, valign=c, margin=0pt 3pt 0pt 3pt]{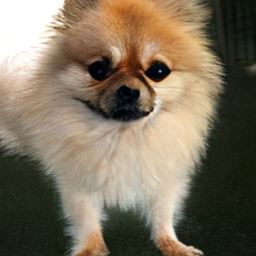} & 
\includegraphics[width=0.106\textwidth, valign=c, margin=0pt 3pt 0pt 3pt]{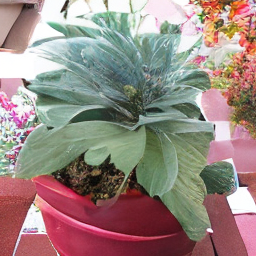} & 
\includegraphics[width=0.106\textwidth, valign=c, margin=0pt 3pt 0pt 3pt]{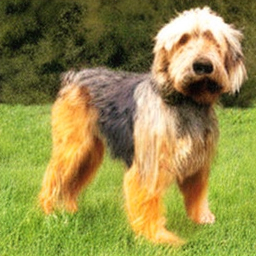} \\

\hline

% --- White Wolf Section ---
\cellcolor{colorPretrain}Pretrain & & 
\includegraphics[width=0.106\textwidth, valign=c, margin=0pt 3pt 0pt 3pt]{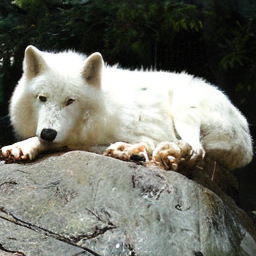} & 
\includegraphics[width=0.106\textwidth, valign=c, margin=0pt 3pt 0pt 3pt]{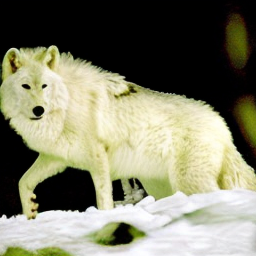} & 
\includegraphics[width=0.106\textwidth, valign=c, margin=0pt 3pt 0pt 3pt]{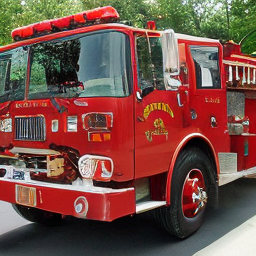} & 
\includegraphics[width=0.106\textwidth, valign=c, margin=0pt 3pt 0pt 3pt]{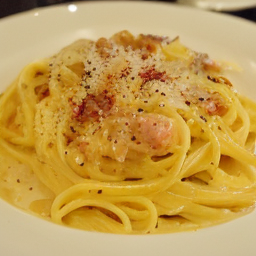} & 
\includegraphics[width=0.106\textwidth, valign=c, margin=0pt 3pt 0pt 3pt]{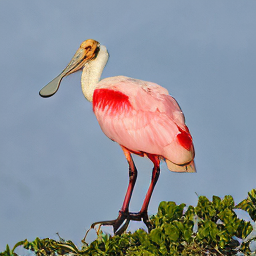} & 
\includegraphics[width=0.106\textwidth, valign=c, margin=0pt 3pt 0pt 3pt]{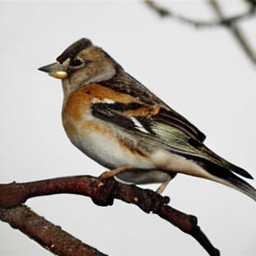} & 
\includegraphics[width=0.106\textwidth, valign=c, margin=0pt 3pt 0pt 3pt]{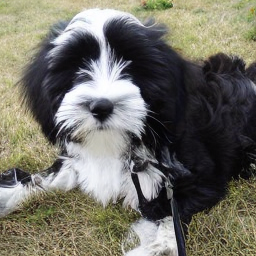} \\

PruneForget & \smash{\raisebox{+0.3\height}{\rotatebox{90}{\scriptsize White wolf}}} & 
\includegraphics[width=0.106\textwidth, valign=c, margin=0pt 3pt 0pt 3pt]{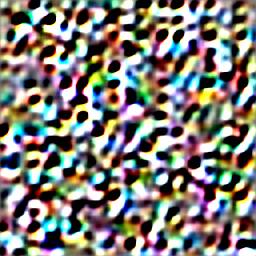} & 
\includegraphics[width=0.106\textwidth, valign=c, margin=0pt 3pt 0pt 3pt]{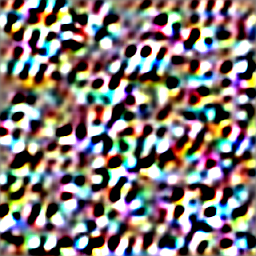} & 
\includegraphics[width=0.106\textwidth, valign=c, margin=0pt 3pt 0pt 3pt]{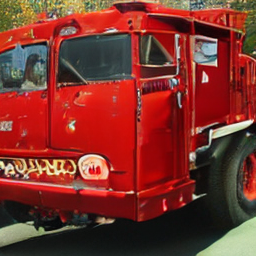} & 
\includegraphics[width=0.106\textwidth, valign=c, margin=0pt 3pt 0pt 3pt]{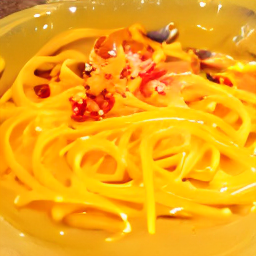} & 
\includegraphics[width=0.106\textwidth, valign=c, margin=0pt 3pt 0pt 3pt]{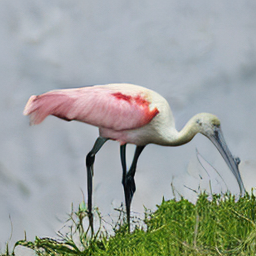} & 
\includegraphics[width=0.106\textwidth, valign=c, margin=0pt 3pt 0pt 3pt]{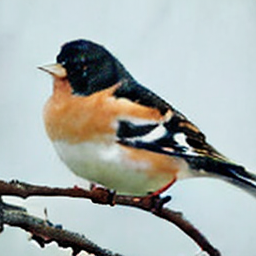} & 
\includegraphics[width=0.106\textwidth, valign=c, margin=0pt 3pt 0pt 3pt]{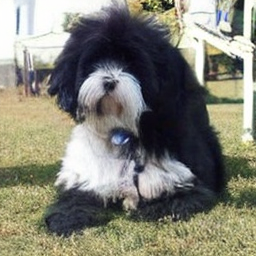} \\

\hline

% --- Arctic Fox Section ---
\cellcolor{colorPretrain}Pretrain & & 
\includegraphics[width=0.106\textwidth, valign=c, margin=0pt 3pt 0pt 3pt]{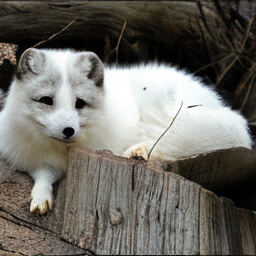} & 
\includegraphics[width=0.106\textwidth, valign=c, margin=0pt 3pt 0pt 3pt]{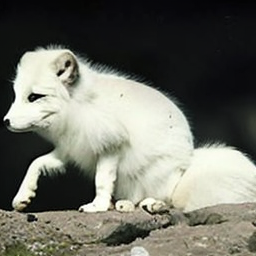} & 
\includegraphics[width=0.106\textwidth, valign=c, margin=0pt 3pt 0pt 3pt]{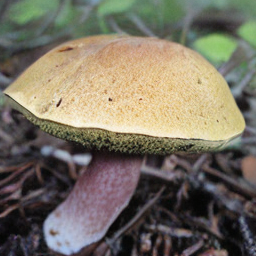} & 
\includegraphics[width=0.106\textwidth, valign=c, margin=0pt 3pt 0pt 3pt]{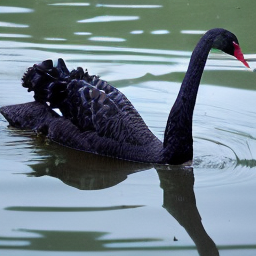} & 
\includegraphics[width=0.106\textwidth, valign=c, margin=0pt 3pt 0pt 3pt]{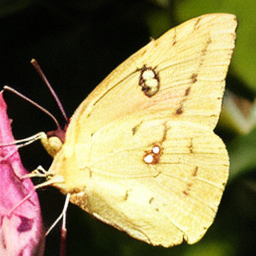} & 
\includegraphics[width=0.106\textwidth, valign=c, margin=0pt 3pt 0pt 3pt]{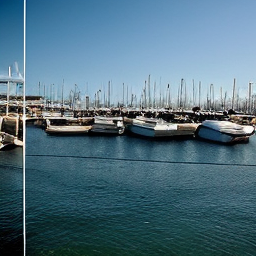} & 
\includegraphics[width=0.106\textwidth, valign=c, margin=0pt 3pt 0pt 3pt]{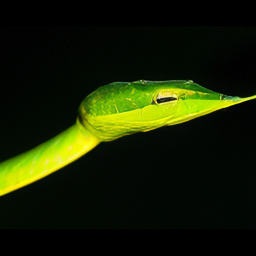} \\

PruneForget & \smash{\raisebox{+0.3\height}{\rotatebox{90}{\scriptsize Arctic fox}}} & 
\includegraphics[width=0.106\textwidth, valign=c, margin=0pt 3pt 0pt 3pt]{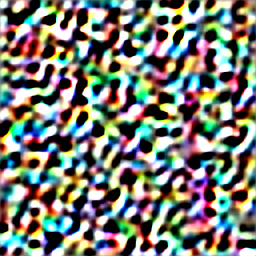} & 
\includegraphics[width=0.106\textwidth, valign=c, margin=0pt 3pt 0pt 3pt]{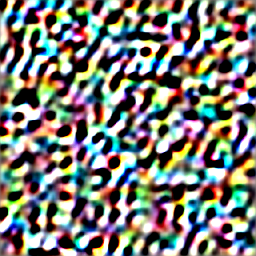} & 
\includegraphics[width=0.106\textwidth, valign=c, margin=0pt 3pt 0pt 3pt]{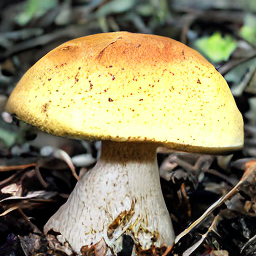} & 
\includegraphics[width=0.106\textwidth, valign=c, margin=0pt 3pt 0pt 3pt]{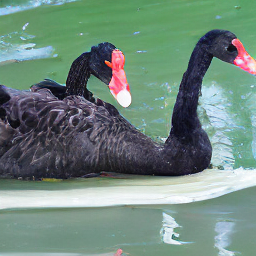} & 
\includegraphics[width=0.106\textwidth, valign=c, margin=0pt 3pt 0pt 3pt]{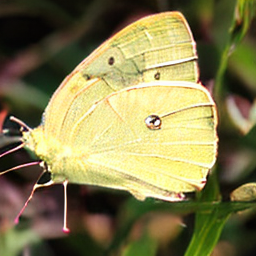} & 
\includegraphics[width=0.106\textwidth, valign=c, margin=0pt 3pt 0pt 3pt]{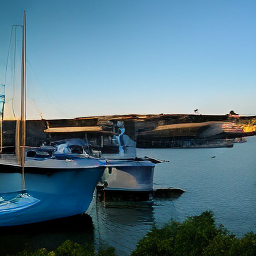} & 
\includegraphics[width=0.106\textwidth, valign=c, margin=0pt 3pt 0pt 3pt]{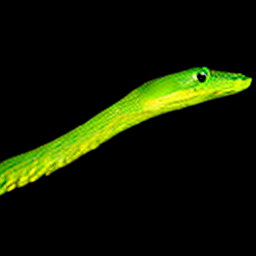} \\

\hline

% --- Otter Section ---
\cellcolor{colorPretrain}Pretrain & & 
\includegraphics[width=0.106\textwidth, valign=c, margin=0pt 3pt 0pt 3pt]{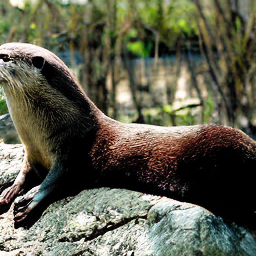} & 
\includegraphics[width=0.106\textwidth, valign=c, margin=0pt 3pt 0pt 3pt]{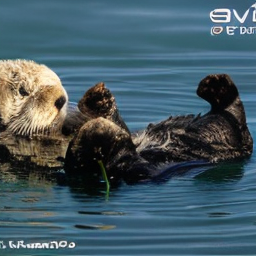} & 
\includegraphics[width=0.106\textwidth, valign=c, margin=0pt 3pt 0pt 3pt]{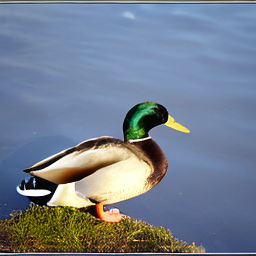} & 
\includegraphics[width=0.106\textwidth, valign=c, margin=0pt 3pt 0pt 3pt]{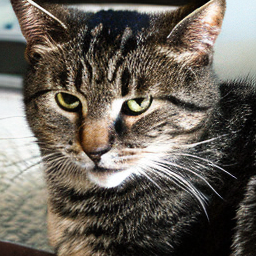} & 
\includegraphics[width=0.106\textwidth, valign=c, margin=0pt 3pt 0pt 3pt]{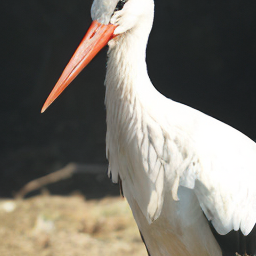} & 
\includegraphics[width=0.106\textwidth, valign=c, margin=0pt 3pt 0pt 3pt]{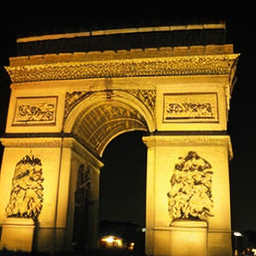} & 
\includegraphics[width=0.106\textwidth, valign=c, margin=0pt 3pt 0pt 3pt]{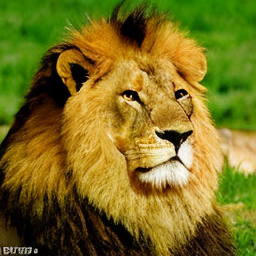} \\

PruneForget & \smash{\raisebox{+1.0\height}{\rotatebox{90}{\scriptsize Otter}}} & 
\includegraphics[width=0.106\textwidth, valign=c, margin=0pt 3pt 0pt 3pt]{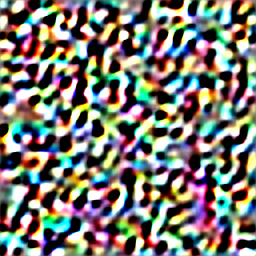} & 
\includegraphics[width=0.106\textwidth, valign=c, margin=0pt 3pt 0pt 3pt]{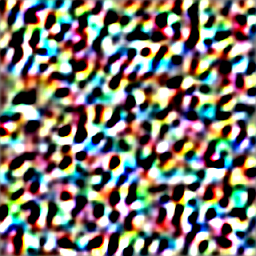} & 
\includegraphics[width=0.106\textwidth, valign=c, margin=0pt 3pt 0pt 3pt]{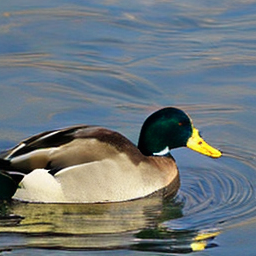} & 
\includegraphics[width=0.106\textwidth, valign=c, margin=0pt 3pt 0pt 3pt]{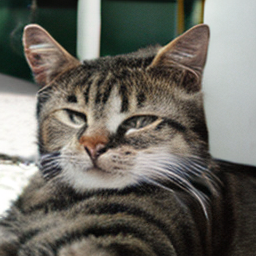} & 
\includegraphics[width=0.106\textwidth, valign=c, margin=0pt 3pt 0pt 3pt]{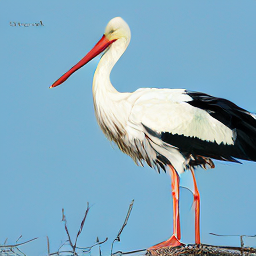} & 
\includegraphics[width=0.106\textwidth, valign=c, margin=0pt 3pt 0pt 3pt]{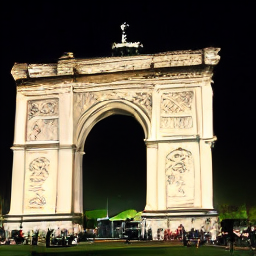} & 
\includegraphics[width=0.106\textwidth, valign=c, margin=0pt 3pt 0pt 3pt]{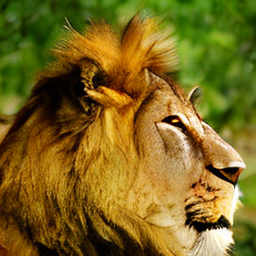} \\

\specialrule{.15em}{.05em}{.05em}
\end{tabular}
}
\label{fig:supp_dit_imgnet}
\end{figure*}
\myparagraph{Qualitative results on ImageNet (DiT-XL/2).}
In~\figref{fig:supp_dit_imgnet}, we present images generated by the unlearned models for the five evaluated unlearn classes: \textit{Cacatua galerita}, \textit{Golden retriever}, \textit{White wolf}, \textit{Arctic fox}, and \textit{Otter}. While the pre-trained model generates recognizable images for all categories, PruneForget successfully erases the forgotten concepts. Specifically, for the unlearn classes (columns I1–I2), the outputs from PruneForget degrade into pure noise. Conversely, for the retain classes (columns C1–C5), PruneForget preserves high visual fidelity, producing images that closely match those generated by the pre-trained model. %These visual results show that PruneForget effectively unlearns specific class information while maintaining the model's overall generative capabilities.

\begin{table}[t]
    \setlength{\tabcolsep}{3pt}
    \renewcommand{\arraystretch}{1.15}
\centering
\caption{\textbf{CIFAR-10 (ResNet-18) under global pruning.} Evaluating PruneForget under global pruning on 50\% random unlearning demonstrates superior alignment to the oracle (\hlavg{\textbf{Avg $\Delta$}}) and higher speedup compared to the unlearning-unaware baseline.}
\resizebox{\columnwidth}{!}{
\begin{tabular}{l|ccc | >{\columncolor{avg_highlight}}c | cc | cc}
\specialrule{.15em}{.05em}{.05em}
\multirow{2}{*}{\textbf{Method}} & \multicolumn{8}{c}{\textbf{Random Unlearn 50\%}} \\
 & \textbf{$\Delta$UA $\downarrow$} & \textbf{$\Delta$RA $\downarrow$} & \textbf{$\Delta$TA $\downarrow$} & \textbf{Avg $\Delta$ $\downarrow$} & \textbf{$D_{\text{KL}}$ $\downarrow$} & \textbf{Time} & \textbf{MACs} & \textbf{Speedup} \\
\hline
\cellcolor{colorOracle}Retrain$\to$3SP & \cellcolor{colorOracle} 0.00 (7.21$_{\pm 0.25}$) & \cellcolor{colorOracle} 0.00 (100.00$_{\pm 0.00}$) & \cellcolor{colorOracle} 0.00 (92.54$_{\pm 0.29}$) & \cellcolor{colorOracle} 0.00 & \cellcolor{colorOracle} 0.00 & \cellcolor{colorOracle} 1471 & \cellcolor{colorOracle} 175.5M & \cellcolor{colorOracle} 3.17$\times$ \\
Pretrain$\to$3SP & 1.79 (5.42$_{\pm 0.14}$) & \textbf{\textcolor{colorBest}{0.00}} (100.00$_{\pm 0.00}$) & 0.99 (93.53$_{\pm 0.16}$) & 0.93 & \textbf{\textcolor{colorBest}{0.23}} & 596 & 186.9M & 2.98$\times$ \\
% 3SP$\to$SFRon & 5.15 (12.36$_{\pm 0.24}$) & 0.40 (99.60$_{\pm 0.07}$) & 2.67 (89.87$_{\pm 0.22}$) & 2.74 & 0.25 & 681 & 229.7M & 2.42$\times$ \\
% SFRon$\to$3SP & \textit{\textbf{0.45}} (6.76$_{\pm 0.18}$) & \textbf{\textcolor{colorBest}{0.00}} (100.00$_{\pm 0.00}$) & 0.18 (92.72$_{\pm 0.22}$) & \textit{\textbf{0.21}} & 0.25 & 689 & 244.6M & 2.28$\times$ \\
% $\ell_1$-MU$\to$2SP & 5.20 (12.41$_{\pm 0.22}$) & 3.25 (96.75$_{\pm 0.32}$) & 5.17 (87.37$_{\pm 0.23}$) & 4.54 & 0.35 & 452 & \textbf{\textcolor{colorBest}{106.5M}} & \textbf{\textcolor{colorBest}{5.23$\times$}} \\
% Un-pruning & 3.48 (10.68$_{\pm 0.36}$) & \textit{\textbf{0.09}} (99.91$_{\pm 0.03}$) & 3.63 (88.91$_{\pm 0.39}$) & 2.40 & 0.50 & 518 & \textit{\textbf{127.0M}} & \textit{\textbf{4.38$\times$}} \\
% Bilevel & 6.71 (13.91$_{\pm 1.30}$) & 3.03 (96.97$_{\pm 1.46}$) & 6.48 (86.06$_{\pm 1.24}$) & 5.41 & 0.64 & 708 & 229.7M & 2.42$\times$ \\
% LLM-Eraser & 11.49 (18.70$_{\pm 2.34}$) & 14.49 (85.51$_{\pm 2.66}$) & 11.51 (81.03$_{\pm 2.39}$) & 12.50 & 0.42 & 95 & 219.3M & 2.54$\times$ \\
\hline
PruneForget & \textbf{\textcolor{colorBest}{0.05}} (7.26$_{\pm 0.29}$) & \textbf{\textcolor{colorBest}{0.00}} (100.00$_{\pm 0.00}$) & \textbf{\textcolor{colorBest}{0.12}} (92.42$_{\pm 0.29}$) & \textbf{\textcolor{colorBest}{0.06}} & 0.24 & 648 & 137.4M & \textbf{\textcolor{colorBest}{4.05}}$\times$ \\
\specialrule{.15em}{.05em}{.05em}
\end{tabular}
}
\label{tab:cls_cifar_rand_gp}
\end{table}

\begin{table}[t]
    \setlength{\tabcolsep}{3pt}
    \renewcommand{\arraystretch}{1.15}
\centering
\caption{\textbf{TinyImageNet (Swin-T) under global pruning.} Evaluating PruneForget under global pruning on 50\% random unlearning shows lower \hlavg{\textbf{Avg $\Delta$}} than the baseline under the same speedup.}
\resizebox{\columnwidth}{!}{
\begin{tabular}{l|ccc | >{\columncolor{avg_highlight}}c | cc | cc}
\specialrule{.15em}{.05em}{.05em}
\multirow{2}{*}{\textbf{Method}} & \multicolumn{8}{c}{\textbf{Random Unlearn 50\%}} \\
 & \textbf{$\Delta$UA $\downarrow$} & \textbf{$\Delta$RA $\downarrow$} & \textbf{$\Delta$TA $\downarrow$} & \textbf{Avg $\Delta$ $\downarrow$} & \textbf{$D_{\text{KL}}$ $\downarrow$} & \textbf{Time} & \textbf{MACs} & \textbf{Speedup} \\
\hline
\cellcolor{colorOracle}Retrain$\to$2SP & \cellcolor{colorOracle} 0.00 (22.95$_{\pm 0.18}$) & \cellcolor{colorOracle} 0.00 (98.68$_{\pm 0.07}$) & \cellcolor{colorOracle} 0.00 (77.29$_{\pm 0.21}$) & \cellcolor{colorOracle} 0.00 & \cellcolor{colorOracle} 0.00 & \cellcolor{colorOracle} 5429 & \cellcolor{colorOracle} 2962.6M & \cellcolor{colorOracle} 1.50$\times$ \\
Pretrain$\to$2SP & 2.37 (20.58$_{\pm 0.14}$) & 
\textbf{\textcolor{colorBest}{0.22}} (98.90$_{\pm 0.05}$) & 0.27 (77.56$_{\pm 0.26}$) & 0.95 & \textbf{\textcolor{colorBest}{0.31}} & 3663 & 2960.4M & 1.51$\times$ \\
% 2SP$\to$SFRon & 2.83 (25.79$_{\pm 0.37}$) & \textbf{\textcolor{colorBest}{0.06}} (98.61$_{\pm 0.08}$) & 4.24 (73.05$_{\pm 0.36}$) & 2.38 & 0.48 & 4438 & 2959.9M & 1.51$\times$ \\
% SFRon$\to$2SP & 2.51 (20.44$_{\pm 0.24}$) & 0.38 (99.06$_{\pm 0.09}$) & 0.34 (77.63$_{\pm 0.27}$) & 1.08 & \textit{\textbf{0.36}} & 4604 & \textbf{\textcolor{colorBest}{2916.8M}} & \textbf{\textcolor{colorBest}{1.53$\times$}} \\
% $\ell_1$-MU$\to$2SP & 33.97 (56.92$_{\pm 0.32}$) & 46.69 (51.98$_{\pm 0.72}$) & 34.47 (42.82$_{\pm 0.55}$) & 38.38 & 1.59 & 4305 & 2980.7M & 1.50$\times$ \\
% Un-pruning & \textbf{\textcolor{colorBest}{0.67}} (23.62$_{\pm 6.17}$) & 21.02 (77.66$_{\pm 6.24}$) & 9.21 (68.08$_{\pm 4.76}$) & 10.30 & 0.96 & 3585 & 3769.8M & 1.18$\times$ \\
% Bilevel & 15.58 (7.38$_{\pm 0.16}$) & 2.17 (96.51$_{\pm 0.17}$) & 5.16 (82.45$_{\pm 0.19}$) & 7.63 & 1.42 & 17814 & 2962.3M & 1.50$\times$ \\
% LLM-Eraser & 7.69 (15.26$_{\pm 0.28}$) & 3.08 (95.60$_{\pm 0.19}$) & 2.86 (80.15$_{\pm 0.22}$) & 4.54 & 0.51 & 2033 & 2974.7M & 1.50$\times$ \\
\hline
PruneForget & 
\textbf{\textcolor{colorBest}{1.16}} (21.79$_{\pm 0.17}$) & 0.35 (99.02$_{\pm 0.05}$) & \textbf{\textcolor{colorBest}{0.12}} (77.17$_{\pm 0.20}$) & \textbf{\textcolor{colorBest}{0.54}} & 0.37 & 4506 & 2942.9M & 1.51$\times$ \\
\specialrule{.15em}{.05em}{.05em}
\end{tabular}
}
\label{tab:cls_swin_rand_gp}
\end{table}

\begin{figure}[t]
    \centering
    % (a) Default Model
    \begin{subfigure}[b]{0.49\textwidth}
        \centering
        \label{fig:resnet_default}
        \includegraphics[width=\textwidth, trim={1cm 0 0.3cm 0}]{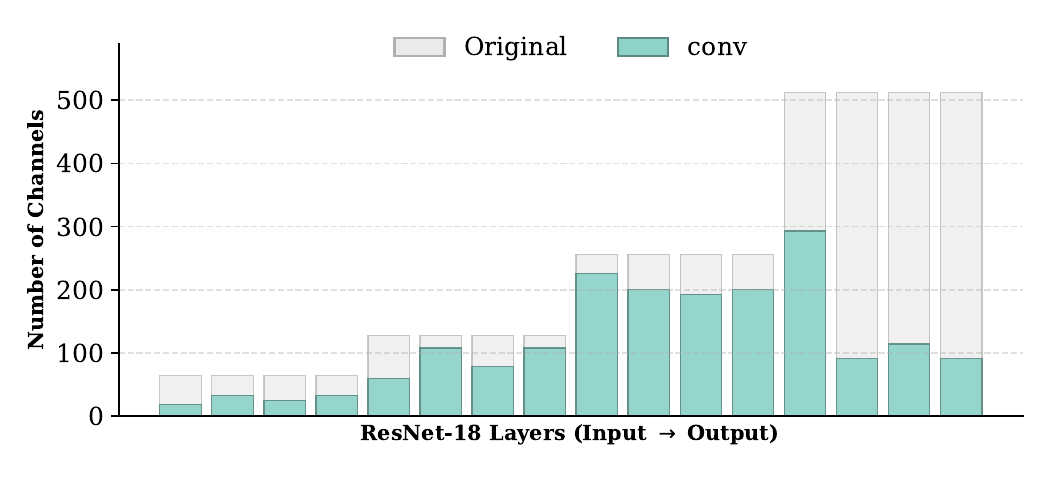}
        \caption{Pretrain$\to$3SP} 
    \end{subfigure}
    \hfill
    % (b) Our Method
    \begin{subfigure}[b]{0.49\textwidth}
        \centering
        \label{fig:resnet_ours}
        \includegraphics[width=\textwidth,trim={0.3cm 0 1cm 0}]{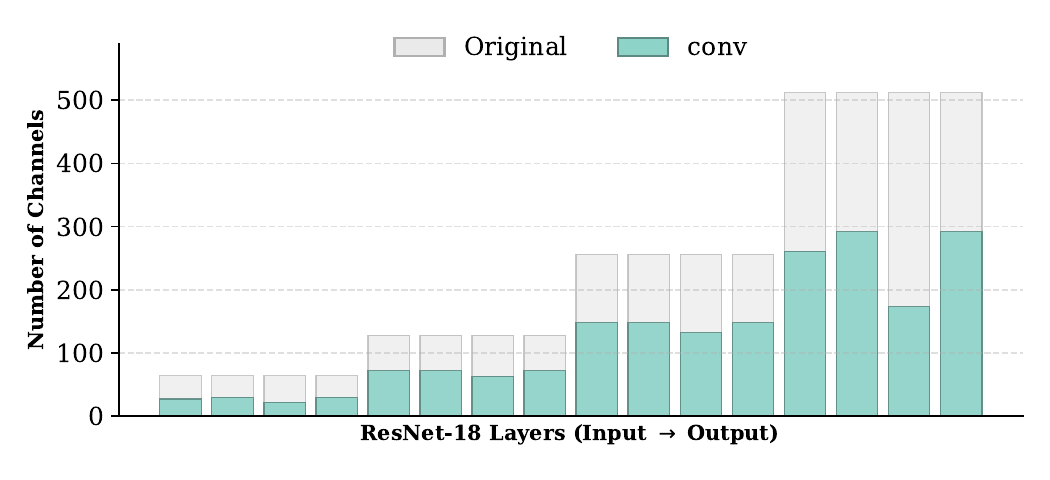}
        \caption{PruneForget}
    \end{subfigure}
    \caption{\textbf{Channel sparsity distribution of pruned ResNet-18.} 
    In (a), the pruning process operates independently of the unlearning phase; conversely, in (b), the proposed PruneForget integrates unlearning awareness into the pruning stage. This distinction leads to noticeably different parameter distributions across the network. Specifically, PruneForget tends to preserve significantly more channels in the final stage.
    }
% \vspace{-0.5cm}
    \label{fig:supp_sparsity}
\end{figure}

\begin{figure}[t]
    \centering
    % (a) Default Model
    \begin{subfigure}[b]{0.492\textwidth}
        \centering
        \label{fig:swin_default}
        \includegraphics[width=.9\textwidth, trim={1cm 0 0.3cm 0}]{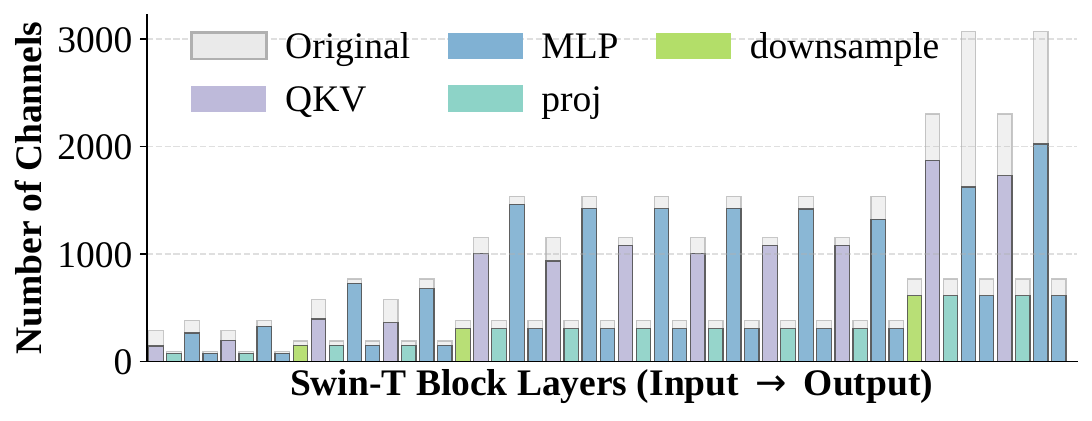}
        \caption{Pretrain$\to$2SP} 
    \end{subfigure}
    \hfill
    % (b) Our Method
    \begin{subfigure}[b]{0.492\textwidth}
        \centering
        \label{fig:swin_ours}
        \includegraphics[width=.9\textwidth,trim={0.3cm 0 1cm 0}]{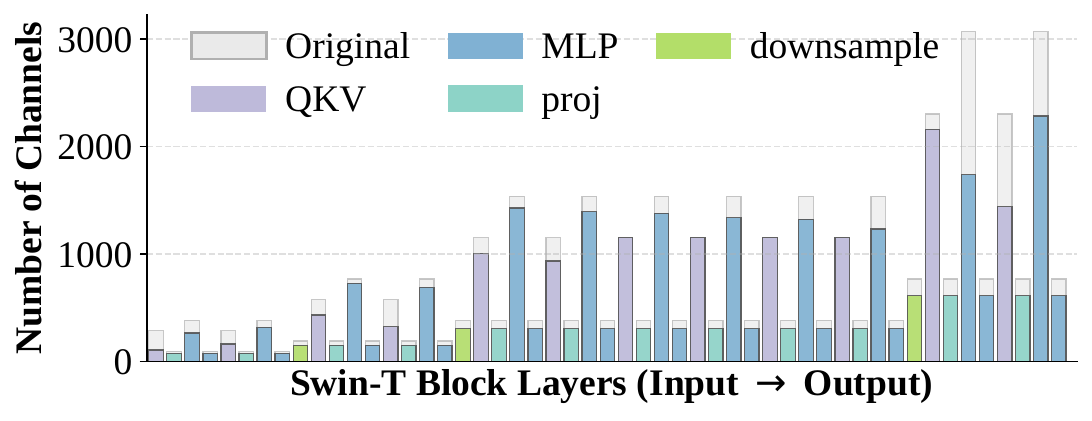}
        \caption{PruneForget}
    \end{subfigure}
    \caption{\textbf{Channel sparsity distribution of pruned Swin-T.} 
    In (a), the pruning process operates independently of the unlearning phase; conversely, in (b), the proposed PruneForget integrates unlearning awareness into the pruning stage. This results in the difference in pruned parameter distributions across the model. Specifically, we observe that PruneForget prunes the initial QKV layer more aggressively than in (a). While the majority of pruning for both methods occurs toward the end of the model, (b) more significantly prunes the QKV layer at the end.
    }
    \label{fig:sparsity}
\end{figure}
\section{Additional Results using Global Pruning}\label{sec:supp_global}

% \ray{TODO: Transition from uniform pruning to global a little smoother.}

In previous sections, we evaluated our method under uniform pruning, which removes an equal proportion of channels per layer to create hardware-friendly architectures. To achieve better accuracy retention, an alternative strategy is global pruning, which pools parameters across all layers to adaptively preserve the network's most critical components, which is supported by Group-level Pruning~\cite{fang2023depgraph} and Isomorphic Pruning~\cite{fang2024isomorphic}. Note that this approach results in varying MACs even at identical pruning ratios. We evaluate our method under this setting to demonstrate its effectiveness in a global pruning scenario, which also shows the structural impact of unlearn-aware pruning.

% \clover{TODO: Highlight difference b/w global vs. uniform pruning}

% \myparagraph{Visualization of sparsity patterns under global pruning.}
 We compare PruneForget against the unlearn-unaware baselines (Pretrain$\to$$n$SP) on ResNet-18 and Swin-T under 50\% random unlearning. 
% Unlike the uniform pruning evaluated in \tabref{tab:cls_cifar_rand_pr50} and \tabref{tab:cls_swin_rand_pr20}, global pruning keeps the total number of pruned channels constant while adaptively allocating layer-wise sparsity, resulting in varying MACs even at identical pruning ratios. 
As shown in \tabref{tab:cls_cifar_rand_gp} and \tabref{tab:cls_swin_rand_gp}, PruneForget consistently attains a lower {$\text{Avg }\Delta$} despite operating with fewer or comparable MACs. 

Next, we analyze the channel sparsity distributions from the different methods. In~\figref{fig:supp_sparsity}, we visualize the channel distribution for ResNet-18. We observe that integrating the unlearning objective directly into pruning yields different sparse architectures. PruneForget retains noticeably more channels in the deeper stages, whereas the baseline aggressively sparsifies the final layer. For Swin-T shown in \figref{fig:sparsity}, PruneForget prunes the attention (QKV) projections more heavily in both the initial and final blocks. While we did not observe an interpretable reason as to why PruneForget prunes these layers, this finding seems to be consistent with existing works on unlearning, where they also focus on optimizing the attention layers~\cite{gandikota2023erasing}.

% \input{sec/supp_limitation}

% \clearpage
% \input{sec/updated_exp}

\end{document}